\documentclass[pdflatex,sn-nature]{sn-jnl}
\usepackage{titlesec}

\titleformat{\subsubsection}[runin]
  {\normalfont\normalsize\bfseries}
  {}
  {0pt}
  {}
  [.\hspace{0.5em}]

\usepackage{graphicx}%
\usepackage{subcaption}
\usepackage{placeins}
\usepackage{multirow}%
\usepackage{amsmath,amssymb,amsfonts}%
\usepackage{amsthm}%
\usepackage{booktabs}%
\usepackage{setspace}
\usepackage{tikz}
\usepackage{adjustbox}
\usepackage[table]{xcolor}
\usepackage{mathtools,bm}
\usepackage{tabularx,array,longtable,rotating}
\usepackage{makecell}

\newcommand{\FL}{\mathcal{F}^{L}}
\newcommand{\FE}{\mathcal{F}^{E}}

\newcolumntype{Y}{>{\centering\arraybackslash}X}
\newcounter{extendeddatafigure}
\renewcommand{\theextendeddatafigure}{\arabic{extendeddatafigure}}

\newcommand{\edcaption}[2]{%
  \refstepcounter{extendeddatafigure}%
  \caption*{\textbf{Extended Data Fig. \theextendeddatafigure\ | #1} #2}%
}
\newcounter{extendeddatatable}
\renewcommand{\theextendeddatatable}{\arabic{extendeddatatable}}

\newcommand{\edtablecaption}[2]{%
  \refstepcounter{extendeddatatable}%
  \caption*{\textbf{Extended Data Table \theextendeddatatable\ | #1} #2}%
}
\theoremstyle{thmstyleone}%
\theoremstyle{thmstyletwo}%

\theoremstyle{thmstylethree}%

\unnumbered
\begin{document}

\title[Lagrangian--Eulerian Graph Neural ODEs]{Learning and Forecasting Multiscale Spatiotemporal Epidemic Dynamics with Lagrangian--Eulerian Graph Neural ODEs}

\author[1]{\fnm{Jinyi} \sur{Duan}}
\email{DJY1111@stu.xjtu.edu.cn}



\author*[3]{\fnm{Qian} \sur{Li}}
\email{liqian@xaut.edu.cn}

\author*[1,2]{\fnm{Biao} \sur{Tang}}
\email{biaotang@xjtu.edu.cn}

\affil[1]{
\orgdiv{School of Mathematics and Statistics},
\orgname{Xi'an Jiaotong University},
\orgaddress{
\city{Xi'an},
\postcode{710049},
\country{People's Republic of China}
}
}

\affil[2]{
\orgdiv{Center for Intersection of Mathematics and Life Sciences},
\orgname{Xi'an Jiaotong University},
\orgaddress{
\city{Xi'an},
\postcode{710049},
\country{People's Republic of China}
}
}

\affil[3]{
	\orgdiv{School of Mathematics and Statistics},
	\orgname{Xi'an University of Technology},
	\orgaddress{
		\city{Xi'an},
		\postcode{710054},
		\country{People's Republic of China}
	}
}

\abstract{Accurate medium- to long-range epidemic forecasting requires extrapolating high-dimensional, heterogeneous and nonstationary dynamics from partial observations. Yet hybrid models often leave implicit which observations inform which epidemic processes and how learned graph interactions modify system state. We introduce LE-EpiGNODE, which replaces generic message passing with process-specific, mechanism-guided neural operators: age-contact information drives Lagrangian infection generation, whereas mobility drives conservative Eulerian population transport within a structure-preserving universal differential equation. We evaluated LE-EpiGNODE across 14 controlled synthetic mechanisms spanning time-modulated, state-dependent, nonlinear and coupled dynamics, three complementary COVID-19 settings, 21 comparator models and five targeted ablations. Forecasting gains tended to become more pronounced at longer horizons; at the longest horizon, LE-EpiGNODE achieved the lowest WMAPE in all three real-world datasets, reducing error by 17.7--40.0\% relative to the best baseline. Controlled ablations linked these gains to the Lagrangian–Eulerian decomposition and process-aligned priors, while synthetic ground truth showed that learned operators and induced state fluxes track known generating dynamics. These results support mechanism-organized graph neural dynamics as a route to reliable long-horizon extrapolation with directly testable, process-level mechanistic interpretation beyond trajectory accuracy.}

\maketitle
\setstretch{1.08}
\section{Introduction}
Reliable epidemic forecasts are critical for anticipating healthcare demand, allocating public-health resources, and adapting interventions before transmission conditions change. Yet accurate medium- to long-range prediction remains difficult because epidemic trajectories emerge from high-dimensional, nonstationary, and multiscale dynamics. Age-dependent contact patterns shape local infection generation~\cite{monod2021age,mistry2021mixing}, whereas population mobility reorganizes spatial connectivity and redistributes epidemiological states~\cite{chang2021}. These heterogeneous processes are coupled through a shared epidemic state, so changes at one scale can reshape transmission at another. Meanwhile, the underlying system is only partially observed, with incidence, contact, and mobility data providing complementary but incomplete views of latent states and interactions. Long-horizon forecasts therefore require models that extrapolate these coupled dynamics, rather than simply extend observed epidemic trajectories~\cite{rahmandad2022,kendall2024}.

Existing approaches address complementary parts of this problem. Mechanistic age-structured and metapopulation models explicitly organize contact-mediated transmission and mobility-driven coupling~\cite{colizza2007,brockmann2011,arenas2020}, but their interaction functions become difficult to specify as heterogeneity, dimensionality, and temporal variation increase. Graph-based spatiotemporal models instead learn nonlinear relational structure directly from data~\cite{Liu2024ARO,deng2020colagnn,panagopoulos2021transfer,xie2023epignn,qiu2024msgnn}. Hybrid models have connected these directions at progressively deeper levels: MepoGNN and M-Graphormer infer parameters, contact-related quantities, or propagation graphs that enter predefined epidemic dynamics~\cite{cao2023mepognn,wang2025multi}, whereas EARTH incorporates a learnable transmission graph into an epidemiology-aware Neural ODE~\cite{Wan_EpiODE_ICML25}. More generally, universal differential equations (UDEs) embed trainable unresolved functions within partially known governing equations~\cite{rackauckas2020universal}. In epidemic modeling, this paradigm has been used to learn difficult-to-specify spatiotemporal heterogeneous processes and unresolved nonlinear functions within compartmental dynamics, further supporting mechanistic reconstruction and longer-term prediction~\cite{tang2024heterogeneity,ma2026kanude}. These developments establish that neural flexibility and mechanistic epidemic structure can be combined at parameter, graph, and continuous-dynamics levels.

Despite this progress, two related issues remain. Heterogeneous observations are often combined within shared transmission representations, learned graphs, or parameter-estimation modules, leaving the correspondence between an information source and the physical process it directly constrains largely implicit. Mechanistic knowledge is also commonly introduced through predicted parameters, compartmental equations, or global training constraints, while the semantics and state action of graph messages themselves remain generic. Moving from discrete graph models to Neural ODEs or UDEs does not by itself resolve this issue: a continuous neural vector field can still obscure which physical process a learned interaction represents and how that interaction is permitted to modify the epidemic state. This distinction is consequential rather than merely interpretive. Age-structured contact and mobility constrain processes with fundamentally different state actions~\cite{mistry2021mixing,nouvellet2021reduction}: infection generation creates compartmental transitions, whereas population transport redistributes existing states under conservation~\cite{gomez2018critical}. If both are absorbed into a common latent interaction mechanism, different learned components may compensate for one another while reproducing similar trajectories. Such compensation can remain hidden within the training regime but need not persist during extrapolation beyond it~\cite{gallo2022identifiability}; consequently, trajectory accuracy alone does not ensure dynamically faithful long-horizon prediction. The remaining methodological challenge is therefore to align heterogeneous information with process-specific operators and to let epidemic mechanisms determine the physical semantics and admissible state action of their graph messages.

\begin{figure*}[!t]
    \centering
    \includegraphics[width=\textwidth]{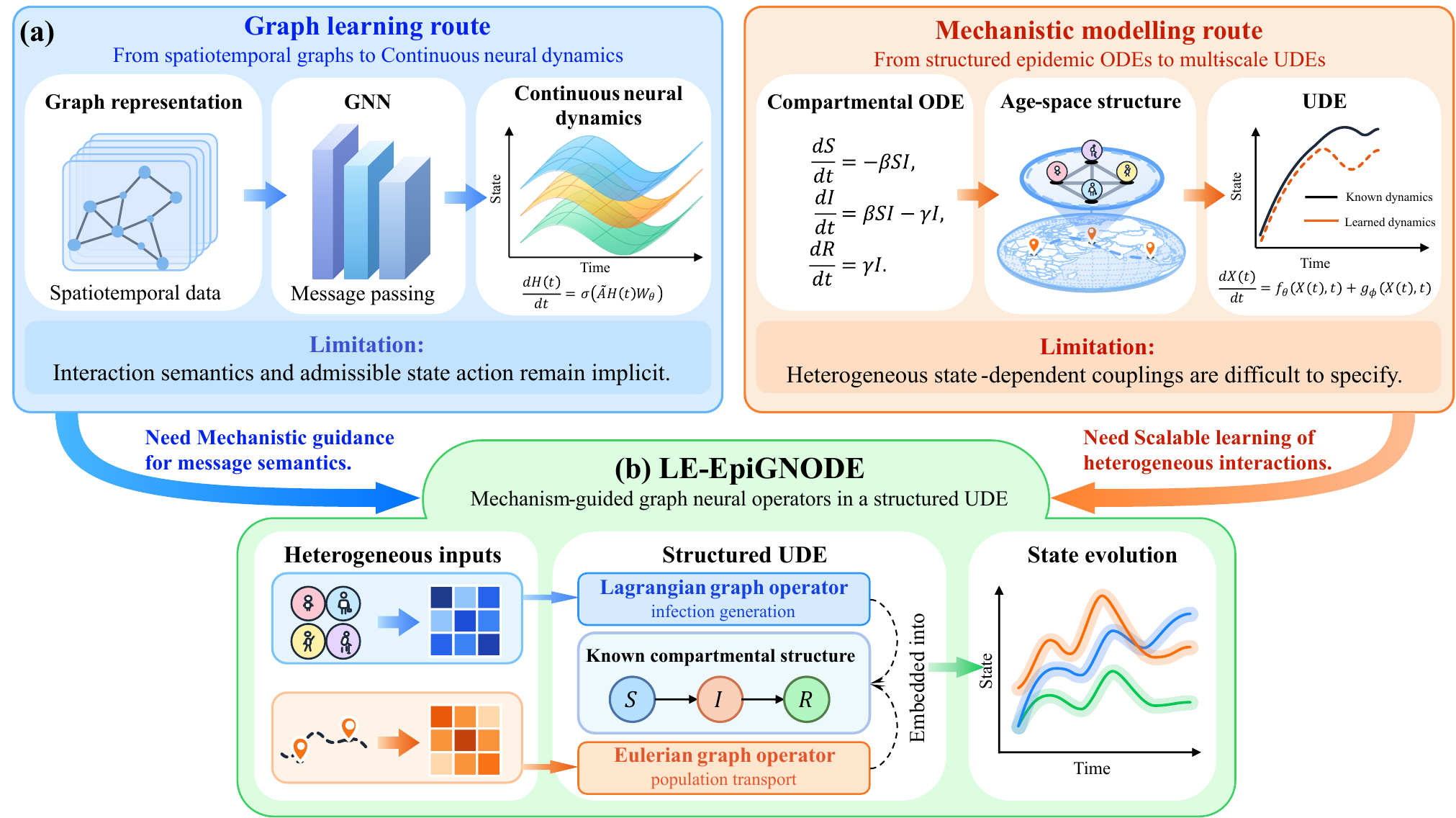}

    \caption{
    \textbf{Conceptual motivation and framework of LE-EpiGNODE.}
    \textbf{(a)}, Conceptual synthesis motivating LE-EpiGNODE. From the graph-learning side, flexible spatiotemporal message passing progresses toward continuous neural interactions but requires governing-dynamics guidance to define message semantics and admissible state updates. From the mechanistic side, compartmental models progress toward age- and spatially structured UDEs with learnable unresolved functions but require scalable graph neural learning to represent high-dimensional heterogeneous couplings. These reciprocal requirements converge on mechanism-guided graph neural operators embedded within structured governing dynamics.
    \textbf{(b)}, LE-EpiGNODE instantiates this convergence through a Lagrangian infection-generation operator and an Eulerian population-transport operator embedded in a structured UDE. Graph neural modules learn state- and time-dependent interaction strengths, while the epidemic mechanism specifies process semantics, admissible state updates, and conservation structure.
    }
    \label{fig:motivation-framework}
\end{figure*}

These limitations motivate the two complementary design routes summarized in Fig.~\ref{fig:motivation-framework}a. From the graph-learning side, spatiotemporal message passing can be extended toward continuous neural interactions, but these models still require governing-dynamics guidance to determine what a learned interaction means and how it may update the epidemic state. From the mechanistic side, compartmental ODEs can be extended to age- and spatially structured UDEs, but high-dimensional, state-dependent couplings across population groups and regions require a scalable mechanism for learning heterogeneous relational structure. The two routes therefore expose complementary deficiencies and converge on the same operator-level requirement: graph neural learning is needed to represent unresolved heterogeneous interactions, while structured governing dynamics are needed to determine the physical meaning and admissible state action of those learned interactions.

We address this requirement with LE-EpiGNODE, a Lagrangian--Eulerian Epidemic Graph Neural Ordinary Differential Equation framework (Fig.~\ref{fig:motivation-framework}b). Mechanistic information enters at three connected levels: it defines distinct relational topologies for the two epidemic processes, assigns heterogeneous observations to the corresponding pathways, and prescribes how messages from each pathway may modify the epidemic state. Age-contact information informs a Lagrangian infection-generation operator, whereas mobility information informs an Eulerian population-transport operator. Both operators are nonlinear, state- and time-dependent components of the governing vector field and interact through a shared epidemic state. Lagrangian messages generate infections from infectious sources through prescribed compartmental transitions, whereas Eulerian messages conservatively redistribute the full compartmental state across regions. Graph neural networks therefore learn unresolved interaction strengths and temporal responses, while the epidemic mechanism fixes what each message represents and what state transformation it is permitted to induce. This mechanism-organized, operator-level message passing is the central methodological distinction of LE-EpiGNODE.

\section{Results}

We evaluated LE-EpiGNODE across 14 controlled synthetic generation conditions, three complementary real-world COVID-19 settings, 21 mechanistic, sequence, graph-based and hybrid comparators, and five targeted ablations. The synthetic systems provide known generating dynamics for controlled tests of long-horizon extrapolation and dynamical fidelity, whereas the real-world datasets test age-structured, mobility-driven and coupled age--space forecasting under partial observation. Detailed data-generation, dataset, mobility-input and baseline configurations are provided in Supplementary Tables~1--8.

\subsection{Long-horizon forecasting across heterogeneous epidemic dynamics}

We first examined whether the Lagrangian--Eulerian formulation can maintain forecasting accuracy as predictions extend beyond the observed epidemic phase. The central question is whether the learned operators can extrapolate changing interaction dynamics, including explicit temporal modulation, endogenous state feedback and nonlinear responses, rather than only continue the local trajectory near the forecast origin.

\subsubsection{Synthetic epidemic forecasting}

The synthetic benchmark deliberately varies dynamical form as well as structural scale. Its 14 generation conditions comprise four Lagrangian infection-generation mechanisms---the Baseline Mass-Action Contact Mechanism (BMCM), Time-Decaying Contact Mechanism (TDCM), Infection-Avoidant Contact Mechanism (IACM) and Nonlinear Mass-Action Contact Mechanism (NMCM)---and two Eulerian population-transport mechanisms, Baseline Risk-Aware Gravity Mobility (BRGM) and Controlled Risk-Aware Gravity Mobility (CRGM), together with their eight pairwise multiscale combinations. These mechanisms span baseline mass-action, explicit temporal modulation, endogenous state feedback and nonlinear responses across isolated and coupled dynamical settings. Representative trajectories of the isolated mechanisms are shown in Supplementary Fig.~1, while their complementary age- and region-level signatures in the multiscale systems are illustrated in Supplementary Fig.~2. Within each structural setting, all models were trained using the same training period and forecast origin and were evaluated at horizons of 1, 4, 12 and 20 weeks. This common protocol probes the transition from short-term trajectory continuation to long-horizon extrapolation under qualitatively different forms of epidemic nonstationarity and multiscale coupling; complete mechanism definitions and generation settings are provided in Supplementary Methods, section 1.1, and Supplementary Tables~1--3.

\begin{figure}[!t]
    \centering

    \includegraphics[width=1.0\linewidth]{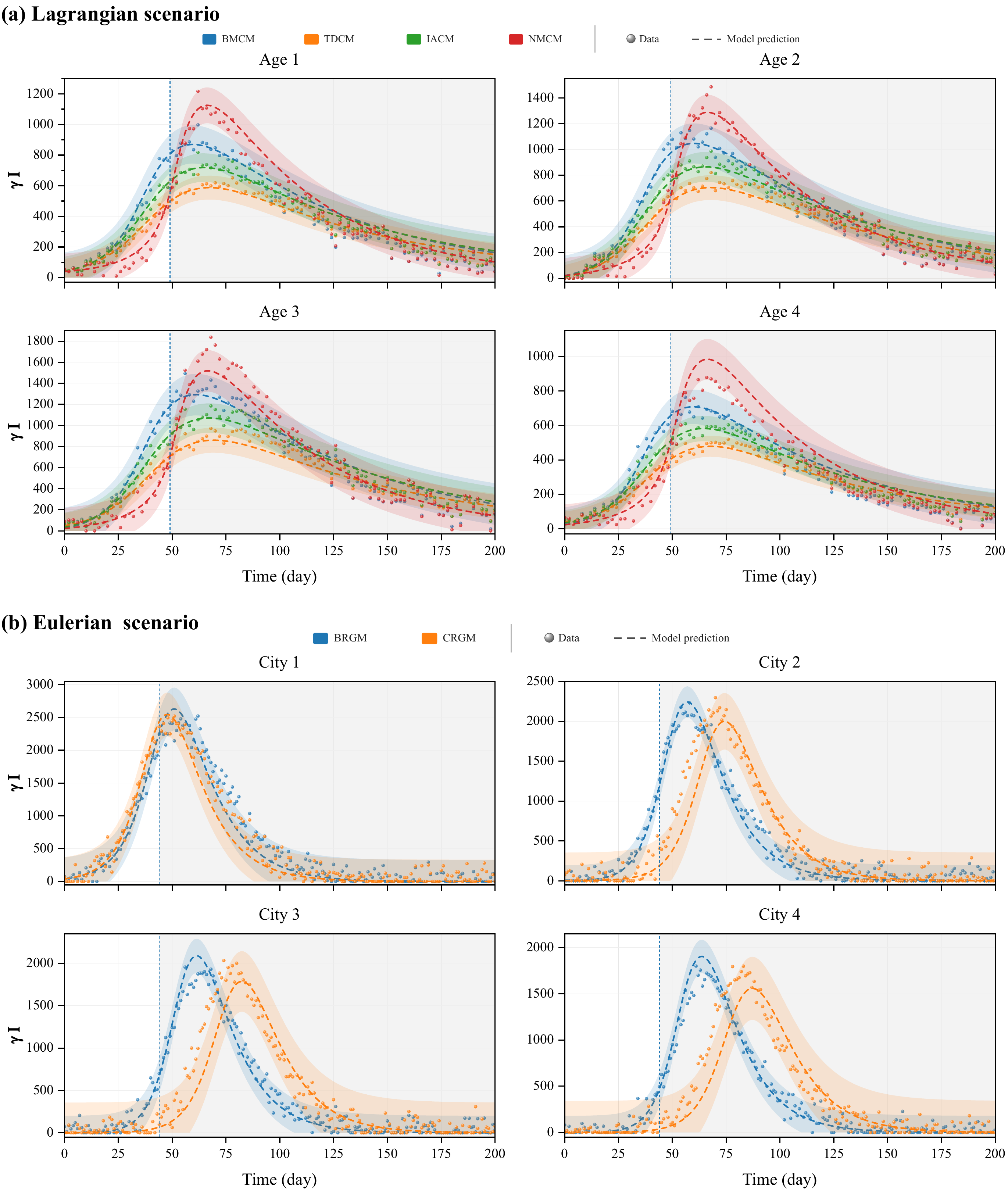}

    \caption{
    \textbf{Fitting and forecasting trajectories in the synthetic scenarios.}
    \textbf{(a)}, Lagrangian scenario, where infection generation is governed by age-structured transmission dynamics within regions.
    \textbf{(b)}, Eulerian scenario, where population transport couples regions while local infection generation and removal remain active.
    The vertical dashed line marks the forecast origin, and the gray-shaded region denotes the out-of-sample forecasting period.
    Colored envelopes show post-hoc normal-reference residual bands centered on the model predictions.
    All models are trained using only observations before the forecast origin.
    }
    \label{fig:sim-curves}
\end{figure}

Figure~\ref{fig:sim-curves} shows representative fitting and forecasting trajectories under distinct Lagrangian and Eulerian interaction mechanisms, while Extended Data Fig.~\ref{fig:ed-multiscale-forecast} extends the analysis to all eight paired mechanisms in the multiscale setting. Because population structure and the observation protocol are held fixed within each structural setting, differences in temporal response and post-peak behavior reflect the prescribed interaction dynamics. Table~\ref{tab:forecast-summary} summarizes the corresponding WMAPE comparisons across forecast horizons, while Extended Data Table~\ref{tab:ed-synthetic-forecasting} provides the complete MAE and WMAPE results for all comparator models.

Across the 12 setting--horizon combinations, LE-EpiGNODE achieved the lowest WMAPE in 10 cases. Here, the best-performing baseline refers to the comparator with the lowest WMAPE within the same epidemic setting and forecast horizon. In the Lagrangian setting, LE-EpiGNODE achieved the lowest error at 1, 4 and 12 weeks; at 20 weeks, LE-EpiGNODE and MepoSIR were nearly tied, with WMAPE values of 12.45\% and 12.21\%, respectively. In the Eulerian setting, LE-EpiGNODE achieved the lowest WMAPE at all four horizons. At 20 weeks, its WMAPE was 21.11\%, compared with 61.63\% for SIR, the best-performing baseline at that horizon, representing a 65.7\% relative reduction in WMAPE. Across all comparator models evaluated at this horizon, the relative reduction ranged from 65.7\% to 84.0\%, indicating a substantial separation from the full baseline set under mobility-driven long-horizon extrapolation. In the multiscale setting, LE-EpiGNODE's ranking improved with horizon, reaching first place at 4, 12 and 20 weeks. At 20 weeks, its WMAPE was 13.36\%, compared with 25.52\% for MepoSIR, the best-performing baseline at that horizon, representing a 47.6\% relative reduction in WMAPE.

These results reveal an overall horizon-dependent pattern. The relative advantage of LE-EpiGNODE was larger at the medium and long horizons than at the shortest horizon in the Eulerian and multiscale settings. At the shortest horizon in the multiscale setting, several sequence and spatiotemporal graph baselines achieved lower errors. The strong performance of MepoSIR in the comparatively simple Lagrangian setting also shows that a well-matched mechanistic inductive bias can remain effective. This pattern is consistent with the central design hypothesis of LE-EpiGNODE, while the structural ablations below provide a more direct test of its Lagrangian--Eulerian organization.

\subsubsection{Real-world epidemic forecasting}

\begin{figure*}[!htpb]
    \centering

    \includegraphics[width=\textwidth]{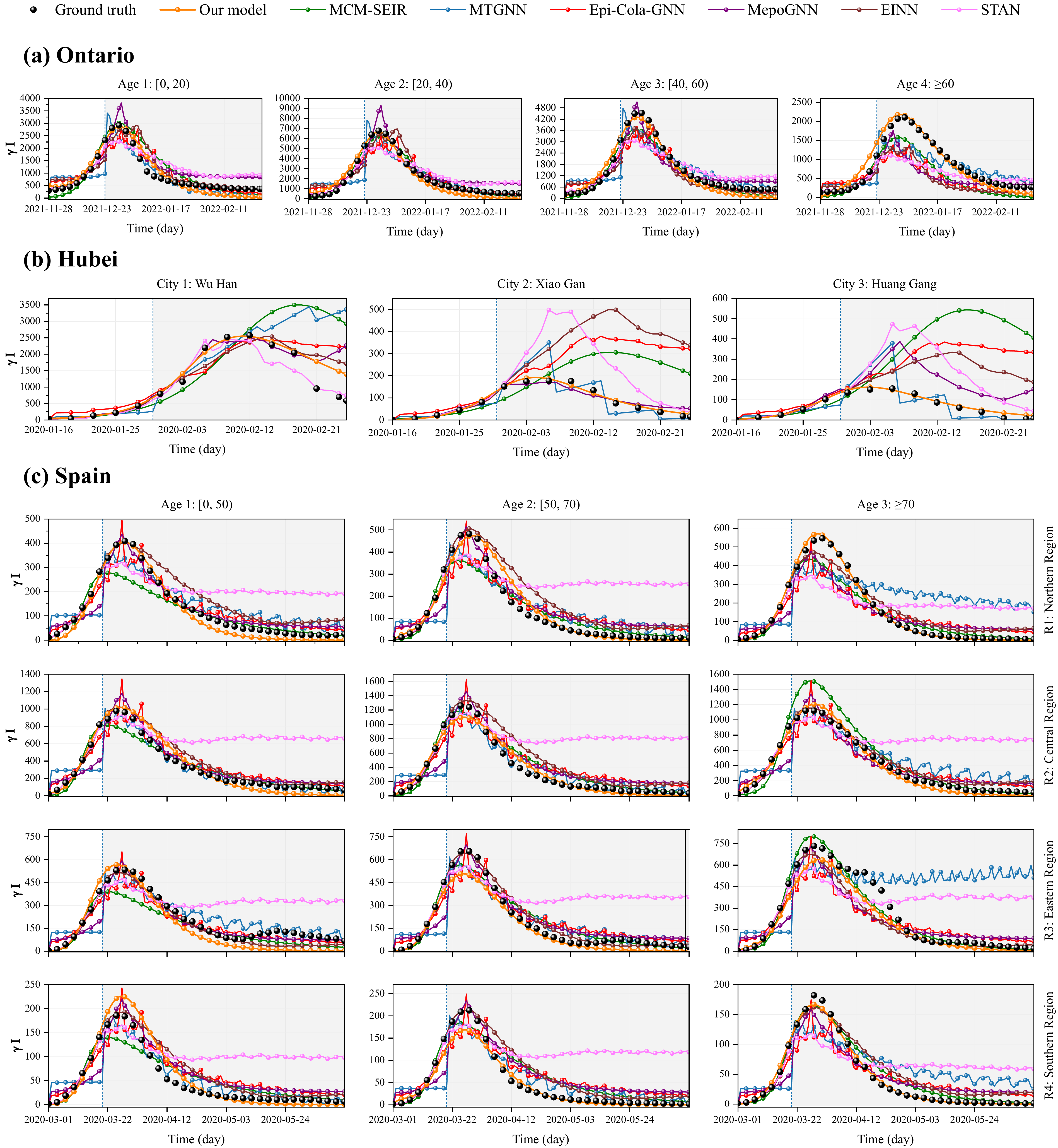}

    \caption{
    \textbf{Representative fitting and forecasting trajectories of LE-EpiGNODE and selected baselines on the real-world COVID-19 datasets.}
    \textbf{(a)}, Ontario.
    \textbf{(b)}, Hubei.
    \textbf{(c)}, Spain.
    The vertical dashed line marks the forecast origin, and the gray-shaded region denotes the out-of-sample forecasting period.
    Models are trained using observations before the forecast origin.
    Ontario, Hubei, and Spain correspond to the Lagrangian, Eulerian, and multiscale propagation settings, respectively.
    }
    \label{fig:real-forecast}

\end{figure*}

We next examined whether the forecasting performance observed in the controlled systems extends to real epidemic observations. We evaluated LE-EpiGNODE on three COVID-19 datasets representing complementary epidemic settings: age-structured dynamics in Ontario, Canada; mobility-coupled spatial dynamics across three major cities in Hubei, China, without age stratification; and coupled age--space dynamics in Spain. All models used only observations available before the forecast origin. Dataset construction, forecasting windows and implementation details are described in Methods. Figure~\ref{fig:real-forecast} presents representative fitting and forecasting trajectories across the three epidemic settings. Table~\ref{tab:forecast-summary} summarizes the corresponding forecasting performance, while Extended Data Table~\ref{tab:ed-real-forecasting} provides the complete MAE and WMAPE comparisons across all evaluated forecast horizons and comparator models.

LE-EpiGNODE achieved the lowest WMAPE at every evaluated horizon in Ontario and Hubei. In Spain, LE-EpiGNODE ranked fourth at 7 days and third at 14 days, then ranked first at both 49 and 93 days. At the longest horizon in Ontario, LE-EpiGNODE achieved a WMAPE of 16.20\%, compared with 21.76\% for MCM-SEIR, the best-performing baseline at that horizon, representing a 25.6\% relative reduction in WMAPE. In Hubei, the corresponding values were 15.87\% for LE-EpiGNODE and 26.45\% for MepoGNN, representing a 40.0\% relative reduction. In Spain, LE-EpiGNODE achieved 20.43\% compared with 24.81\% for MCM-SEIR, representing a 17.7\% relative reduction. Spain's stronger relative performance at longer horizons parallels the synthetic multiscale pattern.

\begin{table*}[!t]
\centering
\caption{
\textbf{Forecasting performance across synthetic and real-world settings.}
Values are WMAPE (\%); lower values indicate better forecasting.
Each WMAPE entry is reported as LE-EpiGNODE / best-performing baseline for the corresponding setting and forecast horizon, with the lower value shown in bold.
The corresponding baseline model and the rank of LE-EpiGNODE among all evaluated models are also reported.
Shaded rows denote the longest evaluated horizon, for which the relative WMAPE reduction of LE-EpiGNODE against the best-performing baseline is shown; negative values indicate higher error for LE-EpiGNODE.
Complete MAE and WMAPE results across all comparator models and forecast horizons are provided in Extended Data Tables~\ref{tab:ed-synthetic-forecasting} and~\ref{tab:ed-real-forecasting}.
}
\label{tab:forecast-summary}

\scriptsize
\renewcommand{\arraystretch}{1.2}
\setlength{\tabcolsep}{4pt}

\begin{adjustbox}{max width=\textwidth}
\begin{tabular}{llclcc}
\toprule
\textbf{Setting}
& \textbf{Horizon}
& \textbf{WMAPE (\%)}
& \textbf{Best baseline}
& \textbf{LE rank}
& \textbf{Reduction (\%)} \\
\midrule

\multicolumn{6}{@{}l}{\textbf{(a) Synthetic scenarios}} \\
\addlinespace[3pt]

Lagrangian
& 1 week
& \textbf{5.67} / 6.31
& DCRNN
& 1
& \\

&
4 weeks
& \textbf{5.84} / 7.91
& DCRNN
& 1
& \\

&
12 weeks
& \textbf{8.45} / 9.19
& MepoSIR
& 1
& \\

\rowcolor{black!5}
&
20 weeks
& 12.45 / \textbf{12.21}
& MepoSIR
& 2
& $-2.0$ \\

\addlinespace[4pt]

Eulerian
& 1 week
& \textbf{13.41} / 26.82
& Graph WaveNet
& 1
& \\

&
4 weeks
& \textbf{16.44} / 44.91
& AGCRN
& 1
& \\

&
12 weeks
& \textbf{18.32} / 58.12
& SIR
& 1
& \\

\rowcolor{black!5}
&
20 weeks
& \textbf{21.11} / 61.63
& SIR
& 1
& \textbf{65.7} \\

\addlinespace[4pt]

Multiscale
& 1 week
& 17.16 / \textbf{13.23}
& DCRNN
& 6
& \\

&
4 weeks
& \textbf{13.50} / 27.17
& DCRNN
& 1
& \\

&
12 weeks
& \textbf{10.65} / 23.04
& MepoSIR
& 1
& \\

\rowcolor{black!5}
&
20 weeks
& \textbf{13.36} / 25.52
& MepoSIR
& 1
& \textbf{47.6} \\

\midrule

\multicolumn{6}{@{}l}{\textbf{(b) Real-world COVID-19 datasets}} \\
\addlinespace[2pt]

Ontario
& 7 days
& \textbf{3.43} / 7.42
& M-Graphormer
& 1
& \\

&
14 days
& \textbf{4.49} / 12.17
& MCM-SEIR
& 1
& \\

&
35 days
& \textbf{7.80} / 18.33
& MCM-SEIR
& 1
& \\

\rowcolor{black!5}
&
67 days
& \textbf{16.20} / 21.76
& MCM-SEIR
& 1
& \textbf{25.6} \\

\addlinespace[4pt]

Hubei
& 7 days
& \textbf{17.15} / 19.22
& AGCRN
& 1
& \\

&
14 days
& \textbf{8.07} / 14.41
& MepoGNN
& 1
& \\

\rowcolor{black!5}
&
26 days
& \textbf{15.87} / 26.45
& MepoGNN
& 1
& \textbf{40.0} \\

\addlinespace[4pt]

Spain
& 7 days
& 9.61 / \textbf{3.46}
& EINN
& 4
& \\

&
14 days
& 9.66 / \textbf{6.09}
& EINN
& 3
& \\

&
49 days
& \textbf{14.09} / 17.63
& MepoGNN
& 1
& \\

\rowcolor{black!5}
&
93 days
& \textbf{20.43} / 24.81
& MCM-SEIR
& 1
& \textbf{17.7} \\

\bottomrule
\end{tabular}
\end{adjustbox}
\end{table*}

Forecasting in Hubei and Spain additionally requires population transport to be extrapolated beyond the observation window. The auxiliary mobility network predicts future origin--destination (OD) mobility indices from the evolving epidemic state and time without using future OD observations. Its WMAPE increases to 40.15\% at 26 days in Hubei and 29.06\% at 93 days in Spain (Extended Data Table~\ref{tab:ed-od-errors}; Extended Data Figs.~\ref{fig:ed-hubei-mobility} and~\ref{fig:ed-spain-mobility}), whereas the corresponding epidemic-forecast WMAPE values remain at 15.87\% and 20.43\%. This separation suggests that the Eulerian pathway can retain forecast-relevant transport structure despite imperfect pointwise prediction of future OD mobility values. Because mobility and epidemic WMAPE quantify different prediction targets, this comparison is interpreted qualitatively rather than as a direct error ratio.

\subsection{Process decomposition and aligned priors drive forecasting gains}

To identify which components underlie the long-horizon forecasting gains, we evaluated five targeted ablations of the Lagrangian--Eulerian decomposition, the two dynamical pathways and their corresponding priors, as defined in Methods. We first examined these effects in the controlled synthetic systems and then tested whether the same process--information correspondence persisted in the real-world datasets. Table~\ref{tab:combined-ablation} reports the complete MAE and WMAPE results across all evaluated forecast horizons for both synthetic and real-world settings.

\subsubsection{Synthetic-system ablations}

Removing the Lagrangian--Eulerian structure consistently degraded long-horizon forecasting in the synthetic systems (Table~\ref{tab:combined-ablation}). In the multiscale setting, \textit{w/o LE structure} increased the 20-week WMAPE from 13.36\% to 58.26\%. Removing the pathway corresponding to the represented process had a similarly pronounced effect: WMAPE increased from 12.45\% to 50.13\% without $\mathcal{G}_{L}$ in the Lagrangian setting and from 21.11\% to 78.27\% without $\mathcal{G}_{E}$ in the Eulerian setting. Removing either pathway also degraded performance when the two processes were coupled. The particularly large deterioration after removing the Eulerian pathway is consistent with the synthetic mobility mechanisms, in which transport is both epidemic-state dependent and, for CRGM, explicitly time modulated while remaining population conserving. In the multiscale setting, the large gap between the full model and \textit{w/o LE structure} further supports separating infection generation from transport when two processes with different state actions evolve simultaneously, rather than asking one generic graph vector field to represent both.

The process-aligned priors provided additional gains. Removing the age-contact prior increased the 20-week WMAPE from 12.45\% to 72.08\% in the Lagrangian setting, whereas removing OD-informed mobility pretraining increased it from 21.11\% to 44.57\% in the Eulerian setting. The same variants also increased errors in the multiscale setting. Because the contact and OD priors enter different pathways, this process-matched deterioration supports the value of information--process alignment: each prior contributes most directly when assigned to the dynamical process it is intended to constrain.

\subsubsection{Real-world ablations}

The real-world experiments showed a similar correspondence between processes and information sources (Table~\ref{tab:combined-ablation}), with each pathway contributing in accordance with the dynamical role it represents in the given setting. At the longest horizon, \textit{w/o LE structure} increased WMAPE from 16.20\% to 34.78\% in Ontario, from 15.87\% to 45.45\% in Hubei and from 20.43\% to 55.43\% in Spain. Removing $\mathcal{G}_{L}$ had the largest effect in age-structured Ontario, increasing WMAPE to 58.20\%, whereas removing $\mathcal{G}_{E}$ increased WMAPE to 26.06\% in mobility-coupled Hubei. Both pathways contributed in Spain, where age-dependent infection generation and spatial transport act in the same system: removing $\mathcal{G}_{L}$ or $\mathcal{G}_{E}$ increased WMAPE from 20.43\% to 35.86\% or 23.44\%, respectively.

The effects of the priors followed the same correspondence. Removing the age-contact prior increased the longest-horizon WMAPE from 16.20\% to 28.08\% in Ontario, while removing mobility pretraining increased it from 15.87\% to 32.22\% in Hubei. Both priors also improved forecasting in Spain, with smaller individual effects than in the settings that more directly represented the corresponding process.

\begin{table*}[!t]
\centering
\caption{
\textbf{Ablation performance across synthetic and real-world settings.}
Each cell reports MAE / WMAPE (\%); lower values indicate better forecasting.
The complete LE-EpiGNODE model is highlighted in bold.
Ablations targeting the processes and priors emphasized in each setting are shown.
A dash denotes a forecast horizon that was not evaluated in that setting.
}
\label{tab:combined-ablation}

\scriptsize
\renewcommand{\arraystretch}{1.08}
\setlength{\tabcolsep}{4pt}

\begin{adjustbox}{max width=\textwidth}
\begin{tabular}{lcccc}
\toprule

\multicolumn{5}{@{}l}{\textbf{(a) Synthetic scenarios}} \\
\addlinespace[2pt]

\textbf{Variant}
& \textbf{1 week}
& \textbf{4 weeks}
& \textbf{12 weeks}
& \textbf{20 weeks} \\
\midrule

\multicolumn{5}{c}{\textit{Lagrangian}} \\
\midrule
LE-EpiGNODE
& \textbf{44.38 / 5.67}
& \textbf{51.09 / 5.84}
& \textbf{58.66 / 8.45}
& \textbf{63.30 / 12.45} \\

w/o LE structure
& 69.48 / 8.87
& 101.01 / 11.54
& 214.94 / 30.97
& 374.75 / 73.74 \\

w/o $\mathcal G_L$
& 176.15 / 22.49
& 185.64 / 21.22
& 207.70 / 29.93
& 254.75 / 50.13 \\

w/o $e_{uv}^{L}$
& 54.58 / 6.97
& 58.87 / 6.73
& 128.53 / 18.52
& 366.32 / 72.08 \\

\midrule
\multicolumn{5}{c}{\textit{Eulerian}} \\
\midrule
LE-EpiGNODE
& \textbf{152.42 / 13.41}
& \textbf{241.97 / 16.44}
& \textbf{171.32 / 18.32}
& \textbf{123.12 / 21.11} \\

w/o LE structure
& 350.17 / 30.80
& 725.73 / 49.30
& 761.46 / 81.44
& 751.05 / 128.76 \\

w/o $\mathcal G_E$
& 606.44 / 53.34
& 1009.28 / 68.56
& 723.92 / 77.43
& 456.54 / 78.27 \\

w/o $\operatorname{MLP}_{\mathrm{mob}}$ pretraining
& 680.31 / 59.83
& 633.91 / 43.06
& 398.60 / 42.63
& 259.98 / 44.57 \\

\midrule
\multicolumn{5}{c}{\textit{Multiscale}} \\
\midrule
LE-EpiGNODE
& \textbf{123.68 / 17.16}
& \textbf{136.39 / 13.50}
& \textbf{113.54 / 10.65}
& \textbf{110.01 / 13.36} \\

w/o LE structure
& 156.55 / 21.72
& 347.94 / 34.45
& 477.26 / 44.75
& 479.65 / 58.26 \\

w/o $\mathcal G_L$
& 303.89 / 42.16
& 376.24 / 37.25
& 340.62 / 31.94
& 272.62 / 33.12 \\

w/o $e_{uv}^{L}$
& 163.46 / 22.68
& 185.74 / 18.39
& 165.42 / 15.51
& 142.63 / 17.33 \\

w/o $\mathcal G_E$
& 427.17 / 59.27
& 680.95 / 67.41
& 805.11 / 75.50
& 635.77 / 77.23 \\

w/o $\operatorname{MLP}_{\mathrm{mob}}$ pretraining
& 228.77 / 31.74
& 290.25 / 28.73
& 288.06 / 27.01
& 222.20 / 26.99 \\

\midrule
\multicolumn{5}{@{}l}{\textbf{(b) Real-world COVID-19 datasets}} \\
\addlinespace[2pt]

\multicolumn{5}{c}{\textit{Ontario}} \\
\midrule
\textbf{Variant}
& \textbf{7 days}
& \textbf{14 days}
& \textbf{35 days}
& \textbf{67 days} \\
\midrule

LE-EpiGNODE
& \textbf{128.57 / 3.43}
& \textbf{165.71 / 4.49}
& \textbf{190.60 / 7.80}
& \textbf{248.66 / 16.20} \\

w/o LE structure
& 985.10 / 26.31
& 1348.30 / 36.52
& 835.93 / 34.21
& 533.63 / 34.78 \\

w/o $\mathcal G_L$
& 2028.34 / 54.18
& 2102.24 / 56.93
& 1346.87 / 55.11
& 893.13 / 58.20 \\

w/o $e_{uv}^{L}$
& 408.94 / 10.92
& 400.34 / 10.84
& 593.86 / 24.30
& 430.92 / 28.08 \\

\midrule
\multicolumn{5}{c}{\textit{Hubei}} \\
\midrule
\textbf{Variant}
& \textbf{7 days}
& \textbf{14 days}
& \textbf{26 days}
& {} \\
\midrule

LE-EpiGNODE
& \textbf{88.04 / 17.15}
& \textbf{58.25 / 8.07}
& \textbf{100.34 / 15.87}
& -- \\

w/o LE structure
& 117.18 / 22.82
& 116.65 / 16.16
& 287.37 / 45.45
& -- \\

w/o $\mathcal G_E$
& 105.46 / 20.54
& 84.11 / 11.65
& 164.80 / 26.06
& -- \\

w/o $\operatorname{MLP}_{\mathrm{mob}}$ pretraining
& 123.33 / 24.02
& 225.73 / 31.28
& 203.72 / 32.22
& -- \\

\midrule
\multicolumn{5}{c}{\textit{Spain}} \\
\midrule
\textbf{Variant}
& \textbf{7 days}
& \textbf{14 days}
& \textbf{49 days}
& \textbf{93 days} \\
\midrule

LE-EpiGNODE
& \textbf{54.78 / 9.61}
& \textbf{54.63 / 9.66}
& \textbf{43.89 / 14.09}
& \textbf{37.60 / 20.43} \\

w/o LE structure
& 104.06 / 18.26
& 126.40 / 22.36
& 98.86 / 31.74
& 101.98 / 55.43 \\

w/o $\mathcal G_L$
& 160.33 / 28.13
& 153.58 / 27.16
& 99.87 / 32.06
& 65.97 / 35.86 \\

w/o $e_{uv}^{L}$
& 110.94 / 19.47
& 100.91 / 17.85
& 59.70 / 19.17
& 43.12 / 23.43 \\

w/o $\mathcal G_E$
& 132.73 / 23.29
& 123.99 / 21.93
& 58.30 / 18.72
& 43.13 / 23.44 \\

w/o $\operatorname{MLP}_{\mathrm{mob}}$ pretraining
& 132.73 / 23.29
& 113.18 / 20.02
& 58.08 / 18.65
& 42.62 / 23.16 \\

\bottomrule
\end{tabular}
\end{adjustbox}
\end{table*}


\subsection{Process-specific operators enable dynamical validation and mechanistic interpretation beyond trajectory accuracy}

Trajectory-level accuracy measures output fidelity, whereas LE-EpiGNODE also exposes internal dynamical quantities with explicit process semantics and mechanism-defined state actions. This makes it possible to test whether apparently accurate trajectories or accumulated structures are supported by faithful evolving dynamics. We therefore evaluate these quantities against synthetic ground truth and then examine their process-level organization in real-world forecasting, where the true generating operators are unobserved.

\subsubsection{Dynamical fidelity in synthetic systems}

The three evaluation levels mirror the construction of LE-EpiGNODE. The instantaneous process-specific operators $\mathcal{F}^{b}$, with $b\in\{L,E\}$ denoting the Lagrangian or Eulerian branch, quantify the interaction strengths learned for each dynamical process by the neural components; the corresponding state fluxes $\Phi$ quantify the dynamical effect of those interactions after the mechanism-defined state action is applied; and the temporally integrated Lagrangian and Eulerian interaction fluxes (LIFs and EIFs) summarize cumulative process organization over the evaluation interval. The hierarchy $\mathcal{F}^{b}(t)\rightarrow\Phi(t)\rightarrow\mathrm{LIF/EIF}$ links learned interaction strengths, their instantaneous dynamical effects and cumulative process organization. WMAPE measures relative magnitude error, PCC measures agreement in the temporal and edge-wise variation of instantaneous operators and fluxes, and SRCC measures rank agreement of integrated interaction structures. Lower WMAPE and higher PCC or SRCC therefore indicate stronger agreement with the known generating dynamics; complete definitions are provided in Methods.

Figure~\ref{fig:integrated-flux}a--d compares the true and inferred integrated interaction structures, while Extended Data Table~\ref{tab:ed-dynamics-consistency} reports the complete quantitative dynamical-fidelity metrics. Complete edge-wise operator trajectories across the synthetic mechanisms are provided in Supplementary Figs.~3--21. LE-EpiGNODE showed strong agreement with the known generating dynamics at both instantaneous and accumulated levels across the three synthetic settings. In the Lagrangian setting, $\mathcal{F}^{L}$-PCC, $\Phi^{L}$-PCC and LIF-SRCC were 0.980, 0.959 and 0.971, respectively. The corresponding values in the Eulerian setting were 0.996, 0.981 and 0.923. In the multiscale setting, the model retained strong agreement for both pathways: $\mathcal{F}^{L}$-PCC and $\Phi^{L}$-PCC were 0.936 and 0.920 for the Lagrangian branch, while $\mathcal{F}^{E}$-PCC and $\Phi^{E}$-PCC were 0.999 and 0.990 for the Eulerian branch; the corresponding LIF-SRCC and EIF-SRCC values were 0.835 and 0.998.

The prior-removal variants provide a stricter test of the information--process alignment and reveal an important distinction between accumulated structure and evolving dynamics. Removing the age-contact prior left the LIF-SRCC unchanged at 0.971 in the Lagrangian setting, but increased the $\Phi^{L}$-WMAPE from 25.89\% to 196.68\% and reduced the $\Phi^{L}$-PCC from 0.959 to 0.074. Thus, an apparently preserved cumulative interaction ranking can coexist with a severe loss of agreement in instantaneous flux variation. This discrepancy is particularly relevant for time-modulated and state-dependent mechanisms, for which tracking the temporal evolution of the operator is a stricter requirement than reproducing its integrated ordering. Removing OD-informed mobility pretraining caused a broader deterioration in the Eulerian setting, increasing the $\mathcal{F}^{E}$-WMAPE from 6.93\% to 53.49\% and reducing the EIF-SRCC from 0.923 to 0.448. These results show that process-aligned priors constrain not only trajectory forecasts but also the evolving dynamical quantities represented by their corresponding pathways. More generally, the defined process semantics and state actions make these discrepancies observable as explicit dynamical quantities within LE-EpiGNODE.

\subsubsection{Process-level interpretation in real epidemics}

\begin{figure*}[p]
    \centering

    \includegraphics[
        width=\textwidth,
        height=0.82\textheight,
        keepaspectratio
    ]{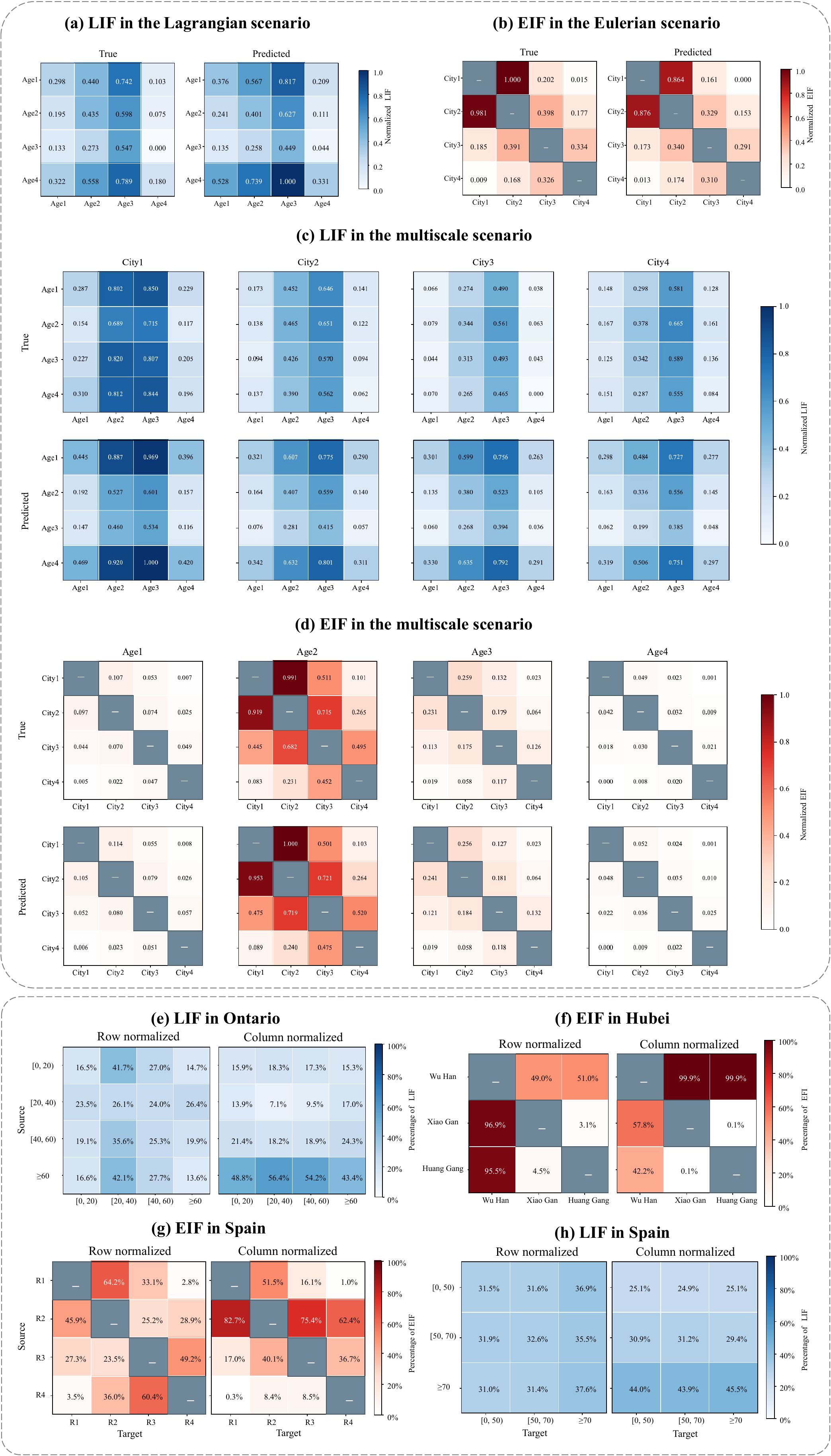}

\caption{
\textbf{Integrated interaction fluxes for dynamical validation and process-level interpretation.}
\textbf{(a)}, LIF in the Lagrangian synthetic scenario.
\textbf{(b)}, EIF in the Eulerian synthetic scenario.
\textbf{(c)}, LIF in the multiscale synthetic scenario.
\textbf{(d)}, EIF in the multiscale synthetic scenario.
For synthetic systems, true and inferred fluxes are integrated over the full simulation horizon and min--max normalized for visualization.
\textbf{(e)}, LIF in Ontario.
\textbf{(f)}, EIF in Hubei.
\textbf{(g)}, EIF in Spain.
\textbf{(h)}, LIF in Spain.
For real-world datasets, LIF is aggregated over regions and EIF over age groups to characterize age- and region-level interaction structure, respectively.
Row and column normalization show the target distribution from each source and the source composition of each target.
Diagonal conservation terms are excluded from EIF.
}
    \label{fig:integrated-flux}
\end{figure*}

Because the generating operators are unavailable for the COVID-19 datasets, the real-world analysis uses the same operator- and flux-level quantities validated in synthetic systems to inspect how the fitted model organizes age-related and spatial information during forecasting. The LIF and EIF maps in Fig.~\ref{fig:integrated-flux}e--h summarize the node-level composition of cumulative interactions along the two dynamical pathways. LIF is aggregated over the spatial dimension to characterize the relative organization of cumulative infection-generation flux across age groups, whereas EIF is aggregated over the age dimension to characterize the relative organization of cumulative infectious-state transport flux across regions. The resulting normalized profiles therefore provide complementary views of the age and spatial structures learned by LE-EpiGNODE.

The Lagrangian patterns reveal heterogeneous interactions across age groups. In Ontario, messages from several source groups are directed predominantly towards the $[20,40)$ group, while the $\geq60$ group also provides substantial input to multiple targets. Spain similarly exhibits unequal interaction strengths across age groups. Age-specific differences in social contacts, susceptibility, infectivity and household exposure are known to shape the age distribution of COVID-19 transmission~\cite{monod2021age,caetano2024drivers,hu2021infectivity,bi2021household}, providing relevant epidemiological context for interpreting these age-related message patterns.

The Eulerian patterns show clearer spatial directionality. In Hubei, the learned fluxes form a Wuhan-centered structure, with surrounding cities receiving strong cumulative messages originating from Wuhan. Population movements from Wuhan played an important role in shaping the early spatial distribution of COVID-19 across China~\cite{jiang2020mobility,jia2020population}. In Spain, central regions produce stronger outward messages towards other regions, where regional mobility, multiple epidemic introductions, lockdowns and spatial spillovers affected the development of the first epidemic wave~\cite{mazzoli2021interplay,orea2022spanish}.

The LIF and EIF maps characterize information propagation within LE-EpiGNODE. Accordingly, we interpret them as process-level explanations of how age-related and spatial information contributes to the model's forecasts, while contact rates, individual transmission events and causal transmission pathways remain outside the scope of this analysis. Their value is that the same quantities whose dynamical fidelity is testable against synthetic ground truth remain inspectable in real forecasts, providing a consistent bridge from controlled process validation to real-world model interpretation.


\section{Discussion}

LE-EpiGNODE was motivated by a mismatch between the organization of epidemic processes and the way heterogeneous information is often incorporated into forecasting models. Age-contact patterns and population mobility are not interchangeable covariates: they constrain infection generation and population transport, which act on the epidemic state in fundamentally different ways and may both vary with epidemic conditions and time. Mechanistic models preserve these process roles but can become restrictive when high-dimensional interactions are nonlinear or nonstationary, whereas flexible graph models can learn heterogeneous dependencies without explicitly fixing what a learned interaction represents or how it is allowed to modify the state. LE-EpiGNODE addresses this gap by embedding separate Lagrangian infection-generation and Eulerian population-transport operators within a structured UDE, aligning age-contact and mobility information with their corresponding pathways while retaining mechanism-defined state actions. The results indicate that this organization is particularly useful when forecasting moves beyond local trajectory continuation and requires extrapolation of evolving multiscale dynamics.

The forecasting results show broad gains that tend to become more pronounced at longer prediction horizons. LE-EpiGNODE achieved the lowest WMAPE at the longest horizon in all three real-world datasets. Its relative advantage increased from the shortest to the longest horizon in the Eulerian synthetic setting and in Hubei, while the multiscale synthetic setting and Spain shifted from weaker short-horizon rankings to first place at longer horizons. In Ontario, the relative WMAPE reduction was larger at intermediate horizons than at the longest horizon. In the Lagrangian synthetic setting, LE-EpiGNODE was slightly less accurate than MepoSIR at 20 weeks. Taken together, these results indicate that LE-EpiGNODE remains competitive near the observed regime while its advantage tends to become more consistent when
forecasting requires extrapolation of evolving state-dependent, time-varying and
coupled multiscale interaction dynamics.

The ablation and process-level analyses jointly clarify the contribution of mechanism-organized message passing. Replacing the Lagrangian–Eulerian formulation with a generic Graph Neural ODE, removing individual pathways, or omitting their aligned contact and OD priors degraded forecasting in process-dependent ways across synthetic and real-world settings. These results support separating infection generation from conservative transport and aligning each information source with the process it constrains. Beyond forecasting, mechanism-defined message semantics and state actions make learned interaction strengths, induced state fluxes and integrated interaction structures directly testable against synthetic ground truth. LE-EpiGNODE showed strong agreement with the generating dynamics, whereas removing the contact prior left the cumulative Lagrangian ranking nearly unchanged despite a sharp deterioration in instantaneous flux agreement. This discrepancy shows that agreement in accumulated structure does not guarantee fidelity of evolving dynamics, particularly under time-modulated or state-dependent mechanisms. The $\mathcal{F}^{b}\rightarrow\Phi\rightarrow\mathrm{LIF/EIF}$ hierarchy thus connects learned interactions, their dynamical effects and cumulative process organization, enabling direct dynamical validation in controlled systems and process-level model interpretation in real epidemics, where the true generating operators remain unobserved.

Several limitations remain. First, latent epidemic states and learned process fluxes may be only partially identifiable from reported incidence, particularly when observations are sparse, delayed or affected by changes in reporting. Additional epidemiological and behavioral measurements could provide stronger constraints on the hidden dynamics. Second, the auxiliary mobility network is pretrained on historical OD observations and frozen during epidemic-model training. Although its outputs remain dependent on epidemic state and time, its fixed parameters may limit adaptation to abrupt mobility changes caused by interventions or behavioral shifts. Third, the synthetic mechanisms are controlled and deliberately stylized, whereas the real-world evaluation is restricted to three COVID-19 settings; broader validation across pathogens, intervention regimes, spatial scales and surveillance systems is therefore needed.

The central methodological contribution of LE-EpiGNODE is mechanism-organized operator learning: heterogeneous observations are mapped to the physical processes they inform, graph messages are given explicit process semantics, and the governing dynamics constrain the state transformations those messages can induce while neural modules learn unresolved state- and time-dependent interaction strengths. This design is mechanism-informed in construction and makes the learned dynamics interpretable and, in controlled systems, directly testable against known generating processes. The resulting operator-level organization links flexible continuous-time learning with mechanistic structure, supports long-horizon epidemic extrapolation, and enables process-level mechanistic interpretation beyond trajectory accuracy. In real data, these quantities are used for model interpretation rather than causal mechanism estimation. More broadly, the same principle may be useful for multiscale dynamical systems in which different data streams constrain processes with distinct physical state actions.

\section{Methods}

\subsection{LE-EpiGNODE dynamical formulation}

LE-EpiGNODE formulates multiscale epidemic dynamics as a structured universal differential equation (UDE) in which unresolved infection-generation and population-transport interactions are represented by process-specific graph neural operators (Fig.~\ref{fig:motivation-framework}b). The mechanistic backbone defines the relational role and admissible state action of each pathway, while the neural components learn their state- and time-dependent interaction strengths.

We consider a spatial epidemic system consisting of $M$ regions, each divided into $A$ age groups. For region $k$ and age group $j$, let $S_j^k(t)$, $I_j^k(t)$ and $R_j^k(t)$ denote the susceptible, infectious and removed populations, respectively. The population size of age group $i$ in region $k$ is denoted by $P_i^k(t)=S_i^k(t)+I_i^k(t)+R_i^k(t)$. Let $C_{ij}^{k}$ represent the contact intensity from infectious-source age group $i$ to susceptible-target age group $j$ in region $k$, $\beta(t)$ denote a potentially time-varying transmission coefficient, $\gamma_j^k$ denote the removal rate, $\eta_j$ denote the mobility tendency of age group $j$, and $L^{lk}$ denote the transport intensity from region $l$ to region $k$. A representative age-structured spatial epidemic system is

\begin{equation}
\left\{
\begin{aligned}
\frac{\mathrm d S_j^k}{\mathrm d t}
&=
-\sum_{i=1}^{A}\beta(t) C_{ij}^{k}S_j^{k}\frac{I_i^{k}}{P_i^{k}}
+\sum_{l=1}^{M}\eta_j L^{lk}S_j^{l},
\\
\frac{\mathrm d I_j^k}{\mathrm d t}
&=
\sum_{i=1}^{A}\beta(t) C_{ij}^{k}S_j^{k}\frac{I_i^{k}}{P_i^{k}}
-\gamma_j^k I_j^{k}
+\sum_{l=1}^{M}\eta_j L^{lk}I_j^{l},
\\
\frac{\mathrm d R_j^k}{\mathrm d t}
&=
\gamma_j^k I_j^{k}
+\sum_{l=1}^{M}\eta_j L^{lk}R_j^{l}.
\end{aligned}
\right.
\label{eq:mechanistic_model}
\end{equation}
The diagonal entries of the transport operator account for outflow, and $\sum_{k=1}^{M}L^{lk}=0$ ensures conservation of the transported population from each origin region $l$. In the observation model used below, case reporting is assumed to coincide with the $I\rightarrow R$ transition, so the reported-incidence rate is identified with the removal flow $\gamma_j^k I_j^k(t)$.

Equation~\eqref{eq:mechanistic_model} separates two relational processes with different state actions: contact-mediated infection generation changes compartmental state, whereas population transport redistributes existing states across regions without changing compartmental identity. LE-EpiGNODE retains this mechanistic organization but replaces the unresolved, potentially nonstationary interaction functions with learnable operators. The resulting formulation preserves compartment-transition direction, explicit dependence of infection generation on infectious source states, and population conservation during spatial transport.

Accordingly, LE-EpiGNODE follows the universal differential equation paradigm~\cite{rackauckas2020universal}:
\[
\frac{\mathrm d X}{\mathrm d t}
=
f_{\mathrm{mech}}
\!\left(
X;
\mathcal{F}^{L}(X,t),
\mathcal{F}^{E}(X,t)
\right),
\]
where \(f_{\mathrm{mech}}\) denotes the mechanism-defined governing vector field that encodes the retained compartmental organization and mechanistic constraints, while \(\mathcal{F}^{L}\) and \(\mathcal{F}^{E}\) represent the learnable Lagrangian infection-generation and Eulerian population-transport operators, respectively. Unlike a single generic neural correction, the learnable operators are organized according to processes with distinct mechanistic roles, while the governing mechanisms determine their admissible state transformations.

To represent the two processes, we define a shared age--space node set in which each region--age combination corresponds to one node:
$$
v=(k,j)\in\mathcal{V},
\qquad
k\in\{1,\ldots,M\},
\qquad
j\in\{1,\ldots,A\}.
$$
The total number of nodes is $|\mathcal{V}|=MA$. For node $v=(k,j)$, the epidemic state is
$$
X_v(t)=
\left(
X_v^S(t),
X_v^I(t),
X_v^R(t)
\right)^\top
=
\left(
S_j^k(t),
I_j^k(t),
R_j^k(t)
\right)^\top.
$$

We construct two graph views over the shared node set $\mathcal{V}$. The Lagrangian graph $\mathcal{G}_{L}=(\mathcal{V},\mathcal{E}_{L})$ represents within-region age-dependent interactions, with
$$
\mathcal{E}_{L}
=
\left\{
(k,i)\rightarrow(k,j):
k\in\{1,\ldots,M\},
\ i,j\in\{1,\ldots,A\}
\right\}.
$$
An edge $u=(k,i)\rightarrow v=(k,j)$ indicates that infectious individuals in age group $i$ can contribute to infection generation in age group $j$ within the same region.

The Eulerian graph $\mathcal{G}_{E}=(\mathcal{V},\mathcal{E}_{E})$ represents mobility-driven population transport, with
$$
\mathcal{E}_{E}
=
\left\{
(l,j)\rightarrow(k,j):
l\neq k,\ 
l,k\in\{1,\ldots,M\},
\ j\in\{1,\ldots,A\}
\right\}.
$$
An edge $u=(l,j)\rightarrow v=(k,j)$ represents the movement of individuals in age group $j$ from region $l$ to region $k$. Although $\mathcal{G}_{L}$ and $\mathcal{G}_{E}$ share the same nodes and epidemic states, they define distinct relational supports for the two dynamical processes: $\mathcal{G}_{L}$ supports infection-generating interactions, whereas $\mathcal{G}_{E}$ supports spatial redistribution of existing epidemic states. Their process identity is therefore fixed before the interaction strengths are learned.

\subsection{Process-specific graph neural operators in the UDE formulation}

We parameterize the unresolved interactions on these two graph supports using separate state- and time-dependent neural operators. The Lagrangian operator $\mathcal{F}^{L}$ is informed by age-contact structure and learns infection-generation interaction strengths over $\mathcal{G}_{L}$, whereas the Eulerian operator $\mathcal{F}^{E}$ is informed by historical origin--destination mobility and learns population-transport strengths over $\mathcal{G}_{E}$. Their outputs remain coupled through the shared epidemic state, while the mechanistic vector field maps them to different state actions: infection-generating compartment transitions for Lagrangian messages and zero-sum redistribution of the full compartmental state for Eulerian messages.

\subsubsection{Lagrangian infection-generation operator}

The Lagrangian operator models contact-driven infection generation within each region. Following the infection term in Eq.~\eqref{eq:mechanistic_model}, it depends on the infectious-source state, the susceptible-target state, the age-contact structure and continuous time. Age-contact information is introduced as a structural prior whose contribution can vary with the evolving epidemic state.

For a directed Lagrangian edge $u=(k,i)\rightarrow v=(k,j)$, the contact-prior representation is
$$
e_{uv}^{L}=W_{L}C_{ij}^{k},
$$
where $C_{ij}^{k}$ denotes the contact intensity from source age group $i$ to target age group $j$ in region $k$, and $W_{L}$ is a learnable projection into the latent edge space. Given the source state, target state, contact prior and time, the Lagrangian interaction strength is defined as
\begin{equation}
\mathcal{F}_{uv}^{L}(t)
=
\frac{
\operatorname{softplus}
\left[
\operatorname{MLP}_{L}
\left(
X_u(t)
\parallel
X_v(t)
\parallel
e_{uv}^{L}
\parallel
t
\right)
\right]
}{
P_u(t)
},
\qquad
u\rightarrow v\in\mathcal{E}_{L},
\label{eq:lagrangian_operator}
\end{equation}
where $\parallel$ denotes feature concatenation and $P_u(t)=X_u^S(t)+X_u^I(t)+X_u^R(t)$ is the population associated with the source node $u=(k,i)$. The softplus transformation ensures a non-negative interaction strength. Equation~\eqref{eq:lagrangian_operator} uses the observed contact structure as a prior rather than a fixed transmission coefficient, allowing the effective interaction to vary with the source state, target state and time.

The infection-generation flux along edge $u\rightarrow v$ is
$$
\Phi_{uv}^{L}(t)=\mathcal{F}_{uv}^{L}(t)X_u^I(t).
$$
Here, $\mathcal{F}_{uv}^{L}(t)$ represents the effective interaction strength, while multiplication by $X_u^I(t)$ preserves the explicit dependence of infection generation on the infectious-source population.

\subsubsection{Eulerian population-transport operator}

The Eulerian operator models the redistribution of existing epidemic states across regions. Regional populations and spatial separation define a gravity-type representation of spatial connectivity~\cite{simini2012}, while epidemic activity and time allow transport intensity to vary with transmission conditions~\cite{nouvellet2021reduction}. Historical origin--destination observations are used to inform the auxiliary mobility function; the corresponding pretraining procedure is described below.

For an Eulerian edge $u=(l,j)\rightarrow v=(k,j)$, the spatial representation between regions $l$ and $k$ is
$$
e_{lk}^{E}
=
\frac{(N^l)^{\kappa_1}(N^k)^{\kappa_2}}
{\exp(\lambda d_{lk})},
\qquad
l,k\in\{1,\ldots,M\},
$$
where $N^l$ and $N^k$ are the fixed reference total populations of the source and destination regions, $d_{lk}$ is their geographical distance or generalized transport impedance, and $\kappa_1$, $\kappa_2$ and $\lambda$ control the population and distance effects.

Regional epidemic activity is constructed from the current infectious states and the removal rates:
$$
\bar I^k(t)
=
\frac{1}{N^k}
\sum_{j=1}^{A}
\gamma_j^kX_{(k,j)}^{I}(t),
$$
where $\bar I^k(t)$ represents the normalized reported-incidence activity implied by the infectious states in region $k$. The auxiliary mobility network takes a dataset-specific input vector $z_{lk}(t)$ that combines source and destination epidemic activity with the spatial representation and the temporal or regional features used in each dataset. The predicted OD mobility value is
\begin{equation}
\hat{\mathcal{M}}^{lk}(t)
=
\operatorname{softplus}
\left[
\operatorname{MLP}_{\mathrm{mob}}
\left(
z_{lk}(t)
\right)
\right].
\label{eq:mobility_predictor}
\end{equation}
For the synthetic and Hubei settings, $z_{lk}(t)$ contains the source and destination epidemic activities, the gravity-type spatial representation $e_{lk}^{E}$ and normalized time. For Spain, additional periodic temporal features and source and destination region identities are included. The complete dataset-specific input definitions are provided in Supplementary Methods, section 1.2, and summarized in Supplementary Table~5. The pretraining and forecasting use of $\operatorname{MLP}_{\mathrm{mob}}$ are specified in the learning procedure below.

The predicted OD mobility value is converted into the Eulerian population-transport operator, whose off-diagonal entries are effective per-capita transport rates, as
\begin{equation}
\mathcal{F}_{uv}^{E}(t)
=
\begin{cases}
\eta_j
\operatorname{softplus}
\left(
\alpha
\dfrac{\hat{\mathcal{M}}^{lk}(t)}{N^l}
+
\epsilon
\right),
& u=(l,j),\ v=(k,j),\ l\neq k,
\\[1ex]
-\displaystyle\sum_{w\neq u}\mathcal{F}_{uw}^{E}(t),
& u=v,
\end{cases}
\label{eq:eulerian_operator}
\end{equation}
where $\eta_j\geq0$ denotes the relative mobility tendency of age group $j$, while $\alpha$ and $\epsilon$ are model calibration parameters that map population-normalized mobility observations to effective transport rates. The mobility data and their construction are described in the corresponding source studies~\cite{yuan2020wuhan,poncedeleon2021flowmaps}. The diagonal entry is defined as the negative sum of the outgoing interactions, giving
$$
\sum_v\mathcal{F}_{uv}^{E}(t)=0.
$$
This zero-sum constraint allows the Eulerian operator to redistribute existing epidemic states without creating or removing population mass. The corresponding infectious-state transport flux is
$$
\Phi_{uv}^{E}(t)=\mathcal{F}_{uv}^{E}(t)X_u^I(t).
$$
The two fluxes therefore have different meanings: $\Phi_{uv}^{L}(t)$ represents newly generated infections associated with an infectious source, whereas $\Phi_{uv}^{E}(t)$ represents the spatial transport of an existing infectious population.

\subsubsection{Coupled epidemic dynamics}

The process-specific operators are coupled through the shared epidemic state, while fixed transition vectors determine how their learned interaction strengths act on the compartments. We define
$$
b_{SI}
=
\begin{bmatrix}
-1\\
1\\
0
\end{bmatrix},
\qquad
b_{IR}
=
\begin{bmatrix}
0\\
-1\\
1
\end{bmatrix},
$$
where $b_{SI}$ represents the transition from susceptible to infectious states and $b_{IR}$ represents the transition from infectious to removed states.

Combining Lagrangian infection generation, the self-loop compartment transition and Eulerian population transport, we denote the complete state- and time-dependent operator by $\mathcal{F}=\mathcal{F}_{\theta}(X(t),t)$. Its action on the epidemic state at node $v$ is
\begin{equation}
\begin{aligned}
\frac{\mathrm d X_v}{\mathrm d t}
=
(\mathcal{F}X)_v
={}&
\underbrace{
\left(
\sum_{u\in\mathcal{N}_{L}(v)\cup\{v\}}
\mathcal{F}_{uv}^{L}(t)X_u^{I}(t)
\right)b_{SI}
}_{\text{Lagrangian message passing}}
\\
&+
\underbrace{
\gamma_vX_v^{I}(t)b_{IR}
}_{\text{self-loop message}}
+
\underbrace{
\sum_{u\in\mathcal{N}_{E}(v)\cup\{v\}}
\mathcal{F}_{uv}^{E}(t)X_u(t)
}_{\text{Eulerian message passing}} .
\end{aligned}
\label{eq:dual_view_dynamics}
\end{equation}
Here, $\mathcal{N}_{L}(v)$ and $\mathcal{N}_{E}(v)$ denote the incoming neighbors of node $v$ in the Lagrangian and Eulerian graphs, respectively, and $\gamma_v=\gamma_j^k$ for $v=(k,j)$. The three terms correspond to infection generation, local compartmental progression and conservative spatial transport. Equation~\eqref{eq:dual_view_dynamics} therefore separates what is learned from how the state is updated: $\mathcal{F}^{L}$ and $\mathcal{F}^{E}$ learn edge-wise interaction strengths, whereas the transition vectors and zero-sum transport structure fix their mechanism-defined state actions.

The transition vectors satisfy
$$
\boldsymbol{1}^{\top}b_{SI}
=
\boldsymbol{1}^{\top}b_{IR}
=
0,
\qquad
\boldsymbol{1}
=
\begin{bmatrix}
1\\
1\\
1
\end{bmatrix},
$$
and the Eulerian operator satisfies $\sum_v\mathcal{F}_{uv}^{E}(t)=0$. Consequently, neither the compartment transitions nor spatial transport change the total population represented by the system. Population conservation is therefore encoded directly in the vector field rather than imposed through an additional penalty term.

Let $f_{\theta}(X,t)$ denote the complete vector field given by the right-hand side of Eq.~\eqref{eq:dual_view_dynamics}, equivalently $f_{\theta}(X,t)=\mathcal{F}_{\theta}(X,t)X$. The epidemic trajectory is obtained by
$$
X(t)
=
X(t_0)
+
\int_{t_0}^{t}
f_{\theta}\!\left(X(\tau),\tau\right)
\,\mathrm d\tau.
$$
This notation distinguishes the complete nonlinear vector field from the process-specific operators $\mathcal{F}^{L}$ and $\mathcal{F}^{E}$.

\subsubsection{SEIR extension for real-world epidemics}

The formulation above uses an SIR backbone for notational clarity and for the controlled synthetic experiments, where a minimal compartmental structure helps isolate the effects of infection generation and population transport. The Lagrangian--Eulerian decomposition itself is not tied to SIR. For the real-world COVID-19 analyses, we extend the same operator construction to an SEIR backbone by introducing an exposed compartment while leaving the two graph pathways unchanged. For node $v$,
$$
X_v(t)
=
\left(
X_v^S(t),
X_v^E(t),
X_v^I(t),
X_v^R(t)
\right)^\top,
\qquad
P_v(t)=X_v^S(t)+X_v^E(t)+X_v^I(t)+X_v^R(t),
$$
with compartment-transition direction vectors
$$
b_{SE}=(-1,1,0,0)^\top,
\qquad
b_{EI}=(0,-1,1,0)^\top,
\qquad
b_{IR}=(0,0,-1,1)^\top.
$$
The Lagrangian messages now generate susceptible-to-exposed transitions, the self-loop message describes exposed-to-infectious and infectious-to-removed progression, and the Eulerian messages transport the complete SEIR state across regions. The resulting dynamics are
\begin{equation}
\begin{aligned}
\frac{\mathrm d X_v}{\mathrm d t}
=
(\mathcal{F}X)_v
={}&
\underbrace{
\left(
\sum_{u\in\mathcal{N}_{L}(v)\cup\{v\}}
\mathcal{F}_{uv}^{L}(t)X_u^I(t)
\right)b_{SE}
}_{\text{Lagrangian message passing}}
\\
&+
\underbrace{
\sigma X_v^E(t)b_{EI}
+
\gamma_vX_v^I(t)b_{IR}
}_{\text{self-loop message}}
+
\underbrace{
\sum_{u\in\mathcal{N}_{E}(v)\cup\{v\}}
\mathcal{F}_{uv}^{E}(t)X_u(t)
}_{\text{Eulerian message passing}} .
\end{aligned}
\label{eq:real_seir}
\end{equation}
Because the compartment-transition vectors sum to zero and the Eulerian operator remains zero-sum, the SEIR extension preserves the same population-conservation property as the SIR formulation. Dataset-specific values of the exposed-to-infectious transition rate $\sigma$ are given with the real-world data settings below.

\subsection{Learning structured epidemic dynamics}

Epidemic surveillance provides only partial observations of the underlying dynamics. In the observation model used here, reported incidence is identified with the $I\rightarrow R$ transition flow, whereas the remaining compartment states and the operator-level interaction strengths remain latent. LE-EpiGNODE therefore learns the continuous-time dynamics indirectly through this reporting channel embedded in the structured vector field.

Optimizing both operators solely from reported-case trajectories can lead to ambiguity because different combinations of infection generation and population transport may produce similar observations. We therefore use a two-stage learning procedure. The auxiliary mobility network is first pretrained using historical origin--destination observations. The epidemic model is then optimized from reported cases while the pretrained mobility-network parameters remain fixed. This procedure transfers mobility information into the Eulerian pathway before the two processes are coupled through the epidemic trajectory.

\subsubsection{OD-informed mobility pretraining}

Let $\vartheta$ denote the parameters of the auxiliary mobility network $\operatorname{MLP}_{\mathrm{mob}}$, and let $\theta$ denote the remaining trainable parameters of the epidemic model, including the contact-prior projection, Lagrangian infection network, Eulerian transport correction parameters, removal rates and initial epidemic state.

During the first stage, $\operatorname{MLP}_{\mathrm{mob}}$ is pretrained using the OD observations available within the training period. Because the latent infectious states are unavailable at this stage, the regional epidemic-activity input is constructed from observed reported cases:
$$
\bar I_{\mathrm{obs}}^k(t_n)
=
\frac{1}{N^k\Delta t_n}
\sum_{j=1}^{A}
Y_{\mathrm{obs},n,j}^{k},
\qquad
\Delta t_n=t_n-t_{n-1},
$$
where $Y_{\mathrm{obs},n,j}^{k}$ is the reported case count for age group $j$ in region $k$ over $[t_{n-1},t_n]$. The resulting population-normalized average reporting rate is used at observation times during mobility pretraining, with time units consistent with the epidemic dynamics. The mobility network is optimized by matching the predicted OD mobility value in Eq.~\eqref{eq:mobility_predictor} to the observed OD mobility value:
\begin{equation}
\mathcal{L}_{\mathrm{mob}}(\vartheta)
=
\frac{1}{|\mathcal{D}_{\mathrm{mob}}|}
\sum_{(t,l,k)\in\mathcal{D}_{\mathrm{mob}}}
\left[
\hat{\mathcal{M}}^{lk}(t)
-
\mathcal{M}^{lk}(t)
\right]^2,
\qquad
l\neq k,
\label{eq:mobility_loss}
\end{equation}
where $\mathcal{M}^{lk}(t)$ denotes the dataset-specific OD mobility observation from region $l$ to region $k$, and $\mathcal{D}_{\mathrm{mob}}$ is the set of observations used for pretraining. The optimized parameters are denoted by $\vartheta^{*}$ and remain fixed during subsequent epidemic-model training.

During epidemic training and forecasting, the regional epidemic activity supplied to the frozen mobility network is generated from the evolving infectious states:
$$
\bar I_{\mathrm{dyn}}^k(t)
=
\frac{1}{N^k}
\sum_{j=1}^{A}
\hat{\gamma}_j^k
\hat{X}_{(k,j)}^{I}(t),
$$
where $\hat{\gamma}_j^k$ is the learned removal rate and $\hat{X}_{(k,j)}^{I}(t)$ is the predicted infectious state. This instantaneous activity has the same units as the interval-averaged activity used during mobility pretraining. The epidemic-activity component of $z_{lk}(t)$ is therefore constructed from
$$
\bar I_{*}^k(t)
=
\begin{cases}
\bar I_{\mathrm{obs}}^k(t_n), & \text{during mobility pretraining, }t=t_n,\\
\bar I_{\mathrm{dyn}}^k(t), & \text{during epidemic training and forecasting}.
\end{cases}
$$
During the second stage, the activity components of $z_{lk}(t)$ in Eq.~\eqref{eq:mobility_predictor} are generated from $\bar I_{\mathrm{dyn}}^k(t)$, while the remaining dataset-specific spatial, temporal and regional features are constructed as described in Supplementary Methods, section 1.2. Although $\vartheta^{*}$ remains fixed, the predicted transport intensities continue to vary with the evolving epidemic states and time. The pretrained network therefore acts as an OD-informed mobility prior rather than a fixed sequence of future mobility observations.

\subsubsection{Epidemic training objective}

Reported cases provide an indirect observation of the latent epidemic dynamics through the $I\rightarrow R$ transition flow. With time measured in days, the predicted number of newly reported cases for age group $j$ in region $k$ over $[t_{n-1},t_n]$ is approximated by
$$
\hat{Y}_{n,j}^{k}
\approx
\hat{\gamma}_j^k
\hat{X}_{(k,j)}^{I}(t_n)
\Delta t_n,
\qquad
\Delta t_n=t_n-t_{n-1}=1,
$$
where $\hat{\gamma}_j^k$ is the learned removal rate and $\hat{X}_{(k,j)}^{I}(t_n)$ is the predicted infectious state at $t_n$.

Given the observed number of newly reported cases $Y_{\mathrm{obs},n,j}^{k}$, the epidemic observation loss is
\begin{equation}
\mathcal{L}_{\mathrm{epi}}(\theta)
=
\frac{1}{|\mathcal{D}_{\mathrm{epi}}|}
\sum_{(n,j,k)\in\mathcal{D}_{\mathrm{epi}}}
\left(
\hat{Y}_{n,j}^{k}
-
Y_{\mathrm{obs},n,j}^{k}
\right)^2,
\label{eq:epidemic_loss}
\end{equation}
where $\mathcal{D}_{\mathrm{epi}}$ denotes the set of epidemic observations used for training.

The initial epidemic state is estimated jointly with the model parameters. To place additional weight on consistency with the first observation time, we define
$$
\mathcal{L}_{\mathrm{ic}}(\theta)
=
\frac{1}{|\mathcal{V}|}
\sum_{(k,j)\in\mathcal{V}}
\left(
\hat{Y}_{1,j}^{k}
-
Y_{\mathrm{obs},1,j}^{k}
\right)^2.
$$
The epidemic model is optimized according to
\begin{equation}
\min_{\theta}
\mathcal{L}_{\mathrm{main}}(\theta)
=
\mathcal{L}_{\mathrm{epi}}(\theta)
+
\lambda_{\mathrm{ic}}
\mathcal{L}_{\mathrm{ic}}(\theta),
\qquad
\vartheta=\vartheta^{*},
\label{eq:main_loss}
\end{equation}
where $\lambda_{\mathrm{ic}}$ controls the contribution of the initial-observation term and $\vartheta^{*}$ denotes the fixed parameters of the pretrained mobility network. The compartment-transition directions and population-conservation condition are encoded directly in $\mathcal{F}$ and therefore do not require additional penalty terms.

\subsubsection{Optimization and forecasting}

During training, the predicted epidemic trajectory is obtained by numerically integrating Eq.~\eqref{eq:dual_view_dynamics} for the SIR formulation or Eq.~\eqref{eq:real_seir} for the real-world SEIR formulation. Gradients of the objective in Eq.~\eqref{eq:main_loss} are propagated through the numerical solution to optimize the epidemic parameters and initial state jointly, while the pretrained mobility parameters $\vartheta^{*}$ remain fixed.

During forecasting, the learned dynamics are integrated forward from the forecast origin using the evolving predicted state. The Lagrangian operator updates infection-generation interactions according to the epidemic state, age-contact prior and time. The Eulerian operator updates transport intensities through $z_{lk}(t)$, whose epidemic-activity components are generated internally from $\bar I_{\mathrm{dyn}}^k(t)$ during forecasting. LE-EpiGNODE therefore generates future epidemic trajectories without requiring future reported cases or future OD observations.
\subsection{Data}

We use controlled synthetic datasets and real-world COVID-19 datasets for complementary purposes. The synthetic datasets provide known generating mechanisms and ground-truth operators for evaluating forecasting performance, structural contributions and dynamical fidelity under controlled conditions. The real-world datasets test forecasting from partial observations in settings involving age structure, population mobility or both.

\subsubsection{Synthetic datasets}

Synthetic data are generated by simulating the SIR-based Lagrangian--Eulerian system described above from day $0$ to day $200$, with daily sampling. We consider three structural settings. The Lagrangian setting contains one region and four age groups, isolating age-dependent infection generation without interregional transport. The Eulerian setting contains four regions and one age group, isolating mobility-driven spatial coupling without age stratification. The multiscale setting contains four regions and four age groups and couples both processes within the same epidemic state space.

Within each setting, we vary the generating functions while keeping population sizes, initial conditions, basic epidemiological parameters and observation-noise rules fixed. The four Lagrangian mechanisms---Baseline Mass-Action Contact Mechanism (BMCM), Time-Decaying Contact Mechanism (TDCM), Infection-Avoidant Contact Mechanism (IACM) and Nonlinear Mass-Action Contact Mechanism (NMCM)---span baseline mass-action, explicit temporal modulation, infection-dependent feedback and nonlinear state dependence. The two Eulerian mechanisms---Baseline Risk-Aware Gravity Mobility (BRGM) and Controlled Risk-Aware Gravity Mobility (CRGM)---represent epidemic-risk-aware transport with and without additional explicit temporal control. Their eight pairwise combinations define the multiscale systems, yielding 14 controlled generation conditions in total. This design tests whether a single learned formulation can extrapolate changing process dynamics rather than fit only a fixed mechanistic form. Complete mechanism definitions and generation settings are provided in Supplementary Methods, section 1.1, and Supplementary Tables~1--3.

For each condition, the dataset contains daily epidemic observations and the corresponding contact structure, with directed OD mobility observations additionally retained when transport is present. Gaussian observation noise is added to both epidemic and OD observations, with standard deviation set to $5\%$ of the corresponding trajectory maximum. Because the generating equations are known, latent states, instantaneous Lagrangian and Eulerian operators, and their state fluxes are retained as ground truth for dynamical-recovery analyses. Complete population and contact structures, mechanism-specific parameters and noise settings are provided in Supplementary Methods, section 1.1, and Supplementary Tables~1--3.

\subsubsection{Real-world COVID-19 datasets}

Real-world validation uses three daily COVID-19 datasets representing complementary settings: age-structured transmission in Ontario, mobility-driven spatial dynamics in Hubei, and coupled age--space dynamics in Spain. Unlike the synthetic datasets, their latent states and generating operators are unobserved; we therefore use the real-world datasets to evaluate forecasting from partial observations using the SEIR extension in Eq.~\eqref{eq:real_seir}. Supplementary Table~4 summarizes the epidemic series, process-aligned information and evaluation settings.

Ontario uses age-stratified reported cases from the Ontario Data Catalogue covering 28 November 2021 to 27 February 2022, represented as one region with four age groups: $[0,20)$, $[20,40)$, $[40,60)$ and $\geq60$. An Ontario-specific contact matrix informs the Lagrangian pathway~\cite{mistry2021mixing}, and no interregional transport is included. The exposed-to-infectious rate is set to $\sigma=(3.61\,\mathrm{days})^{-1}$, using an Omicron incubation-period estimate as a modeling approximation~\cite{galmiche2023incubation}.

Hubei uses daily reported cases for Wuhan, Xiaogan and Huanggang from 16 January to 25 February 2020, obtained from the DXY-COVID-19-Data and Wuhan-2019-nCoV repositories, without age stratification. Daily directed OD indices from Baidu mobility data hosted by Harvard Dataverse are used to pretrain the mobility network, while city populations and distances define the gravity representation. The exposed-to-infectious rate is set to $\sigma=(5.2\,\mathrm{days})^{-1}$, using an early-pandemic incubation-period estimate as a modeling approximation~\cite{li2020wuhan}.

Spain uses daily reported cases from the Spanish National Centre for Epidemiology covering 1 March to 21 June 2020, aggregated from NUTS-1 regions into four spatial nodes (R1: ES1+ES2; R2: ES3+ES4; R3: ES5; R4: ES6) and three age groups: $[0,50)$, $[50,70)$ and $\geq70$. Population-adjusted synthetic contact matrices inform the Lagrangian pathway~\cite{prem2021contact}. Daily OD mobility observations obtained from Zenodo, together with regional populations and spatial separation, inform the Eulerian pathway. The exposed-to-infectious rate is set to $\sigma=(5.2\,\mathrm{days})^{-1}$, using an early-pandemic incubation-period estimate as a modeling approximation~\cite{li2020wuhan}.

All models use identical training periods, forecast origins and evaluation horizons within each dataset. Dataset-specific temporal and regional inputs to the auxiliary mobility predictor are described in Supplementary Methods, section 1.2, and summarized in Supplementary Table~5.
\subsection{Evaluation design}

The evaluation addresses three questions: whether LE-EpiGNODE improves long-horizon forecasting, which structural components and process-aligned priors contribute to these gains, and whether the learned operators reproduce known generating dynamics in controlled systems. Synthetic experiments use pre-peak observations for training and evaluate forecast horizons of 1, 4, 12 and 20 weeks. For the real-world datasets, training uses 25 days for Ontario, 15 days for Hubei and 20 days for Spain, with forecast horizons of 7/14/35/67, 7/14/26 and 7/14/49/93 days, respectively (Supplementary Table~4). Within each setting, all models use the same epidemic observations, forecast origins and evaluation horizons.

 \subsubsection{Baseline models}

We compared LE-EpiGNODE with four groups of baselines representing different approaches to epidemic forecasting. Mechanistic baselines include SIR~\cite{kermack1927}, MepoSIR~\cite{colizza2007} and the mobility-coupled metapopulation SEIR model (MCM-SEIR). General forecasting baselines include LSTM~\cite{hochreiter1997}, GRU~\cite{cho2014}, Neural ODE~\cite{chen2018}, and the spatiotemporal graph models STGCN~\cite{yu2018stgcn}, DCRNN~\cite{li2018dcrnn}, AGCRN~\cite{bai2020agcrn}, Graph WaveNet~\cite{wu2019graphwavenet}, MTGNN~\cite{wu2020mtgnn} and Cola-GNN~\cite{deng2020colagnn}. Epidemic-specific graph baselines include EpiGNN~\cite{xie2023epignn}, Epi-Cola-GNN~\cite{liu2023epicola} and CovidGNN~\cite{kapoor2020covidgnn}, while hybrid mechanistic-learning baselines include STAN~\cite{gao2021stan}, EpiGCN (adapted)~\cite{han2023epigcn}, EARTH~\cite{Wan_EpiODE_ICML25}, MepoGNN~\cite{cao2023mepognn}, M-Graphormer~\cite{wang2025multi} and EINN~\cite{rodriguez2023}.

Synthetic experiments compare mechanistic and general forecasting baselines under controlled generating dynamics, while real-world experiments additionally include epidemic-specific graph and hybrid models. Model architectures, input configurations, hyperparameters and implementation details, including the continuous-time adaptation of EpiGCN, are provided in Supplementary Methods, section 1.3, and Supplementary Tables~6--8.

\subsubsection{Ablation settings}

We consider five ablations to assess the contributions of the Lagrangian--Eulerian structure and the priors aligned with each pathway. \textit{w/o LE structure} replaces the structured formulation with a general Graph Neural ODE vector field. \textit{w/o $\mathcal{G}_{L}$} and \textit{w/o $\mathcal{G}_{E}$} remove the Lagrangian infection-generation and Eulerian population-transport pathways, respectively.

To evaluate the process-aligned priors, \textit{w/o $e_{uv}^{L}$} removes the age-contact representation while retaining the remaining Lagrangian architecture and state inputs. \textit{w/o $\operatorname{MLP}_{\mathrm{mob}}$ pretraining} retains the mobility network but replaces OD-supervised pretraining and parameter freezing with random initialization and joint optimization under the epidemic observation objective.

All variants use the same epidemic observations, training periods, forecast origins, evaluation horizons and epidemic-training settings as the complete model, except for the specified ablation.

\subsubsection{Evaluation metrics}

We evaluate LE-EpiGNODE from two complementary perspectives: epidemic forecasting and dynamical consistency. Forecasting performance is measured by mean absolute error (MAE) and weighted mean absolute percentage error (WMAPE):
$$
\operatorname{MAE}
=
\frac{1}{Q}
\sum_{q=1}^{Q}
\left|
\hat{y}_q-y_q
\right|,
\qquad
\operatorname{WMAPE}
=
\frac{
\sum_{q=1}^{Q}
\left|
\hat{y}_q-y_q
\right|
}{
\sum_{q=1}^{Q}|y_q|
}
\times100\%,
$$
where $y_q$ and $\hat{y}_q$ are the observed and predicted values, respectively, and $Q$ is the number of evaluated observations. MAE measures absolute prediction error, whereas WMAPE normalizes the total absolute error by the magnitude of the observed trajectory.

For synthetic systems, known generating dynamics allow direct evaluation of the instantaneous operators $\mathcal{F}^{L}$ and $\mathcal{F}^{E}$ and state fluxes $\Phi$. WMAPE measures magnitude error, while the Pearson correlation coefficient (PCC) measures agreement in variation:
$$
\operatorname{PCC}(a,\hat{a})
=
\frac{
\sum_{q=1}^{Q}
(a_q-\bar a)
(\hat{a}_q-\overline{\hat{a}})
}{
\sqrt{
\sum_{q=1}^{Q}(a_q-\bar a)^2
}
\sqrt{
\sum_{q=1}^{Q}(\hat{a}_q-\overline{\hat{a}})^2
}
},
$$
where $a_q$ and $\hat{a}_q$ are corresponding true and inferred operator or flux entries across evaluated edges and time points. Lower WMAPE values and PCC values closer to 1 indicate better agreement in magnitude and variation, respectively.

To assess cumulative interaction structure, we define the integrated Lagrangian interaction flux (LIF) and integrated Eulerian interaction flux (EIF) over $[t_0,t_T]$:
$$
\operatorname{LIF}_{uv}
=
\int_{t_0}^{t_T}
\Phi_{uv}^{L}(t)
\,\mathrm dt,
\qquad
\operatorname{EIF}_{uv}
=
\int_{t_0}^{t_T}
\Phi_{uv}^{E}(t)
\,\mathrm dt.
$$
Agreement between true and inferred integrated structures is measured by the Spearman rank correlation coefficient (SRCC):
$$
\operatorname{SRCC}(a,\hat{a})
=
\operatorname{PCC}
\left(
\operatorname{rank}(a),
\operatorname{rank}(\hat{a})
\right).
$$
LIF-SRCC is computed over all entries of the integrated Lagrangian flux matrix, whereas EIF-SRCC excludes diagonal entries determined by population conservation. SRCC values closer to 1 indicate stronger agreement in the ranking of cumulative interactions.

For real-world datasets, operator and flux ground truth is unavailable, so integrated fluxes are instead aggregated and normalized for process-level interpretation:
\[
A_{ij}^{L,\mathrm{row}}
=
\frac{\sum_k \operatorname{LIF}_{(k,i),(k,j)}}
{\sum_{j'}\sum_k \operatorname{LIF}_{(k,i),(k,j')}},
\qquad
A_{ij}^{L,\mathrm{col}}
=
\frac{\sum_k \operatorname{LIF}_{(k,i),(k,j)}}
{\sum_{i'}\sum_k \operatorname{LIF}_{(k,i'),(k,j)}},
\]
and
\[
A_{lk}^{E,\mathrm{row}}
=
\frac{\sum_j \operatorname{EIF}_{(l,j),(k,j)}}
{\sum_{k'\neq l}\sum_j \operatorname{EIF}_{(l,j),(k',j)}},
\qquad
A_{lk}^{E,\mathrm{col}}
=
\frac{\sum_j \operatorname{EIF}_{(l,j),(k,j)}}
{\sum_{l'\neq k}\sum_j \operatorname{EIF}_{(l',j),(k,j)}},
\qquad l\neq k.
\]
LIF aggregates infection-generation fluxes over regions to characterize age-to-age interactions, whereas EIF aggregates infectious-state transport fluxes over age groups to characterize region-to-region interactions. Row normalization gives the relative target distribution from each source, and column normalization gives the relative source composition of each target. Diagonal conservation terms are excluded from EIF.
\section*{Data availability}

The original empirical data sources used in this study are publicly accessible. COVID-19 case data for Ontario were obtained from the Ontario Data Catalogue. Case data for Hubei were obtained from the DXY-COVID-19-Data and Wuhan-2019-nCoV repositories. Case data for Spain were obtained from the Spanish National Centre for Epidemiology. The corresponding OD mobility data for Hubei and Spain were obtained from Harvard Dataverse and Zenodo, respectively. The synthetic datasets were generated using the mechanisms summarized in Methods, with the complete generation equations and parameter settings reported in Supplementary Methods, section 1.1, and Supplementary Tables 1--3.

\clearpage
\section*{Extended Data}
\setcounter{extendeddatafigure}{0}
\setcounter{extendeddatatable}{0}

\begin{table*}[!htpb]
\centering
\edtablecaption
{Forecasting performance on synthetic scenarios.}
{Each cell reports MAE / WMAPE (\%); the best and second-best results are shown in bold and underlined, respectively.}
\label{tab:ed-synthetic-forecasting}
\scriptsize
\renewcommand{\arraystretch}{1.15}
\setlength{\tabcolsep}{5pt}
\begin{tabular}{lcccc}
\toprule
\textbf{Model} & \textbf{1 week} & \textbf{4 weeks} & \textbf{12 weeks} & \textbf{20 weeks} \\
\midrule
\multicolumn{5}{c}{\textit{Lagrangian}} \\
\midrule
LE-EpiGNODE & \textbf{44.38 / 5.67} & \textbf{51.09 / 5.84} & \textbf{58.66 / 8.45} & \underline{63.30 / 12.45} \\
SIR & 270.78 / 34.57 & 299.29 / 34.21 & 238.84 / 34.42 & 186.39 / 36.68 \\
MepoSIR & 72.70 / 9.28 & 73.16 / 8.36 & \underline{63.80 / 9.19} & \textbf{62.04 / 12.21} \\
LSTM & 70.16 / 8.96 & 119.28 / 13.63 & 220.36 / 31.75 & 373.05 / 73.40 \\
GRU & 50.56 / 6.45 & 87.05 / 9.95 & 215.85 / 31.10 & 383.14 / 75.39 \\
Neural ODE & 87.01 / 11.11 & 112.97 / 12.91 & 226.10 / 32.58 & 385.50 / 75.85 \\
STGCN & 64.11 / 8.18 & 102.69 / 11.74 & 204.36 / 29.45 & 364.85 / 71.79 \\
DCRNN & \underline{49.47 / 6.31} & \underline{69.23 / 7.91} & 201.67 / 29.06 & 374.82 / 73.75 \\
AGCRN & 64.15 / 8.19 & 92.19 / 10.54 & 210.09 / 30.27 & 376.15 / 74.01 \\
GraphWaveNet & 73.80 / 9.42 & 126.12 / 14.41 & 196.60 / 28.33 & 343.10 / 67.51 \\
MTGNN & 64.24 / 8.20 & 100.21 / 11.45 & 217.10 / 31.28 & 380.89 / 74.95 \\
\midrule
\multicolumn{5}{c}{\textit{Eulerian}} \\
\midrule
LE-EpiGNODE & \textbf{152.42 / 13.41} & \textbf{241.97 / 16.44} & \textbf{171.32 / 18.32} & \textbf{123.12 / 21.11} \\
SIR & 470.21 / 41.36 & 716.53 / 48.67 & \underline{543.40 / 58.12} & \underline{359.46 / 61.63} \\
MepoSIR & 647.62 / 56.96 & 739.74 / 50.25 & 567.70 / 60.72 & 365.66 / 62.69 \\
LSTM & 382.45 / 33.64 & 766.90 / 52.09 & 726.86 / 77.74 & 686.99 / 117.78 \\
GRU & 362.86 / 31.91 & 715.99 / 48.64 & 753.75 / 80.62 & 751.00 / 128.75 \\
Neural ODE & 424.93 / 37.37 & 802.98 / 54.54 & 754.31 / 80.68 & 702.46 / 120.43 \\
STGCN & 438.29 / 38.55 & 808.82 / 54.94 & 756.45 / 80.90 & 707.87 / 121.36 \\
DCRNN & 338.83 / 29.80 & 729.31 / 49.54 & 692.15 / 74.03 & 640.38 / 109.79 \\
AGCRN & 323.60 / 28.46 & \underline{661.11 / 44.91} & 703.84 / 75.28 & 605.89 / 103.88 \\
GraphWaveNet & \underline{304.94 / 26.82} & 665.80 / 45.23 & 742.35 / 79.40 & 770.63 / 132.12 \\
MTGNN & 342.14 / 30.09 & 707.47 / 48.06 & 691.90 / 74.00 & 709.36 / 121.62 \\
\midrule
\multicolumn{5}{c}{\textit{Multiscale}} \\
\midrule
LE-EpiGNODE & 123.68 / 17.16 & \textbf{136.39 / 13.50} & \textbf{113.54 / 10.65} & \textbf{110.01 / 13.36} \\
SIR & 417.17 / 57.88 & 576.00 / 57.02 & 576.04 / 54.02 & 445.71 / 54.14 \\
MepoSIR & 218.34 / 30.29 & 283.60 / 28.08 & \underline{245.70 / 23.04} & \underline{210.09 / 25.52} \\
LSTM & 121.39 / 16.84 & 351.85 / 34.83 & 477.24 / 44.75 & 450.43 / 54.72 \\
GRU & 130.92 / 18.16 & 365.54 / 36.19 & 487.18 / 45.68 & 456.17 / 55.41 \\
Neural ODE & 183.23 / 25.42 & 398.87 / 39.49 & 511.79 / 47.99 & 494.10 / 60.02 \\
STGCN & 109.80 / 15.23 & 297.87 / 29.49 & 445.47 / 41.77 & 466.88 / 56.71 \\
DCRNN & \textbf{95.33 / 13.23} & \underline{274.44 / 27.17} & 437.31 / 41.01 & 466.42 / 56.66 \\
AGCRN & 126.42 / 17.54 & 301.52 / 29.85 & 421.37 / 39.51 & 448.22 / 54.45 \\
GraphWaveNet & 117.47 / 16.30 & 330.10 / 32.68 & 460.83 / 43.21 & 454.27 / 55.18 \\
MTGNN & \underline{108.14 / 15.00} & 285.70 / 28.28 & 423.83 / 39.74 & 446.42 / 54.23 \\
\bottomrule
\end{tabular}
\end{table*}

\begin{table*}[!htpb]
\centering
\edtablecaption
{Forecasting performance on the real-world COVID-19 datasets.}
{Each cell reports MAE / WMAPE (\%); the best and second-best results are shown in bold and underlined, respectively.}
\label{tab:ed-real-forecasting}
\scriptsize
\renewcommand{\arraystretch}{1.15}
\setlength{\tabcolsep}{3pt}
\begin{tabular}{lcccc}
\toprule

\multicolumn{5}{c}{\textit{Ontario}} \\
\midrule
\textbf{Model} & \textbf{7 days} & \textbf{14 days} & \textbf{35 days} & \textbf{67 days} \\
\midrule
LE-EpiGNODE
& \textbf{128.57 / 3.43}
& \textbf{165.71 / 4.49}
& \textbf{190.60 / 7.80}
& \textbf{248.66 / 16.20} \\
MCM-SEIR
& 364.02 / 9.72
& \underline{449.45 / 12.17}
& \underline{447.98 / 18.33}
& \underline{333.83 / 21.76} \\
DCRNN
& 697.37 / 18.63
& 803.41 / 21.76
& 594.27 / 24.32
& 415.31 / 27.06 \\
AGCRN
& 1255.42 / 33.53
& 1546.00 / 41.87
& 932.27 / 38.15
& 584.65 / 38.10 \\
MTGNN
& 708.40 / 18.92
& 826.26 / 22.38
& 504.55 / 20.65
& 365.74 / 23.83 \\
EpiGNN
& 609.43 / 16.28
& 751.34 / 20.35
& 581.71 / 23.80
& 530.21 / 34.55 \\
Cola-GNN
& 870.40 / 23.25
& 1053.45 / 28.53
& 659.56 / 26.99
& 418.94 / 27.30 \\
CovidGNN
& 1349.77 / 36.05
& 1601.26 / 43.37
& 995.45 / 40.73
& 608.55 / 39.66 \\
MepoGNN
& 547.73 / 14.63
& 576.04 / 15.60
& 515.62 / 21.10
& 480.28 / 31.30 \\
M-Graphormer
& \underline{277.70 / 7.42}
& 493.73 / 13.37
& 608.23 / 24.89
& 791.57 / 51.58 \\
EpiGCN
& 870.47 / 23.25
& 969.13 / 26.25
& 694.34 / 28.41
& 540.18 / 35.20 \\
STAN
& 882.64 / 23.58
& 989.26 / 26.79
& 738.46 / 30.22
& 623.62 / 40.64 \\
EINN
& 552.13 / 14.75
& 637.52 / 17.27
& 570.03 / 23.32
& 374.45 / 24.40 \\
Epi-Cola-GNN
& 874.91 / 23.37
& 861.65 / 23.34
& 608.58 / 24.90
& 377.05 / 24.57 \\
EARTH
& 503.39 / 13.45
& 610.27 / 16.53
& 620.73 / 25.40
& 416.66 / 27.15 \\

\midrule
\multicolumn{5}{c}{\textit{Hubei}} \\
\midrule
\textbf{Model} & \textbf{7 days} & \textbf{14 days} & \textbf{26 days} & \\
\midrule
LE-EpiGNODE
& \textbf{88.04 / 17.15}
& \textbf{58.25 / 8.07}
& \textbf{100.34 / 15.87}
& -- \\
MCM-SEIR
& 120.45 / 23.46
& 197.96 / 27.43
& 488.72 / 77.29
& -- \\
DCRNN
& 140.26 / 27.32
& 123.22 / 17.07
& 288.36 / 45.60
& -- \\
AGCRN
& \underline{98.68 / 19.22}
& 107.81 / 14.94
& 291.94 / 46.17
& -- \\
MTGNN
& 146.23 / 28.48
& 131.28 / 18.19
& 326.15 / 51.58
& -- \\
EpiGNN
& 146.70 / 28.57
& 124.16 / 17.20
& 280.76 / 44.40
& -- \\
Cola-GNN
& 126.63 / 24.66
& 116.85 / 16.19
& 256.29 / 40.53
& -- \\
CovidGNN
& 138.60 / 26.99
& 122.71 / 17.00
& 250.55 / 39.62
& -- \\
MepoGNN
& 105.72 / 20.59
& \underline{104.00 / 14.41}
& \underline{167.27 / 26.45}
& -- \\
M-Graphormer
& 130.34 / 25.39
& 188.64 / 26.14
& 244.12 / 38.61
& -- \\
EpiGCN
& 106.79 / 20.80
& 168.90 / 23.40
& 302.04 / 47.76
& -- \\
STAN
& 185.62 / 36.15
& 217.64 / 30.16
& 194.13 / 30.70
& -- \\
EINN
& 103.84 / 20.22
& 186.81 / 25.88
& 280.60 / 44.37
& -- \\
Epi-Cola-GNN
& 105.28 / 20.50
& 178.32 / 24.71
& 316.56 / 50.06
& -- \\
EARTH
& 144.79 / 28.20
& 198.63 / 27.52
& 391.77 / 61.95
& -- \\

\midrule
\multicolumn{5}{c}{\textit{Spain}} \\
\midrule
\textbf{Model} & \textbf{7 days} & \textbf{14 days} & \textbf{49 days} & \textbf{93 days} \\
\midrule
LE-EpiGNODE
& 54.78 / 9.61
& 54.63 / 9.66
& \textbf{43.89 / 14.09}
& \textbf{37.60 / 20.43} \\
MCM-SEIR
& 92.71 / 16.27
& 104.42 / 18.47
& 71.43 / 22.93
& \underline{45.64 / 24.81} \\
DCRNN
& 73.10 / 12.83
& 84.93 / 15.02
& 62.21 / 19.97
& 98.84 / 53.72 \\
AGCRN
& 72.84 / 12.78
& 96.90 / 17.14
& 120.47 / 38.67
& 155.10 / 84.30 \\
MTGNN
& 56.83 / 9.97
& 75.41 / 13.34
& 72.10 / 23.15
& 82.27 / 44.71 \\
EpiGNN
& 67.92 / 11.92
& 100.13 / 17.71
& 115.12 / 36.96
& 134.18 / 72.93 \\
Cola-GNN
& 96.90 / 17.00
& 117.50 / 20.78
& 101.28 / 32.51
& 114.12 / 62.03 \\
CovidGNN
& 85.78 / 15.05
& 116.24 / 20.56
& 137.93 / 44.28
& 190.14 / 103.34 \\
MepoGNN
& \underline{39.89 / 7.00}
& \underline{46.82 / 8.28}
& \underline{54.92 / 17.63}
& 53.09 / 28.86 \\
M-Graphormer
& 146.31 / 25.67
& 146.28 / 25.87
& 92.73 / 29.77
& 83.40 / 45.33 \\
EpiGCN
& 40.66 / 7.13
& 55.71 / 9.85
& 84.62 / 27.17
& 64.01 / 34.79 \\
STAN
& 70.06 / 12.29
& 86.56 / 15.31
& 144.56 / 46.41
& 221.38 / 120.32 \\
EINN
& \textbf{19.70 / 3.46}
& \textbf{34.46 / 6.09}
& 68.97 / 22.14
& 57.62 / 31.32 \\
Epi-Cola-GNN
& 103.15 / 18.10
& 99.67 / 17.63
& 76.47 / 24.55
& 63.01 / 34.24 \\
EARTH
& 107.51 / 18.86
& 119.49 / 21.13
& 155.51 / 49.92
& 181.46 / 98.63 \\
\bottomrule
\end{tabular}
\end{table*}

\clearpage
\begin{table*}[!htbp]
\centering

\edtablecaption
{OD mobility fitting and forecasting errors.}
{Each entry reports MAE / WMAPE (\%) for the directed OD mobility values in Hubei
and Spain. Errors are computed on each dataset's original mobility scale; MAE
values are not directly comparable across datasets.}

\label{tab:ed-od-errors}

\footnotesize
\renewcommand{\arraystretch}{1.15}
\setlength{\tabcolsep}{6pt}

\begin{tabular}{lcccc}
\toprule
\multicolumn{5}{c}{\textit{Hubei}} \\
\midrule
\textbf{Output} & \textbf{Training} & \textbf{7 days} & \textbf{14 days} & \textbf{26 days} \\
\midrule
OD mobility & 0.0033 / 0.99 & 0.0120 / 28.75 & 0.0138 / 33.80 & 0.0161 / 40.15 \\
\bottomrule
\end{tabular}

\vspace{0.8em}

\begin{tabular}{lccccc}
\toprule
\multicolumn{6}{c}{\textit{Spain}} \\
\midrule
\textbf{Output} & \textbf{Training} & \textbf{7 days} & \textbf{14 days}
& \textbf{49 days} & \textbf{93 days} \\
\midrule
OD mobility & 7303.85 / 12.25 & 4270.26 / 16.67 & 5753.13 / 24.24
& 6249.38 / 24.21 & 9649.42 / 29.06 \\
\bottomrule
\end{tabular}

\end{table*}

\begin{table*}[!htbp]
\centering

\edtablecaption
{Dynamical fidelity on synthetic data.}
{$\mathcal F^{b}$-WMAPE/PCC and $\Phi$-WMAPE/PCC evaluate the instantaneous process-specific operator and state flux, respectively, where $b\in\{L,E\}$ denotes the branch listed in each row. IF-SRCC denotes LIF-SRCC for the Lagrangian branch and EIF-SRCC for the Eulerian branch. Lower WMAPE and higher PCC or SRCC indicate stronger agreement with the known generating dynamics.}

\label{tab:ed-dynamics-consistency}

\scriptsize
\renewcommand{\arraystretch}{1.50}
\setlength{\tabcolsep}{4pt}

\begin{adjustbox}{max width=\textwidth}
\begin{tabular}{lllccccc}
\toprule
\textbf{Scenario}
& \textbf{Branch}
& \textbf{Variant}
& \textbf{$\mathcal F^{b}$-WMAPE (\%)}
& \textbf{$\Phi$-WMAPE (\%)}
& \textbf{$\mathcal F^{b}$-PCC}
& \textbf{$\Phi$-PCC}
& \textbf{IF-SRCC} \\
\midrule

\multirow{2}{*}{Lagrangian}
& \multirow{2}{*}{$L$}
& LE-EpiGNODE
& 19.79
& 25.89
& 0.980
& 0.959
& 0.971 \\

&
& w/o $e_{uv}^{L}$
& 48.44
& 196.68
& 0.849
& 0.074
& 0.971 \\

\midrule

\multirow{2}{*}{Eulerian}
& \multirow{2}{*}{$E$}
& LE-EpiGNODE
& 6.93
& 22.58
& 0.996
& 0.981
& 0.923 \\

&
& w/o $\operatorname{MLP}_{\mathrm{mob}}$ pretraining
& 53.49
& 79.37
& 0.683
& 0.664
& 0.448 \\

\midrule

\multirow{6}{*}{Multiscale}
& \multirow{3}{*}{$L$}
& LE-EpiGNODE
& 26.45
& 34.29
& 0.936
& 0.920
& 0.835 \\

&
& w/o $e_{uv}^{L}$
& 30.78
& 42.36
& 0.931
& 0.872
& 0.851 \\

&
& w/o $\operatorname{MLP}_{\mathrm{mob}}$ pretraining
& 39.27
& 54.06
& 0.886
& 0.778
& 0.849 \\

\cmidrule(lr){2-8}

&
\multirow{3}{*}{$E$}
& LE-EpiGNODE
& 5.66
& 16.15
& 0.999
& 0.990
& 0.998 \\

&
& w/o $e_{uv}^{L}$
& 5.80
& 21.70
& 0.999
& 0.980
& 0.997 \\

&
& w/o $\operatorname{MLP}_{\mathrm{mob}}$ pretraining
& 6.65
& 25.42
& 0.998
& 0.980
& 0.997 \\

\bottomrule
\end{tabular}
\end{adjustbox}
\end{table*}

\clearpage
\begin{figure*}[p]
\centering

\includegraphics[
    width=\textwidth,
    height=\textheight,
    keepaspectratio
]{figures/ed-multiscale-forecast.pdf}

\edcaption
{Fitting and forecasting trajectories across all coupled multiscale synthetic mechanisms.}
{\textbf{(a)}, Multiscale scenarios combining the four Lagrangian infection-generation mechanisms (BMCM, TDCM, IACM and NMCM) with BRGM transport.
\textbf{(b)}, Corresponding multiscale scenarios combining the same four Lagrangian mechanisms with CRGM transport.
The vertical dashed line marks the forecast origin, and the gray-shaded region denotes the out-of-sample forecasting period.
Colored envelopes show post-hoc normal-reference residual bands centered on the model predictions.
All models are trained using only observations before the forecast origin.}

\label{fig:ed-multiscale-forecast}

\end{figure*}

\clearpage
\begin{figure*}[p]
\centering

\includegraphics[
    width=\textwidth,
    height=\textheight,
    keepaspectratio
]{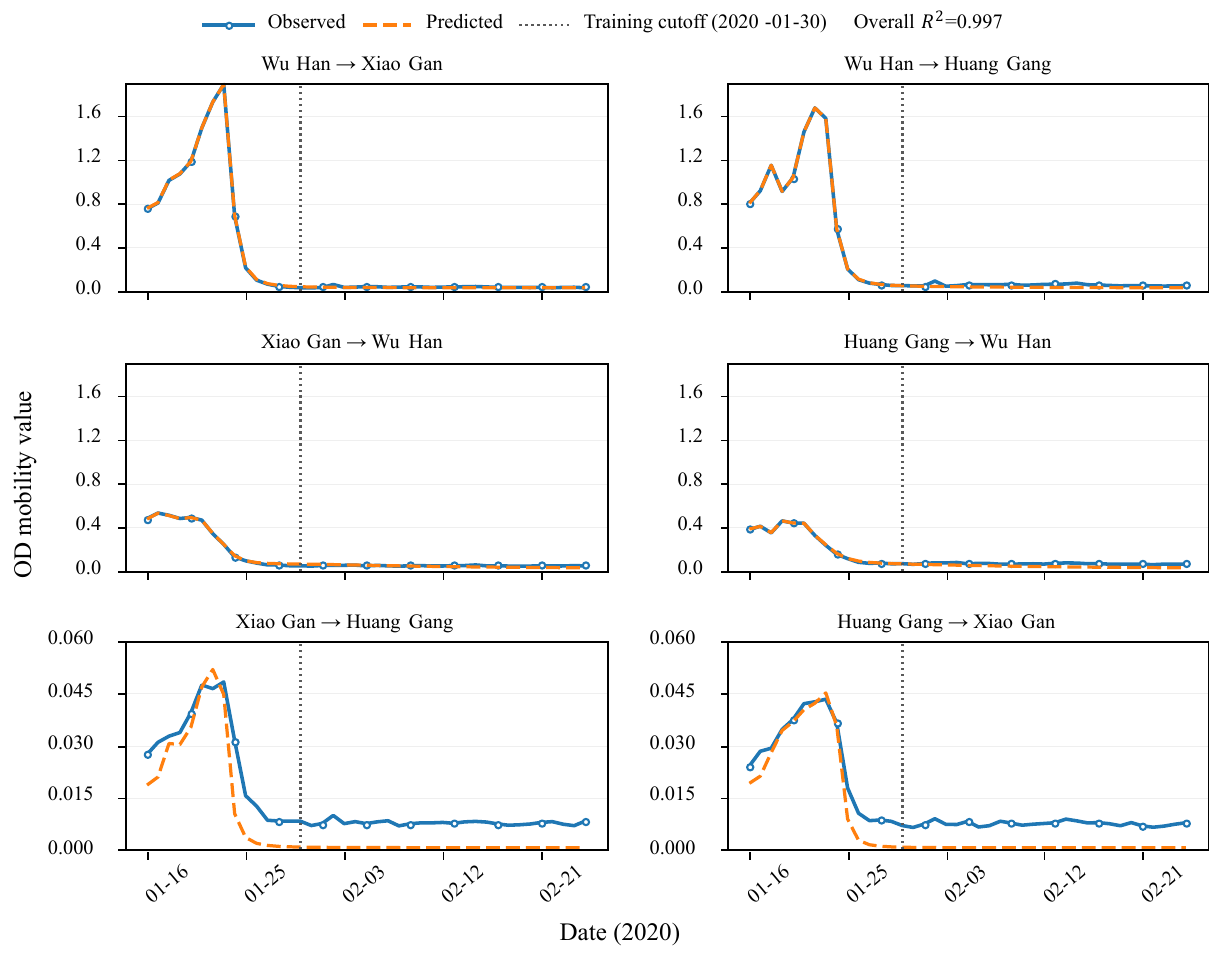}

\edcaption
{OD mobility extrapolation in Hubei.}
{Observed and predicted trajectories are shown for all directed OD edges. The vertical dashed line marks the forecast origin. The auxiliary mobility predictor uses only OD observations available before the forecast origin. The overall $R^2$ displayed in the plot summarizes the complete fitting and forecasting interval; forecasting MAE and WMAPE are reported separately in Extended Data Table~\ref{tab:ed-od-errors}.}
\label{fig:ed-hubei-mobility}

\end{figure*}

\clearpage
\begin{figure*}[p]
\centering

\includegraphics[
    width=\textwidth,
    height=\textheight,
    keepaspectratio
]{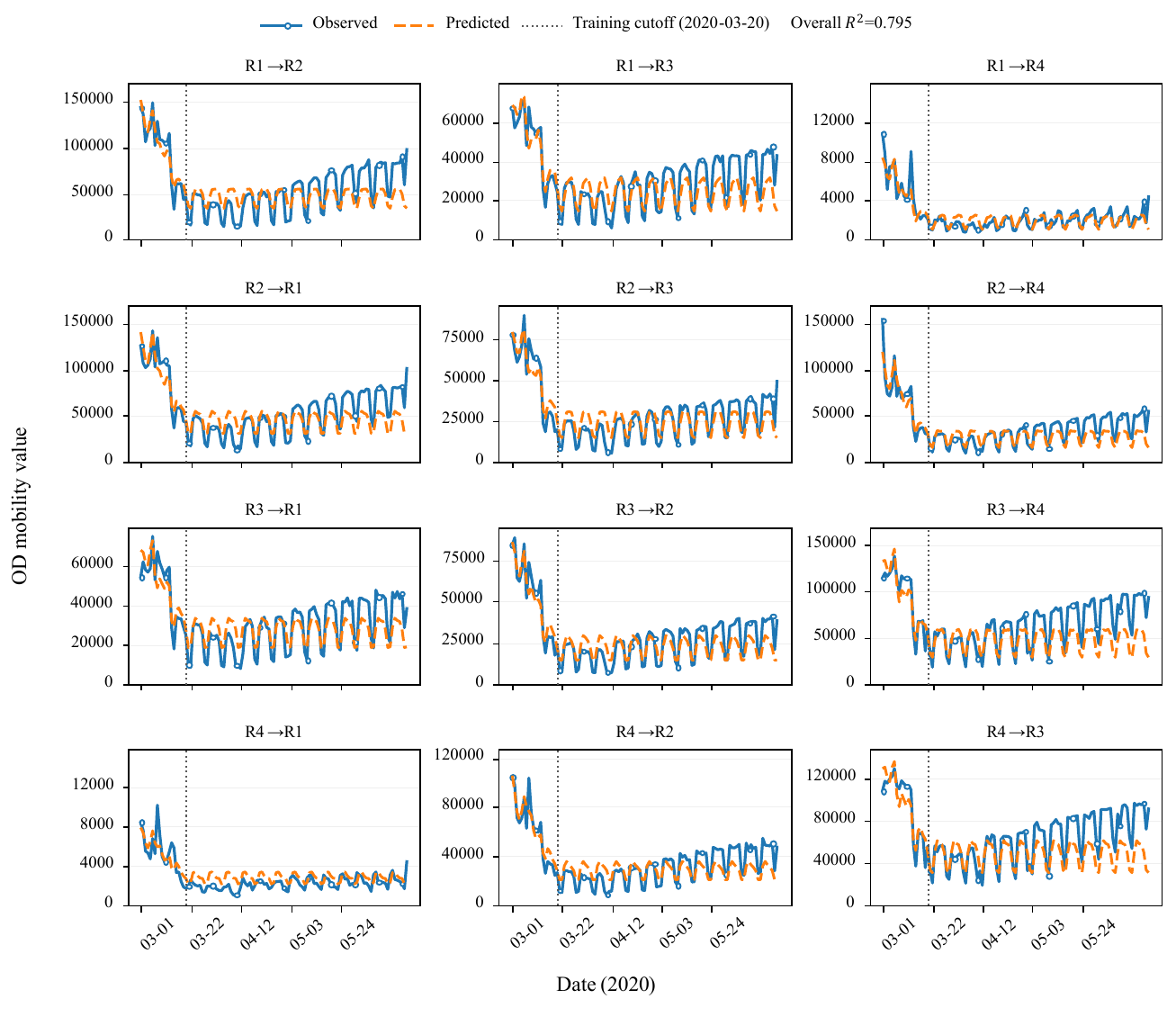}

\edcaption
{OD mobility extrapolation in Spain.}
{Observed and predicted trajectories are shown for all directed OD edges. The vertical dashed line marks the forecast origin. The auxiliary mobility predictor uses only OD observations available before the forecast origin. The overall $R^2$ displayed in the plot summarizes the complete fitting and forecasting interval; forecasting MAE and WMAPE are reported separately in Extended Data Table~\ref{tab:ed-od-errors}.}
\label{fig:ed-spain-mobility}

\end{figure*}

\clearpage
\backmatter

\bibliography{references}

\section*{Acknowledgements}

This research was partially supported by the National Natural Science Foundation of China (grant numbers 12522124 and 12371502). B.T. was also supported by the Young Talent Support Plan of Xi'an Jiaotong University.

\section*{Author contributions}

J.D., Q.L. and B.T. designed the study. J.D. wrote the code, conducted the numerical experiments and analyzed the data. All authors discussed the results and contributed to writing the manuscript.

\section*{Competing interests}

The authors declare no competing interests.

\medskip
\noindent\textbf{Correspondence and requests for materials}
should be addressed to Qian Li or Biao Tang.

\clearpage
\section*{Supplementary Information}

\graphicspath{{supplement_figures/}}

\setcounter{figure}{0}
\setcounter{table}{0}
\setcounter{equation}{0}

\renewcommand{\figurename}{Supplementary Fig.}
\renewcommand{\tablename}{Supplementary Table}
\renewcommand{\thefigure}{\arabic{figure}}
\renewcommand{\thetable}{\arabic{table}}
\renewcommand{\theequation}{S\arabic{equation}}

\renewcommand{\theHfigure}{supp.\arabic{figure}}
\renewcommand{\theHtable}{supp.\arabic{table}}
\renewcommand{\theHequation}{supp.\arabic{equation}}

\numbered
\setcounter{section}{0}
\renewcommand{\thesection}{S\arabic{section}}
\renewcommand{\theHsection}{supp.\arabic{section}}

\titleformat{\subsubsection}[block]
  {\normalfont\normalsize\bfseries}
  {\thesubsubsection}{0.5em}{}

\section{Supplementary Methods}

\subsection{Synthetic Data Generation}
\label{app:synthetic-generation}

This section provides the complete settings for synthetic data generation, including the unified dynamical equations, scenario-level parameters, contact and mobility structures, mechanism-specific parameters, and observation noise. The controlled design comprises three structural settings---Lagrangian infection generation, Eulerian population transport, and coupled multiscale dynamics---and 14 generation conditions in total: four Lagrangian mechanisms, two Eulerian mechanisms, and their eight pairwise multiscale combinations. These conditions span baseline mass-action, explicitly time-modulated, state-dependent, nonlinear, and coupled multiscale interaction dynamics. All systems are sampled daily over $0$--$200$ days. Within each structural setting, population sizes, initial states, basic epidemiological parameters, and observation-noise rules are held fixed so that differences among trajectories primarily reflect the prescribed interaction mechanisms.

\subsubsection{Dynamical System}
\label{app:synthetic-dynamics}

Let $S_j^k(t)$, $I_j^k(t)$, and $R_j^k(t)$ denote the susceptible, infectious and removed populations of age group $j\in\{1,\ldots,A\}$ in region $k\in\{1,\ldots,M\}$, respectively. The simulated trajectories are generated by the following coupled Lagrangian--Eulerian system:
\begin{equation}
\left\{
\begin{aligned}
\frac{dS_j^k}{dt}
&=
-\sum_i
\mathcal F_{i\rightarrow j}^{L}(t) I_i^k
+
\sum_l
\mathcal F_{l\rightarrow k,j}^{E}(t) S_j^l,
\\
\frac{dI_j^k}{dt}
&=
\sum_i
\mathcal F_{i\rightarrow j}^{L}(t) I_i^k
+
\sum_l
\mathcal F_{l\rightarrow k,j}^{E}(t) I_j^l
-
\gamma I_j^k,
\\
\frac{dR_j^k}{dt}
&=
\gamma I_j^k
+
\sum_l
\mathcal F_{l\rightarrow k,j}^{E}(t) R_j^l .
\end{aligned}
\right.
\label{eq:synthetic-system}
\end{equation}

Here, $\mathcal F_{i\rightarrow j}^{L}(t)$ represents the Lagrangian infection effect from age group $i$ to age group $j$, whose specific formulation is provided in Supplementary Table~\ref{tab:appendix-synthetic-mechanisms}; $\gamma$ denotes the removal rate of infectious individuals. In the multiscale scenario, the Lagrangian interaction incorporates spatial heterogeneity through the region-specific contact matrix $C_k$. For simplicity, the region index is not explicitly included in $\mathcal F_{i\rightarrow j}^{L}(t)$. Here, $i$ indexes the infectious source and $j$ the susceptible target. The current source-group population is $P_i^k(t)=S_i^k(t)+I_i^k(t)+R_i^k(t)$.

For Eulerian mobility, $\mathcal M^{lk}(t)$ denotes the OD mobility value from region $l$ to region $k$, $N^l$ denotes the population size of source region $l$, and $\eta_j$ represents the relative mobility propensity of age group $j$. The Eulerian mobility rate between regions for age group $j$ is defined as
\begin{equation}
\mathcal F_{l\rightarrow k,j}^{E}(t)
=
\begin{cases}
\dfrac{\eta_j\mathcal M^{lk}(t)}{N^l},
& l\neq k,
\\[6pt]
-\displaystyle\sum_{m\neq l}
\mathcal F_{l\rightarrow m,j}^{E}(t),
& l=k .
\end{cases}
\label{eq:synthetic-eulerian}
\end{equation}

By this definition, for any source region $l$ and age group $j$, we have
$$
\sum_k
\mathcal F_{l\rightarrow k,j}^{E}(t)
=0.
$$
Therefore, the Eulerian term only describes the redistribution of the same age group across different regions, without introducing additional population increase or decrease.

The three synthetic scenarios are obtained by different configurations of Equation~\eqref{eq:synthetic-system}. The Lagrangian scenario sets $M=1$ and disables Eulerian mobility. The Eulerian scenario sets $A=1$ and fixes the local contact intensity to $1$. The multiscale scenario sets $M=A=4$ and simultaneously activates both Lagrangian contact and Eulerian mobility mechanisms.

\subsubsection{Mechanism Definitions}

Supplementary Table~\ref{tab:appendix-synthetic-mechanisms} summarizes the Lagrangian infection mechanisms and Eulerian mobility mechanisms used for synthetic data generation. The multiscale scenario combines the four Lagrangian mechanisms with the two Eulerian mechanisms in all possible pairs. Regional epidemic activity is defined as
\[
\bar I_*^k(t)
=
\frac{\gamma}{N^k}
\sum_{j=1}^{A} I_j^k(t),
\]
where $N^k$ is the fixed reference total population of region $k$.
The infectious states and population denominator are expressed
on the same population scale.
\begin{table*}[t]
\centering
\caption{\textbf{Synthetic dynamical mechanisms.} Functional forms and dynamical classes used to generate the Lagrangian, Eulerian and multiscale synthetic systems. The eight multiscale conditions combine each Lagrangian mechanism with each Eulerian mechanism. Here, $N^l$ and $N^k$ are the fixed reference total populations of the source and destination regions, respectively, and $N_0$ is the reference population used to scale the gravity terms.}
\label{tab:appendix-synthetic-mechanisms}
\scriptsize
\renewcommand{\arraystretch}{1.22}
\setlength{\tabcolsep}{4pt}

\begin{tabular}{
>{\raggedright\arraybackslash}p{5.7cm}
>{\raggedright\arraybackslash}p{3.5cm}
l}
\toprule
\textbf{Mechanism}
& \textbf{Dynamical class}
& \textbf{Functional form} \\
\midrule

\multicolumn{3}{l}{\textbf{Lagrangian mechanisms}} \\[1pt]

Baseline Mass-Action Contact Mechanism (BMCM)
& Baseline mass-action
& $\displaystyle
\mathcal F_{i\rightarrow j}^{L}(t)
=\beta C_{ij}^{k}\frac{S_j^k(t)}{P_i^k(t)}$ \\[2pt]

Time-Decaying Contact Mechanism (TDCM)
& Explicitly time-modulated
& $\displaystyle
\mathcal F_{i\rightarrow j}^{L}(t)
=\beta e^{-r_Lt}C_{ij}^{k}\frac{S_j^k(t)}{P_i^k(t)}$ \\[2pt]

Infection-Avoidant Contact Mechanism (IACM)
& State-dependent feedback
& $\displaystyle
\mathcal F_{i\rightarrow j}^{L}(t)
=\beta e^{-\alpha_I I_i^k(t)}C_{ij}^{k}
\frac{S_j^k(t)}{P_i^k(t)}$ \\[2pt]

Nonlinear Mass-Action Contact Mechanism (NMCM)
& Nonlinear state dependence
& $\displaystyle
\mathcal F_{i\rightarrow j}^{L}(t)
=\beta[I_i^k(t)]^{q_I}C_{ij}^{k}
\frac{[S_j^k(t)]^{p_S}}{P_i^k(t)}$ \\

\midrule
\multicolumn{3}{l}{\textbf{Eulerian mechanisms}} \\[1pt]

Baseline Risk-Aware Gravity Mobility (BRGM)
& State-dependent mobility
& $\displaystyle
\mathcal{M}_{\mathrm{BRGM}}^{lk}
=\kappa
\left(\frac{N^l}{N_0}\right)^a
\left(\frac{N^k}{N_0}\right)^b
e^{-\lambda d_{lk}-\rho_s\bar I^l-\rho_d\bar I^k}$ \\[3pt]

Controlled Risk-Aware Gravity Mobility (CRGM)
& State- and time-modulated mobility
& $\displaystyle
\mathcal{M}_{\mathrm{CRGM}}^{lk}
=\mathcal{M}_{\mathrm{BRGM}}^{lk}e^{-r_Et}$ \\

\midrule
\multicolumn{3}{l}{\textbf{Multiscale mechanisms}} \\[1pt]

Lagrangian--Eulerian coupled mechanisms
& Coupled multiscale dynamics
& \makecell[l]{
$\{\mathrm{BMCM},\mathrm{TDCM},\mathrm{IACM},\mathrm{NMCM}\}$\\[2pt]
$\qquad\times\{\mathrm{BRGM},\mathrm{CRGM}\}$
} \\

\bottomrule
\end{tabular}
\end{table*}

The mechanisms are not intended to reproduce one particular epidemic outbreak. Instead, they provide controlled systems with known generating rules for evaluating forecasting and process-level dynamical fidelity across qualitatively different interaction dynamics. Together, the 14 conditions cover a stationary baseline, explicit temporal modulation, endogenous state-dependent feedback, nonlinear state dependence, and coupled age--space dynamics. The Lagrangian mechanisms vary infection-generation interactions driven by age-specific contact structure within regions, whereas the Eulerian mechanisms vary conservative population transport across regions.

\paragraph{Lagrangian Infection Mechanisms.}

The Lagrangian mechanisms are used to generate baseline, explicitly time-modulated, state-dependent, and nonlinear variations in local infection-generation intensity.

\noindent\textbf{Baseline Mass-Action Contact Mechanism (BMCM)}
represents the basic transmission process under a stable contact environment. This mechanism assumes that age-specific contact patterns remain approximately unchanged over time. Infection generation is primarily determined by the predefined contact structure, without considering additional behavioral adjustments or intervention effects, and therefore serves as the baseline scenario for other non-stationary transmission mechanisms.

\noindent\textbf{Time-Decaying Contact Mechanism (TDCM)}
introduces explicit temporal modulation through a gradually decreasing interaction intensity. This stylized mechanism represents settings in which effective contacts decline over time because of interventions, increasing risk awareness, or behavioral change.

\noindent\textbf{Infection-Avoidant Contact Mechanism (IACM)}
introduces endogenous state-dependent feedback through infection-responsive contact adjustment. As infection levels increase, effective interaction intensity changes with the evolving epidemic state.

\noindent\textbf{Nonlinear Mass-Action Contact Mechanism (NMCM)}
introduces nonlinear state dependence in infection generation, providing a controlled setting in which interaction strength responds nonlinearly to susceptible and infectious states.

\paragraph{Eulerian Mobility Mechanisms.}

The Eulerian mechanisms are used to generate state-dependent and explicitly time-modulated redistribution of epidemic states through interregional population transport.

\noindent\textbf{Baseline Risk-Aware Gravity Mobility (BRGM)}
defines state-dependent mobility in which interregional transport varies with source- and destination-region epidemic activity as well as population and distance effects.

\noindent\textbf{Controlled Risk-Aware Gravity Mobility (CRGM)}
adds explicit temporal modulation to the risk-aware mobility process, providing a controlled representation of progressively strengthened travel restrictions or other time-dependent mobility suppression.

\subsubsection{Scenario-level Settings}
\label{app:synthetic-settings}

The basic parameters of the three scenarios are summarized in Supplementary Table~\ref{tab:synthetic-base-parameters}. Here, $M$ and $A$ denote the number of regions and age groups, respectively, and $\eta$ represents the relative mobility propensity of different age groups. During synthetic data generation, the compartment states $(S,I,R)$ and the observation $\gamma I$ are scaled by $10^{-4}$ in the Lagrangian and multiscale settings and by $10^{-5}$ in the Eulerian setting. The population sizes and initial conditions reported below are given in individuals.

\begin{table}[t]
\centering
\caption{\textbf{Synthetic scenario settings.} Shared configurations used to generate the Lagrangian, Eulerian and multiscale systems.}
\label{tab:synthetic-base-parameters}
\scriptsize
\renewcommand{\arraystretch}{1.12}
\setlength{\tabcolsep}{4pt}
\begin{tabularx}{\textwidth}{
>{\raggedright\arraybackslash}p{4.0cm}
>{\raggedright\arraybackslash}X
>{\raggedright\arraybackslash}X
>{\raggedright\arraybackslash}X}
\toprule
\textbf{Setting}
& \textbf{Lagrangian}
& \textbf{Eulerian}
& \textbf{Multiscale} \\
\midrule

Graph size
& $M=1,\ A=4$
& $M=4,\ A=1$
& $M=4,\ A=4$ \\

Simulation horizon
& $0$--$200$ days
& $0$--$200$ days
& $0$--$200$ days \\

Sampling interval
& $1$ day
& $1$ day
& $1$ day \\

Transmission rate $\beta$
& $0.01$
& $0.20$
& $0.02$ \\

Removal rate $\gamma$
& $0.02$
& $0.08$
& $0.021$ \\

Mobility propensity $\eta$
& --
& $[1]$
& $[0.106,\,0.685,\,0.156,\,0.053]$ \\

Gravity exponents $(a,b)$
& --
& $(1,1)$
& $(1,1)$ \\

Distance decay $\lambda$
& --
& $0.012$
& $0.012$ \\

Gravity scale $\kappa$
& --
& $3000$
& $500000$ \\

Reference population $N_0$
& --
& $10^{5}$
& $10^{6}$ \\

Risk coefficients $(\rho_s,\rho_d)$
& --
& $(6,10)$
& $(6,10)$ \\

ODE solver
& LSODA
& LSODA
& LSODA \\

\bottomrule
\end{tabularx}
\end{table}

Within each structural setting, the population sizes and initial infection states are fixed across generation conditions. For the Lagrangian scenario,
$$
\bm N^{L}
=
10^{3}
\begin{bmatrix}
82 & 102 & 130 & 65
\end{bmatrix},
\qquad
\bm I^{L}(0)
=
\begin{bmatrix}
1900 & 1000 & 1200 & 1300
\end{bmatrix}.
$$
For the Eulerian scenario,
$$
\bm N^{E}
=
10^{3}
\begin{bmatrix}
130\\
100\\
90\\
80
\end{bmatrix},
\qquad
\bm I^{E}(0)
=
\begin{bmatrix}
500\\
0\\
0\\
0
\end{bmatrix}.
$$
For the multiscale scenario,
$$
\bm N^{M}
=
10^{3}
\begin{bmatrix}
120&165&190&115\\
110&155&180&105\\
100&145&170&95\\
115&160&185&110
\end{bmatrix},
$$
$$
\bm I^{M}(0)
=
\begin{bmatrix}
1900&1000&1200&1300\\
0&0&0&0\\
0&0&0&0\\
0&0&0&0
\end{bmatrix}.
$$
The rows and columns of the population matrices and initial infection matrices correspond to regions and age groups, respectively. All scenarios satisfy
$$
\bm R(0)=0,
\qquad
\bm S(0)=\bm N-\bm I(0).
$$
The initial infections in the Eulerian and multiscale scenarios are concentrated in the first region, providing a clear initial spatial transmission condition and enabling the investigation of interregional redistribution patterns generated by different mobility mechanisms.

\subsubsection{Contact and Mobility Structures}
\label{app:synthetic-structures}

The Lagrangian scenario adopts a fixed baseline contact matrix
$$
C^{L}
=
\begin{bmatrix}
3.20&2.92&2.79&3.17\\
2.88&3.00&2.54&3.13\\
2.71&2.56&2.50&2.81\\
3.23&3.22&2.84&3.50
\end{bmatrix}.
$$
BMCM, TDCM, IACM, and NMCM all generate corresponding Lagrangian infection effects based on this contact structure.

In the Eulerian scenario, each region contains only one age group, and the local contact intensity is uniformly set to $1.0$. The interregional distance matrix is defined as
$$
D^{E}
=
\begin{bmatrix}
0&120&210&260\\
120&0&150&180\\
210&150&0&140\\
260&180&140&0
\end{bmatrix}.
$$

The multiscale scenario adopts a region-dependent baseline contact matrix
$$
C_k=s_kC_0,
\qquad
[s_1,s_2,s_3,s_4]
=
[1.00,\,0.98,\,0.95,\,0.92],
$$
where
$$
C_0
=
\begin{bmatrix}
1.600&1.460&1.395&1.585\\
1.550&1.500&1.420&1.565\\
1.605&1.580&1.450&1.655\\
1.615&1.450&1.370&1.500
\end{bmatrix}.
$$
The corresponding interregional distance matrix is
$$
D^{M}
=
\begin{bmatrix}
0&120&170&310\\
120&0&130&220\\
170&130&0&160\\
310&220&160&0
\end{bmatrix}.
$$

The Lagrangian scenario does not include Eulerian mobility. In the Eulerian and multiscale scenarios, $\mathcal M^{lk}(t)$ is generated by BRGM or CRGM defined in Supplementary Table~\ref{tab:appendix-synthetic-mechanisms}, respectively, and is converted into the corresponding Eulerian mobility rate $\mathcal F_{l\rightarrow k,j}^{E}(t)$ for each age group according to Equation~\eqref{eq:synthetic-eulerian}.
\subsubsection{Mechanism-specific Parameters}
\label{app:synthetic-mechanism-parameters}

In addition to the scenario-level shared parameters described above, TDCM, IACM, NMCM, and CRGM introduce additional mechanism-specific parameters. Their values are provided in Supplementary Table~\ref{tab:synthetic-mechanism-parameters}. BMCM and BRGM serve as the baseline mechanisms for Lagrangian and Eulerian dynamics, respectively, and do not introduce additional temporal modulation parameters.

\begin{table}[t]
\centering
\caption{\textbf{Mechanism-specific parameters.} Parameter values used by the explicitly time-modulated, state-dependent and nonlinear synthetic mechanisms.}
\label{tab:synthetic-mechanism-parameters}
\small
\renewcommand{\arraystretch}{1.1}
\setlength{\tabcolsep}{5pt}
\begin{tabular}{lcccc}
\toprule
\textbf{Scenario}
& \textbf{TDCM: $r_L$}
& \textbf{IACM: $\alpha_I$}
& \textbf{NMCM: $(p_S,q_I)$}
& \textbf{CRGM: $r_E$} \\
\midrule

Lagrangian
& $0.01$
& $0.10$
& $(1.0,\,0.4)$
& -- \\

Eulerian
& --
& --
& --
& $0.30$ \\

Multiscale
& $0.0035$
& $0.045$
& $(1.0,\,0.2)$
& $0.03$ \\

\bottomrule
\end{tabular}
\end{table}

The multiscale scenario combines the four Lagrangian mechanisms \{BMCM, TDCM, IACM, NMCM\} 
with the two Eulerian mechanisms
$\{\mathrm{BRGM},\mathrm{CRGM}\}$
in all possible pairs, resulting in eight coupled Lagrangian--Eulerian mechanism combinations. All combinations share the same population sizes, initial states, basic epidemiological parameters, baseline contact structures, and region distance matrices. Therefore, the systematic differences among generated trajectories mainly arise from the adopted dynamical mechanisms.

\subsubsection{Observation Noise}
\label{app:synthetic-noise}

To avoid making the simulated observations identical to the deterministic numerical trajectories, we add mild observation noise to the infection states and OD mobility data, respectively. The noise is applied only at the observation level and does not participate in the state evolution described by Equation~\eqref{eq:synthetic-system}.

For the noise-free infection sequence $I_v(t)$ of node $v$, the standard deviation of Gaussian noise is set to $5\%$ of its maximum value over the entire simulation period:
$$
\widetilde I_v(t)
=
\max
\left\{
0,\,
I_v(t)+\epsilon_v(t)
\right\},
\qquad
\epsilon_v(t)
\sim
\mathcal N
\left(
0,\,
[0.05I_{v,\max}]^2
\right),
$$
where
$$
I_{v,\max}
=
\max_t I_v(t).
$$

The corresponding observed case count is calculated as
$$
\widetilde Y_v(t)
=
\gamma\widetilde I_v(t).
$$
The values of $S_v(t)$ and $R_v(t)$ remain the results of noise-free numerical integration.

For the Eulerian and multiscale scenarios, noise is further independently added to each directed OD mobility sequence. Let $\mathcal M^{lk}(t)$ denote the noise-free OD mobility value from region $l$ to region $k$. Then,
$$
\widetilde{\mathcal M}^{lk}(t)
=
\max
\left\{
0,\,
\mathcal M^{lk}(t)+\epsilon_{lk}(t)
\right\},
\qquad
\epsilon_{lk}(t)
\sim
\mathcal N
\left(
0,\,
[0.05\mathcal M_{\max}^{lk}]^2
\right),
\quad l\neq k,
$$
where
$$
\mathcal M_{\max}^{lk}
=
\max_t
\left|
\mathcal M^{lk}(t)
\right|.
$$

The noisy OD matrices preserve zero diagonal entries and are used to reconstruct the Eulerian mobility rates for each age group according to Equation~\eqref{eq:synthetic-eulerian}. The infection-state noise and OD mobility noise use fixed random seeds $42$ and $43$, respectively, to ensure reproducible comparisons across different mechanisms.

Therefore, within the same scenario, all generated datasets share identical initial conditions, basic parameters, and observation perturbation rules. Their primary dynamical differences arise from the predefined Lagrangian mechanisms, Eulerian mechanisms, or their combinations.

\subsubsection{Synthetic Data Overview}
\label{app:synthetic-visualization}

\begin{figure}[!htbp]
\centering

\begin{subfigure}{\linewidth}
    \centering
    \includegraphics[
        width=\linewidth,
        height=\textheight,
        keepaspectratio
    ]{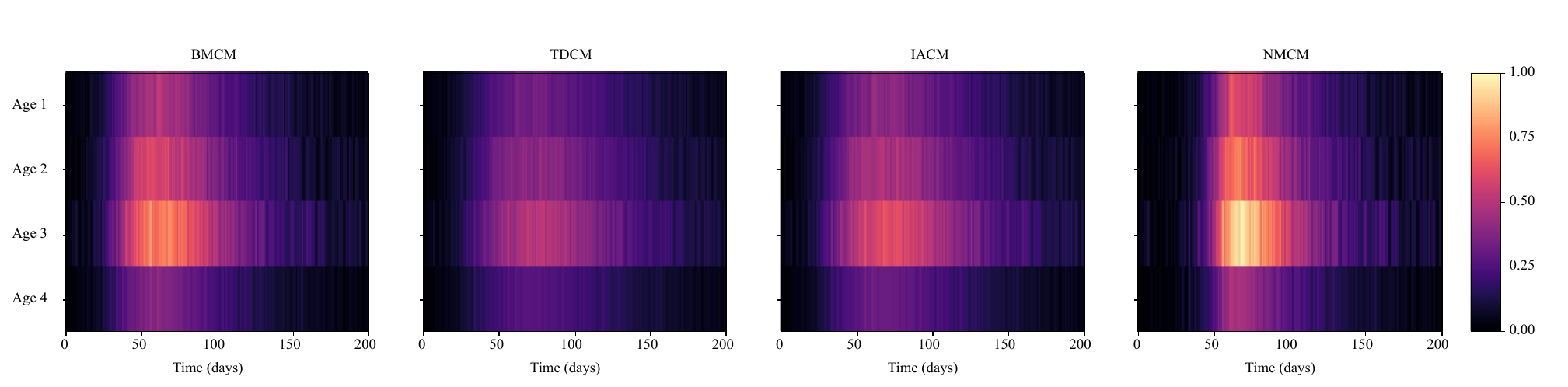}
    \caption{Four Lagrangian infection-generation mechanisms.}
    \label{fig:synthetic-lagrangian}
\end{subfigure}

\vspace{0.6em}

\begin{subfigure}{\linewidth}
    \centering
    \includegraphics[
        width=\linewidth,
        height=\textheight,
        keepaspectratio
    ]{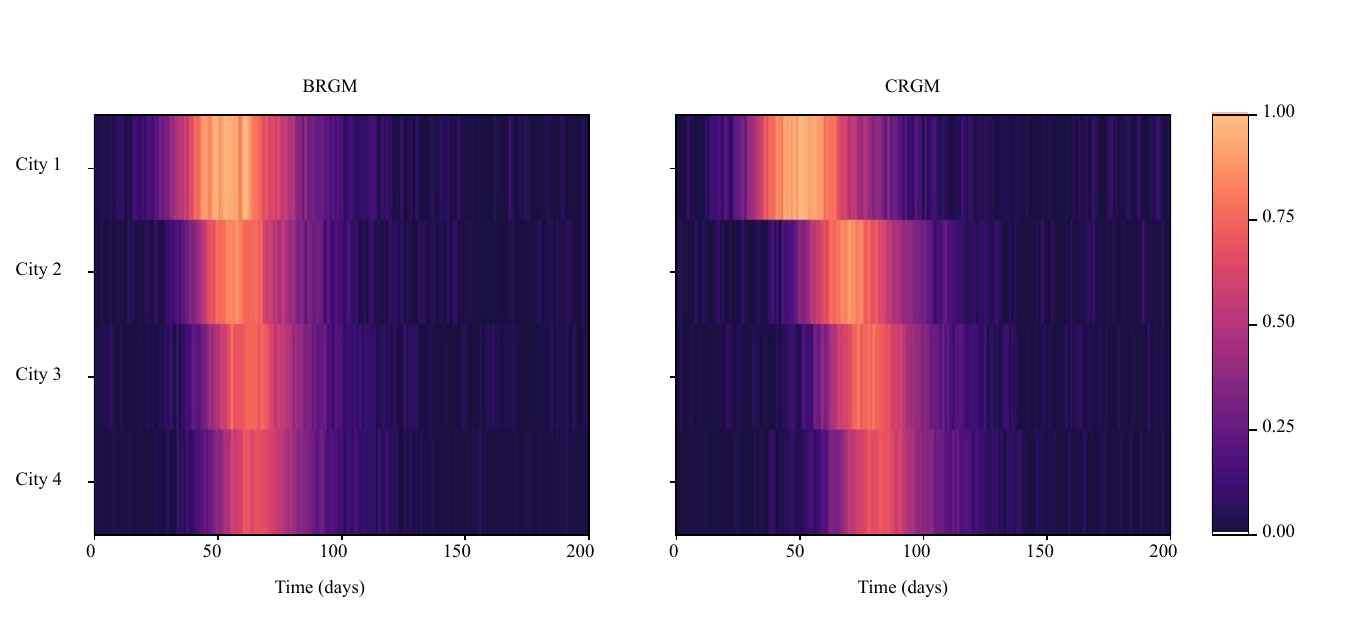}
    \caption{Two Eulerian population-transport mechanisms.}
    \label{fig:synthetic-eulerian}
\end{subfigure}

\caption{
\textbf{Synthetic trajectories in the isolated Lagrangian and Eulerian settings.}
Heatmaps show normalized infectious-state trajectories $I_j^k(t)$ generated by the controlled interaction mechanisms.
Within each structural setting, population sizes, initial states, basic epidemiological parameters and observation rules are held fixed so that differences among trajectories primarily reflect the prescribed dynamical mechanisms.
}
\label{fig:synthetic-groundtruth-local}
\end{figure}

\begin{figure}[p]
\centering

\includegraphics[
    width=\textwidth,
    height=\textheight,
    keepaspectratio
]{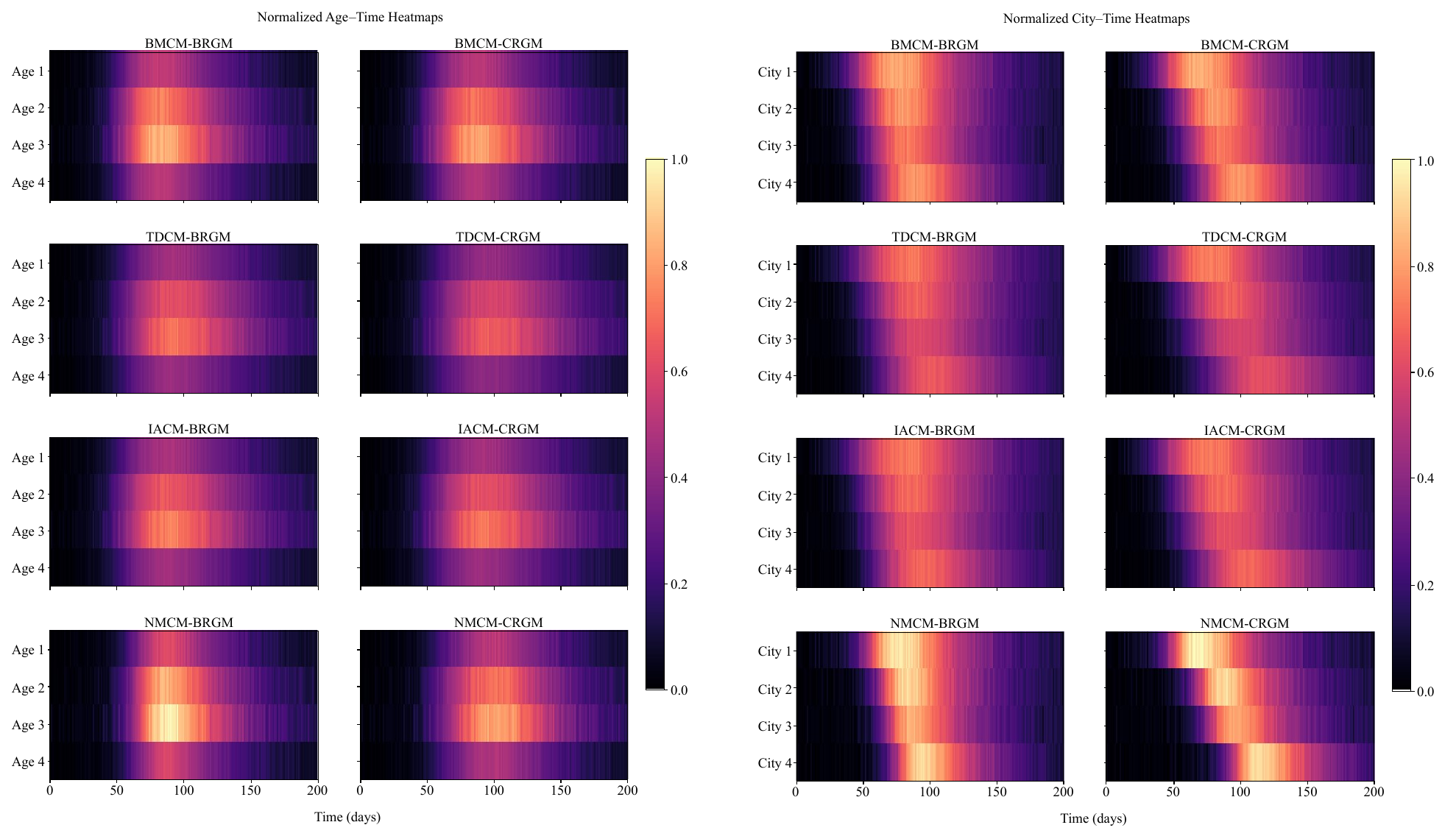}

\caption{
\textbf{Multiscale ground-truth signatures of the coupled synthetic mechanisms.}
Normalized infection-intensity patterns are shown for the eight pairwise combinations
of four Lagrangian mechanisms (BMCM, TDCM, IACM and NMCM) and two Eulerian
mechanisms (BRGM and CRGM). The left block shows age--time patterns across four
age groups, whereas the right block shows the corresponding city--time patterns across
four cities. In both blocks, rows correspond to the four Lagrangian mechanisms and
columns correspond to BRGM and CRGM. Comparisons across rows under the same
Eulerian mechanism highlight Lagrangian mechanism-specific dynamical signatures,
whereas comparisons between BRGM and CRGM within the same row highlight the
additional effects associated with the Eulerian mobility mechanism. Together, these
patterns illustrate the complementary age- and region-level signatures generated by
the coupled multiscale mechanisms.
}
\label{fig:synthetic-groundtruth-multiscale}

\end{figure}

Supplementary Fig.~\ref{fig:synthetic-groundtruth-local} presents infection trajectories
generated by the isolated Lagrangian and Eulerian mechanisms. The Lagrangian setting
contrasts baseline mass-action, explicitly time-modulated, state-dependent and nonlinear
infection-generation dynamics, whereas the Eulerian setting contrasts BRGM with the
additional temporal modulation introduced by CRGM. Supplementary
Fig.~\ref{fig:synthetic-groundtruth-multiscale} extends this overview to the eight pairwise
Lagrangian--Eulerian combinations in the coupled multiscale setting, showing their
complementary age--time and city--time infection-intensity patterns.

Because population sizes, initial states, basic epidemiological parameters and observation rules are held fixed within each structural setting, systematic differences across these trajectories and heatmaps primarily reflect the prescribed interaction mechanisms.

\clearpage

\subsection{Real-world Data and Mobility Predictor Settings}
\label{app:real-mobility}

The observation structures and evaluation settings of the three real-world datasets are summarized in Supplementary Table~\ref{tab:real-data-settings}.
\begin{table*}[!t]
\centering
\caption{
\textbf{Data and evaluation settings for the three real-world COVID-19 datasets.}
}
\label{tab:real-data-settings}

\small
\renewcommand{\arraystretch}{1.30}
\setlength{\tabcolsep}{5pt}

\begin{tabularx}{\textwidth}{
    >{\raggedright\arraybackslash}p{3.0cm}
    >{\raggedright\arraybackslash}X
    >{\raggedright\arraybackslash}X
    >{\raggedright\arraybackslash}X
}
\toprule
\textbf{Setting}
& \textbf{Ontario}
& \textbf{Hubei}
& \textbf{Spain} \\
\midrule

Epidemic setting
& Age structured
& Mobility driven
& Coupled age--space \\

State structure
& 1 region $\times$ 4 age groups
& 3 regions $\times$ 1 age group
& 4 regions $\times$ 3 age groups \\

Epidemic observations
& Age-stratified daily cases
& City-level daily cases
& Region--age daily cases \\

Process-aligned information
& Age-contact matrix
& Daily OD mobility; population and distance
& Age-contact matrices; daily OD mobility; population and distance \\

Observation period
& 28 Nov. 2021--27 Feb. 2022
& 16 Jan. 2020--25 Feb. 2020
& 1 Mar. 2020--21 Jun. 2020 \\

Training
& 25 days
& 15 days
& 20 days \\

Forecasting
& 7 / 14 / 35 / 67 days
& 7 / 14 / 26 days
& 7 / 14 / 49 / 93 days \\

\bottomrule
\end{tabularx}
\end{table*}
This section supplements the input feature settings of the auxiliary mobility predictor $\operatorname{MLP}_{\mathrm{mob}}$ across settings involving Eulerian transport. All such experiments use the same auxiliary mobility-predictor architecture:
\begin{equation}
\hat{\mathcal{M}}^{lk}(t)
=
\operatorname{softplus}
\left[
\operatorname{MLP}_{\mathrm{mob}}
\left(
z_{lk}(t);\vartheta
\right)
\right],
\label{eq:appendix-mobility-predictor}
\end{equation}
where $z_{lk}(t)$ denotes the dataset-specific input vector defined below. The predictor estimates directed OD mobility values using epidemic activity, spatial gravity priors and temporal information, with regional identity encodings additionally included for Spain.

\subsubsection{Common Input Representation}

For a directed edge $l\rightarrow k$, the basic input is defined as
$$
z_{lk}^{\mathrm{base}}(t)
=
\bar I_*^l(t)
\mathbin\Vert
\bar I_*^k(t)
\mathbin\Vert
e_{lk}^{E}
\mathbin\Vert
t_{\mathrm{norm}},
$$
where $\bar I_*^l(t)$ and $\bar I_*^k(t)$ denote the source and destination epidemic activities, using observed reporting activity during mobility pretraining and model-generated reporting activity during epidemic training and forecasting. The term $e_{lk}^{E}$ is the gravity prior constructed from population and spatial distance information, and $t_{\mathrm{norm}}$ represents the normalized temporal coordinate.

\subsubsection{Hubei Input Setting}

The Hubei experiment directly adopts the above basic input representation:
$$
z_{lk}^{\mathrm{HB}}(t)
=
\bar I_*^l(t)
\mathbin\Vert
\bar I_*^k(t)
\mathbin\Vert
e_{lk}^{E}
\mathbin\Vert
t_{\mathrm{norm}}.
$$
The Hubei OD sequences mainly exhibit low-frequency variations. Therefore, no additional periodic temporal features or regional identity features are introduced.

\subsubsection{Spain Input Setting}

The Spain OD sequences exhibit more apparent short-period fluctuations. To enhance the ability of the auxiliary mobility network to capture such temporal patterns, we adopt a training-window normalized and clipped temporal coordinate
$$
\tau(t)
=
\min
\left\{
\max
\left(
\frac{t}{T_{\mathrm{train}}-1},
0
\right),
1
\right\},
$$
and incorporate weekly periodic sine and cosine features:
$$
\psi_{\mathrm{ES}}(t)
=
\left[
\tau(t),
\sin\left(\frac{2\pi t}{7}\right),
\cos\left(\frac{2\pi t}{7}\right)
\right].
$$
In addition, considering the stable spatial heterogeneity among different aggregated regions, we further include one-hot identity encodings $r_l$ and $r_k$ for the source and destination regions. Therefore, the input representation for the Spain experiment is
$$
z_{lk}^{\mathrm{ES}}(t)
=
\bar I_*^l(t)
\mathbin\Vert
\bar I_*^k(t)
\mathbin\Vert
e_{lk}^{E}
\mathbin\Vert
\psi_{\mathrm{ES}}(t)
\mathbin\Vert
r_l
\mathbin\Vert
r_k.
$$

\subsubsection{Input Feature Summary}

Supplementary Table~\ref{tab:mobility-input-settings} summarizes the mobility-predictor inputs for the synthetic, Hubei and Spain datasets. The synthetic Eulerian and multiscale experiments use the same basic representation as Hubei: source and destination epidemic activity, the gravity prior and normalized time, without periodic temporal features or regional identity encodings.

\begin{table}[t]
\centering
\caption{\textbf{Mobility-predictor inputs.} Input features used by the auxiliary mobility network in the synthetic, Hubei and Spain datasets.}
\label{tab:mobility-input-settings}
\small
\renewcommand{\arraystretch}{1.12}
\setlength{\tabcolsep}{6pt}
\begin{tabular}{lccc}
\toprule
\textbf{Input feature}
& \textbf{Synthetic}
& \textbf{Hubei}
& \textbf{Spain} \\
\midrule

Source epidemic intensity
& Yes & Yes & Yes \\

Destination epidemic intensity
& Yes & Yes & Yes \\

Gravity prior
& Yes & Yes & Yes \\

Normalized time
& Yes & Yes & Yes \\

Weekly periodic encoding
& No & No & Yes \\

Source region identity
& No & No & Yes \\

Destination region identity
& No & No & Yes \\

\bottomrule
\end{tabular}
\end{table}

\subsection{Baseline Model Configurations}
\label{app:baseline-configurations}

This section summarizes the implementation details, model architectures, and final hyperparameters of the 21 comparator models used across the main experiments. The comparator set spans mechanistic models, general sequence and spatiotemporal forecasting models, epidemic-specific graph models, and hybrid mechanistic--learning models; not every comparator is used in every setting, as detailed below. Supplementary Table~\ref{tab:baseline-implementation-summary} describes the experimental roles and study-specific implementations; Supplementary Table~\ref{tab:synthetic-baseline-configurations} reports the final configuration ranges used across the 14 synthetic generation conditions; Supplementary Table~\ref{tab:real-baseline-configurations} reports the specific configurations on the three real-world datasets, Ontario, Hubei, and Spain. All configurations correspond to the runs producing the main results rather than to predefined hyperparameter search spaces.

In the synthetic experiment table, numerical ranges represent the minimum and maximum values of the configurations actually adopted across the 14 generation conditions. In the real-world experiment table, unless otherwise specified, three values separated by slashes correspond to Ontario, Hubei, and Spain, respectively. For deep learning baselines, \textit{Hidden/layers} denotes the hidden dimension and number of network layers, \textit{Dropout/batch} denotes the dropout rate and batch size, and \textit{Input window} denotes the number of historical observation days used to construct each prediction sample. \textit{Model-specific settings} summarize model-specific parameters that cannot be included in the above general categories, including ODE solver settings, graph convolution orders, embedding dimensions, and propagation depths. All deep learning baselines use direct 7-day prediction outputs. M-Graphormer is a paper-level reproduction based on the descriptions in the original paper, and EpiGCN is an adaptation of the original EpiGCN graph convolution module for continuous-time prediction.

\begin{longtable}{
>{\raggedright\arraybackslash}p{2.4cm}
>{\raggedright\arraybackslash}p{3.30cm}
>{\raggedright\arraybackslash}p{9.55cm}}
\caption{\textbf{Baseline implementations.} Experimental roles and study-specific configurations of the comparison models.}
\label{tab:baseline-implementation-summary}\\
\toprule
\textbf{Model}
& \textbf{Role}
& \textbf{Implementation in this study} \\
\midrule
\endfirsthead

\multicolumn{3}{c}{\tablename~\thetable\ (continued)}\\
\toprule
\textbf{Model}
& \textbf{Role}
& \textbf{Implementation in this study} \\
\midrule
\endhead

\midrule
\multicolumn{3}{r}{Continued on next page}\\
\endfoot

\bottomrule
\endlastfoot

SIR
& Synthetic baseline
& Independent nodewise SIR dynamics with nonnegative initial compartments and node-specific transmission and removal parameters fitted to the complete training trajectory. \\

MepoSIR
& Synthetic baseline
& Contact-coupled metapopulation SIR using the observed age-contact matrix, region-shared age-specific transmission parameters and node-specific removal parameters. \\

LSTM
& Synthetic baseline
& Stacked LSTM encoder with a direct 7-day regression head, trained on permitted node histories and rolled out autoregressively in 7-day blocks. \\

GRU
& Synthetic baseline
& Stacked GRU encoder with the same direct 7-day prediction and blockwise autoregressive rollout protocol as LSTM. \\

Neural ODE
& Synthetic baseline
& MLP vector field integrated by fixed-step RK4, with the historical window encoded as the initial latent state and decoded into a 7-day trajectory. \\

STGCN
& Synthetic baseline
& Temporal--Chebyshev graph convolution blocks using spatial supports constructed from permitted contact, distance and training-period mobility information. \\

DCRNN
& Synthetic and real baseline
& Diffusion-convolutional GRU encoder with a single autoregressive decoder and scheduled sampling, generating seven future days per block. \\

AGCRN
& Synthetic and real baseline
& Adaptive graph-convolutional recurrence with learned node embeddings and node-specific parameter generation, using only permitted node histories. \\

Graph WaveNet
& Synthetic baseline
& Dilated gated temporal convolutions with residual, skip and adaptive graph-propagation paths, followed by a direct 7-day output head. \\

MTGNN
& Synthetic and real baseline
& Adaptive graph construction, multiscale dilated temporal convolutions and bidirectional MixProp with a direct multi-step head and dataset-specific graph settings. \\

MCM-SEIR
& Real baseline
& Contact- and mobility-coupled metapopulation SEIR using observed contact matrices, a conservative OD-derived mobility generator and fixed dataset-specific incubation rates. \\

EpiGNN
& Real baseline
& Historical GRU encoding, epidemic-aware adaptive adjacency and stacked graph propagation, with the original single-target output replaced by a direct 7-day head. \\

Cola-GNN
& Real baseline
& Recurrent temporal encoding with a learned dynamic correlation graph and graph propagation, extended from single-step to direct 7-day prediction without future-value feedback. \\

CovidGNN
& Real baseline
& Spatio-temporal graph convolution over region--time relations, trained with strictly training-contained sliding windows instead of the original whole-graph train/test mask. \\

MepoGNN
& Real baseline
& Graph WaveNet-style temporal blocks, bidirectional mobility propagation, rate heads and metapopulation SIR recursion using training-period OD supports and calibrated physical readouts. \\

M-Graphormer
& Real baseline
& Paper-level reproduction combining gated TCNs, Graphormer attention, structural encodings, gated dense connections and metapopulation recursion with a direct 7-day head and node-level calibration. \\

EpiGCN
& Real baseline
& EpidemicGCN susceptible--infected--removed message propagation used as a continuous-time vector field with node calibration, excluding unavailable street-view features. \\

STAN
& Real baseline
& Multi-head GAT and GRU temporal encoding with dual data and epidemic heads, using historical $\gamma I$, population and causal first differences in place of unavailable compartment observations. \\

EINN
& Real baseline
& Bidirectional GRU, masked self-attention, Fourier time features, an SEIRm residual and knowledge distillation, trained on expanding prefixes with a direct 7-day target on $\gamma I$. \\

Epi-Cola-GNN
& Real baseline
& Cola-GNN data and epidemic-dynamics heads, both extended from a single target to seven jointly optimized outputs. \\

EARTH
& Real baseline
& Controlled temporal path, continuous graph dynamics, epidemic multi-view representations and attention fusion with a direct 7-day output head. \\

\end{longtable}

\begin{sidewaystable}[p]
\centering
\caption{\textbf{Synthetic baseline configurations.} Ranges summarize the minimum and maximum hyperparameter values used across the 14 synthetic generation conditions, rather than hyperparameter search spaces.
}
\label{tab:synthetic-baseline-configurations}
\scriptsize
\renewcommand{\arraystretch}{2.0}
\setlength{\tabcolsep}{2.8pt}

\begin{tabularx}{\textheight}{
@{}p{2.2cm}
p{4.3cm}
p{2.6cm}
p{3.3cm}
p{2.4cm}
X@{}}
\toprule
\textbf{Model}
& \textbf{Learning rate}
& \textbf{Hidden/layers}
& \textbf{Dropout/batch}
& \textbf{Input window}
& \textbf{Model-specific setting} \\
\midrule

SIR
& $1.6341\times10^{-4}$--$2.5972\times10^{-2}$
& N/A
& N/A
& N/A
& $\beta_{\mathrm{init}}=0.01703$--$0.48997$; $\gamma_{\mathrm{init}}=0.005572$--$0.19871$ \\

MepoSIR
& $1.1078\times10^{-4}$--$4.5878\times10^{-2}$
& N/A
& N/A
& N/A
& $\beta_{\mathrm{init}}=0.01113$--$0.78695$; $\gamma_{\mathrm{init}}=0.005099$--$0.36765$ \\

LSTM
& $1.6092\times10^{-4}$--$4.5127\times10^{-3}$
& $32$--$96$ / $1$--$3$
& $0$--$0.3$ / $4$--$16$
& $7$--$14$
& stacked LSTM encoder with a direct multi-step regression head \\

GRU
& $1.0611\times10^{-4}$--$3.3113\times10^{-3}$
& $64$--$96$ / $1$--$3$
& $0$--$0.3$ / $4$--$16$
& $7$--$28$
& stacked GRU encoder with a direct multi-step regression head \\

Neural ODE
& $1.0744\times10^{-4}$--$4.9943\times10^{-3}$
& $16$--$96$ / $1$--$3$
& $0$--$0.3$ / $4$--$16$
& $7$--$28$
& Fixed-step RK4 solver; step size $0.5$--$1.0$ \\

STGCN
& $2.1638\times10^{-4}$--$3.3285\times10^{-3}$
& $16$--$96$ / $1$--$3$
& $0$--$0.3$ / $4$--$16$
& $7$--$28$
& Chebyshev order $2$--$4$; temporal kernel $3$--$5$ \\

DCRNN
& $1.3736\times10^{-4}$--$4.8628\times10^{-3}$
& $16$--$96$ / $1$--$3$
& $0$--$0.3$ / $4$--$16$
& $7$--$28$
& diffusion step $1$--$3$ \\

AGCRN
& $1.0276\times10^{-4}$--$3.2996\times10^{-3}$
& $16$--$96$ / $1$--$3$
& $0$--$0.3$ / $4$--$16$
& $7$--$28$
& embedding dimension $8$--$32$; Chebyshev order $2$--$4$ \\

Graph WaveNet
& $2.5721\times10^{-4}$--$3.0792\times10^{-3}$
& $16$--$96$ / $1$--$3$
& $0$--$0.3$ / $4$--$16$
& $7$--$28$
& adaptive graph propagation with dilated temporal convolution \\

MTGNN
& $2.8603\times10^{-4}$--$3.5913\times10^{-3}$
& $16$--$96$ / $1$--$3$
& $0.1$--$0.3$ / $4$--$8$
& $7$--$28$
& embedding dimension $8$--$32$; graph depth $1$--$3$ \\

\bottomrule
\end{tabularx}
\end{sidewaystable}

\begin{sidewaystable}[p]
\centering
\caption{\textbf{Real-world baseline configurations.} Final model architectures and hyperparameters for Ontario, Hubei and Spain.}
\label{tab:real-baseline-configurations}
\scriptsize
\renewcommand{\arraystretch}{2.0}
\setlength{\tabcolsep}{2.8pt}

\begin{tabularx}{\textheight}{
@{}p{2.0cm}
p{4.3cm}
p{3.3cm}
p{4.3cm}
p{2.4cm}
X@{}}
\toprule
\textbf{Model}
& \textbf{Learning rate}
& \textbf{Hidden/layers}
& \textbf{Dropout/batch}
& \textbf{Input window}
& \textbf{Model-specific setting} \\
\midrule

MCM-SEIR
& 0.014/0.005/0.020
& N/A
& N/A
& N/A
& $\beta_{\mathrm{init}}=0.104/0.080/0.300$; 
$\gamma_{\mathrm{init}}=0.13/0.17/0.08$ \\

DCRNN
& 0.0085/0.004/0.001056
& 16/1; 32/1; 16/1
& 0.35/8; 0/8; 0.277/8
& 10/3/10
& diffusion step 1/2/3 \\

AGCRN
& 0.0001349/0.003133/0.002033
& 64/1; 64/1; 32/3
& 0.0258/8; 0.387/8; 0.183/8
& 10/3/10
& embedding 8/16/32;
Chebyshev order 2/2/3 \\

MTGNN
& 0.003008/0.009526/0.0001996
& 16/1; 32/2; 32/1
& 0.252/8; 0.153/8; 0.133/8
& 14/5/7
& embedding 16/16/32;
graph depth 1/2/1 \\

EpiGNN
& 0.0009058/0.003611/0.002874
& 64/3; 64/3; 32/1
& 0.219/8; 0.0332/8; 0.238/8
& 18/5/13
& epidemic-aware adaptive adjacency with a direct 7-day head \\

Cola-GNN
& 0.0001962/0.003817/0.002995
& 16/3; 32/1; 16/2
& 0.106/8; 0.382/8; 0.116/8
& 10/7/7
& recurrent temporal encoder and dynamic correlation graph \\

CovidGNN
& 0.0002901/0.005485/0.006149
& 32/1; 64/3; 64/3
& 0.146/8; 0.389/8; 0.186/8
& 10/7/10
& causal windowed spatio-temporal graph construction \\

MepoGNN
& 0.005707/0.001907/0.001042
& 16/2; 64/2; 16/2
& 0.0691/8; 0.338/4; 0.0926/8
& 7/3/5
& absolute/anchored/absolute physical readout \\

M-Graphormer
& 0.002/0.0007216/0.001
& 16/2; 32/3; 16/2
& 0.20/8; 0.0162/4; 0.05/4
& 10/3/7
& attention heads 4/4/2;
Graphormer blocks 2/1/3 \\

EpiGCN
& 0.004/0.002/0.004
& 16/2; 32/3; 32/2
& 0.10/8; 0.05/4; 0.10/8
& 10/3/7
& ODE substeps 2/4/2;
anchored node-calibrated readout \\

STAN
& 0.003/0.003/0.001
& 16/2; 32/2; 32/2
& 0.05/8; 0.05/8; 0.10/4
& 10/7/7
& GAT heads 1/2/2;
GAT hidden 16/32/32 \\

EINN
& 0.003290/0.002/0.002
& 16/3; 32/2; 32/2
& 0.152/4; 0.05/4; 0.05/4
& 14/3/7
& Fourier features 20/12/12;
ODE substeps 1/2/2 \\

Epi-Cola-GNN
& 0.006/0.003/0.006
& 16/1; 32/1; 32/1
& 0/8; 0.05/8; 0/8
& 14/3/10
& temporal channels 8/16/8 \\

EARTH
& 0.004/0.003/0.001
& 64/1; 32/1; 32/1
& 0/8; 0/8; 0.10/4
& 14/3/7
& epidemic ODE substeps 2/1/1 \\

\bottomrule
\end{tabularx}
\end{sidewaystable}

\FloatBarrier

\section{Supplementary Results}
\newcommand{\EdgeTrajectoryPair}[6]{%
\begin{figure}[!htbp]
    \centering
    \begin{subfigure}[t]{0.96\textwidth}
        \centering
        \includegraphics[width=\linewidth,height=0.35\textheight,keepaspectratio]{#1}
        \caption{#2}
    \end{subfigure}

    \vspace{0.30em}

    \begin{subfigure}[t]{0.96\textwidth}
        \centering
        \includegraphics[width=\linewidth,height=0.35\textheight,keepaspectratio]{#3}
        \caption{#4}
    \end{subfigure}

    \caption{ \textbf{#5} Blue solid lines denote the reference generating operators, and red dashed lines denote the inferred operators. }
    \label{#6}
\end{figure}%
}

\newcommand{\MixedTrajectoryPanel}[2]{%
    \begin{subfigure}[t]{0.485\linewidth}
        \centering
        \includegraphics[
            width=\linewidth,
            keepaspectratio
        ]{#1}
        \caption{#2}
    \end{subfigure}%
}

\newcommand{\MixedOperatorFigure}[5]{%
\begin{sidewaysfigure}[p]
    \centering
    \setcounter{subfigure}{0}

    \MixedTrajectoryPanel
        {#1_1.pdf}
        {#3 1.}
    \hfill
    \MixedTrajectoryPanel
        {#1_2.pdf}
        {#3 2.}

    \vspace{0.6em}

    \MixedTrajectoryPanel
        {#1_3.pdf}
        {#3 3.}
    \hfill
    \MixedTrajectoryPanel
        {#1_4.pdf}
        {#3 4.}

    \caption{ \textbf{Complete ground-truth and inferred \(#4(t)\) trajectories for #2 in the multiscale scenario.} Blue solid lines denote the reference generating operators, and red dashed lines denote the inferred operators. }
    \label{#5}
\end{sidewaysfigure}
}

\subsection{Dynamical Fidelity: Complete Operator Trajectories}
\label{app:operator-trajectories}

This section provides the complete reference and inferred trajectories of the
instantaneous Lagrangian operator $\FL(t)$ and Eulerian operator $\FE(t)$ across
the synthetic generation mechanisms. Blue solid lines denote the reference
generating operators, and red dashed lines denote the inferred operators. The
trajectories are organized as $4\times4$ grids of directed-edge interactions.
In the Lagrangian and Eulerian scenarios, each subfigure shows one grid for a
specified mechanism; in the multiscale scenarios, each figure contains four grids
corresponding to Regions 1--4 for $\FL(t)$ or Age groups 1--4 for $\FE(t)$.
The $\FL(t)$ grids represent directed infection-generation interactions among age
groups within a region, whereas the $\FE(t)$ grids represent directed
population-transport interactions among regions for a given age group.

Overall, the inferred operators reproduce the principal temporal patterns and
relative edge structure of the reference dynamics. Agreement is generally stronger
for $\FE(t)$, for which the major state-dependent and time-modulated transport trends
are closely tracked. The inferred $\FL(t)$ trajectories also capture the characteristic
evolution of the different infection-generation mechanisms, but larger discrepancies
in magnitude or turning points are more apparent during the early stage and in rapidly
varying segments. These deviations may arise because early epidemic observations
provide weaker constraints on instantaneous infection-generation strengths, while
compensation between latent initial states and interaction strengths, smooth neural
parameterizations, and Lagrangian--Eulerian coupling in the multiscale systems may
also contribute.

\clearpage
\subsubsection{Lagrangian Scenario}

\EdgeTrajectoryPair
    {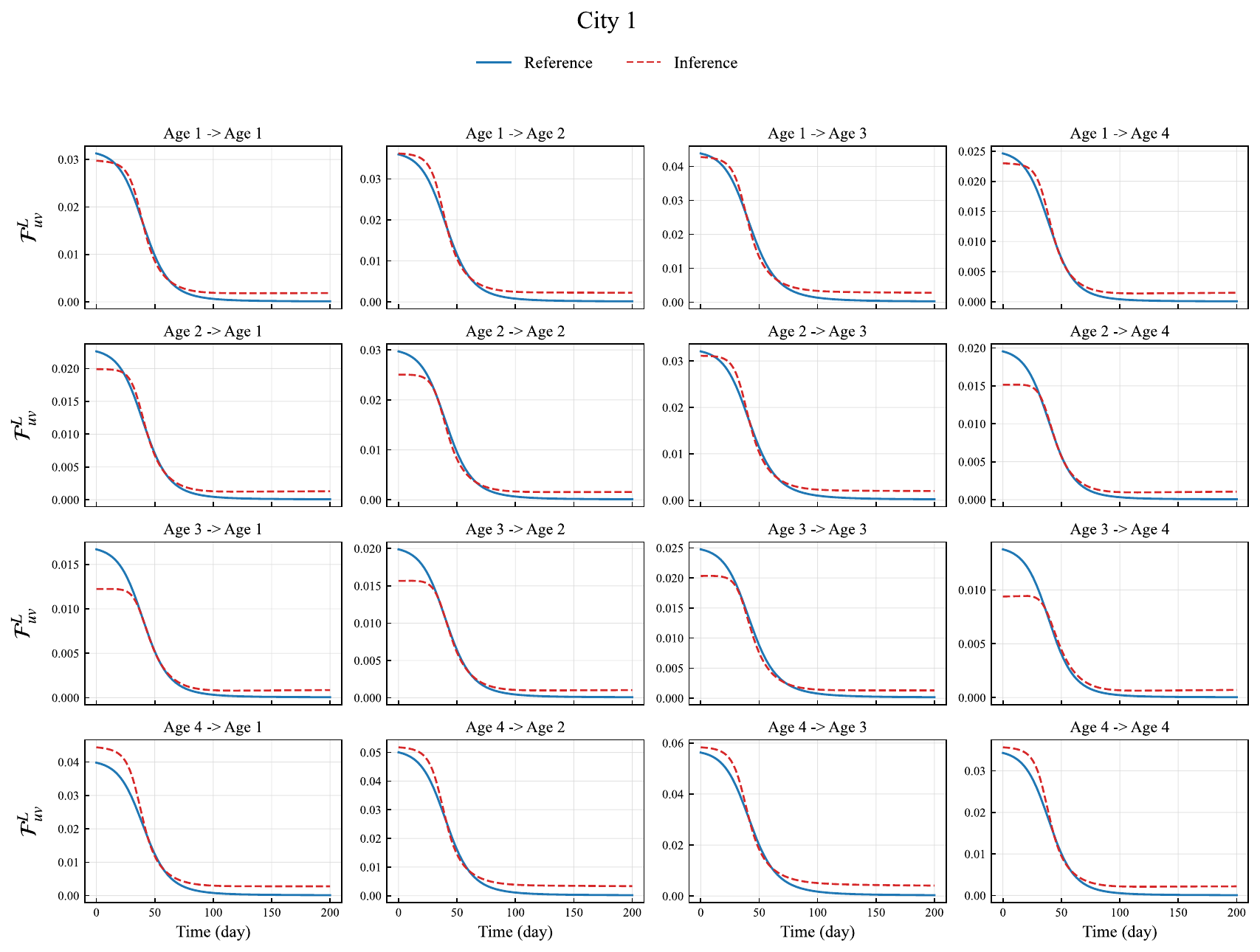}
    {BMCM.}
    {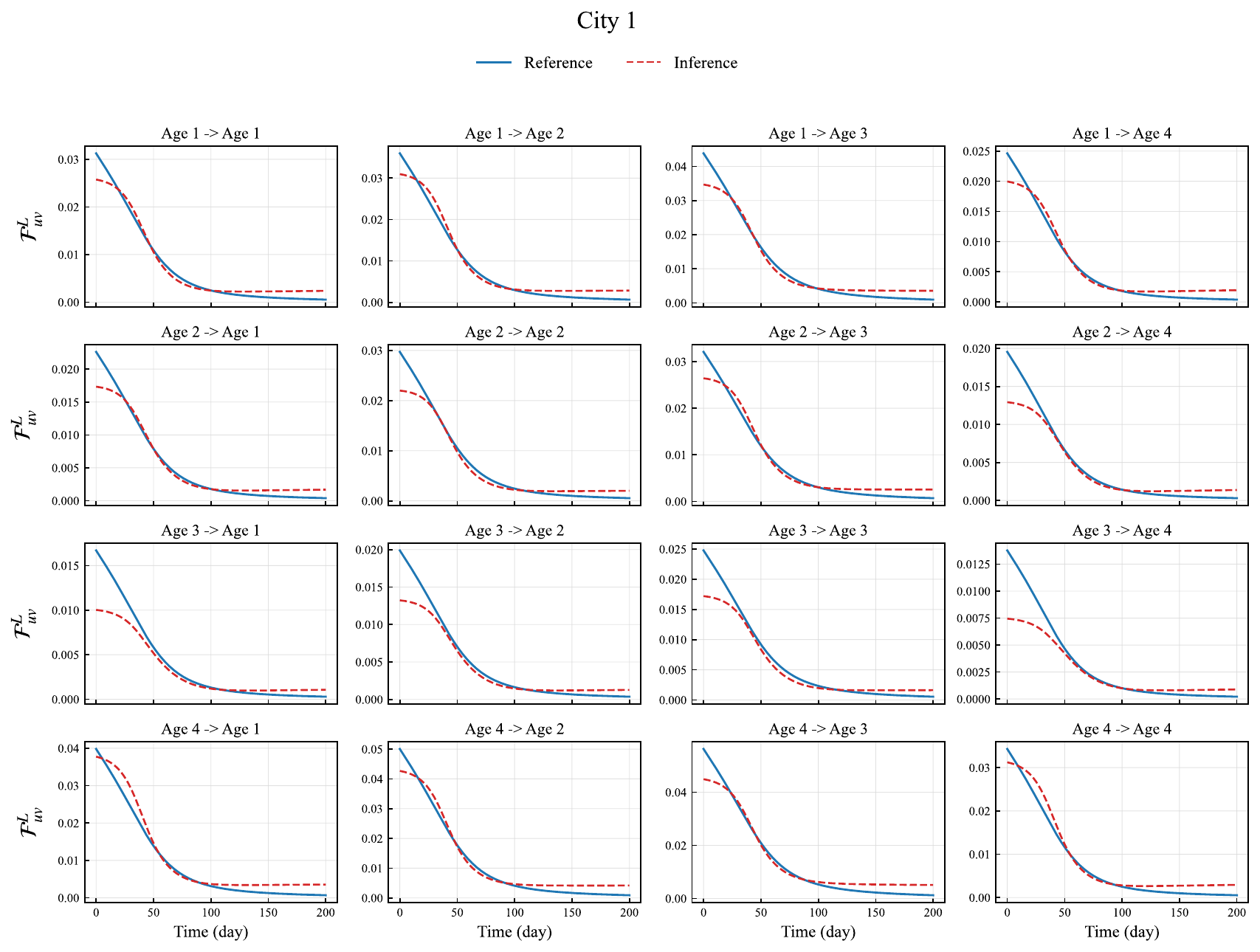}
    {TDCM.}
    {Complete ground-truth and inferred $\FL(t)$ trajectories for BMCM and TDCM in the Lagrangian scenario.}
    {fig:edge-lagrangian-bmcm-tdcm}

\EdgeTrajectoryPair
    {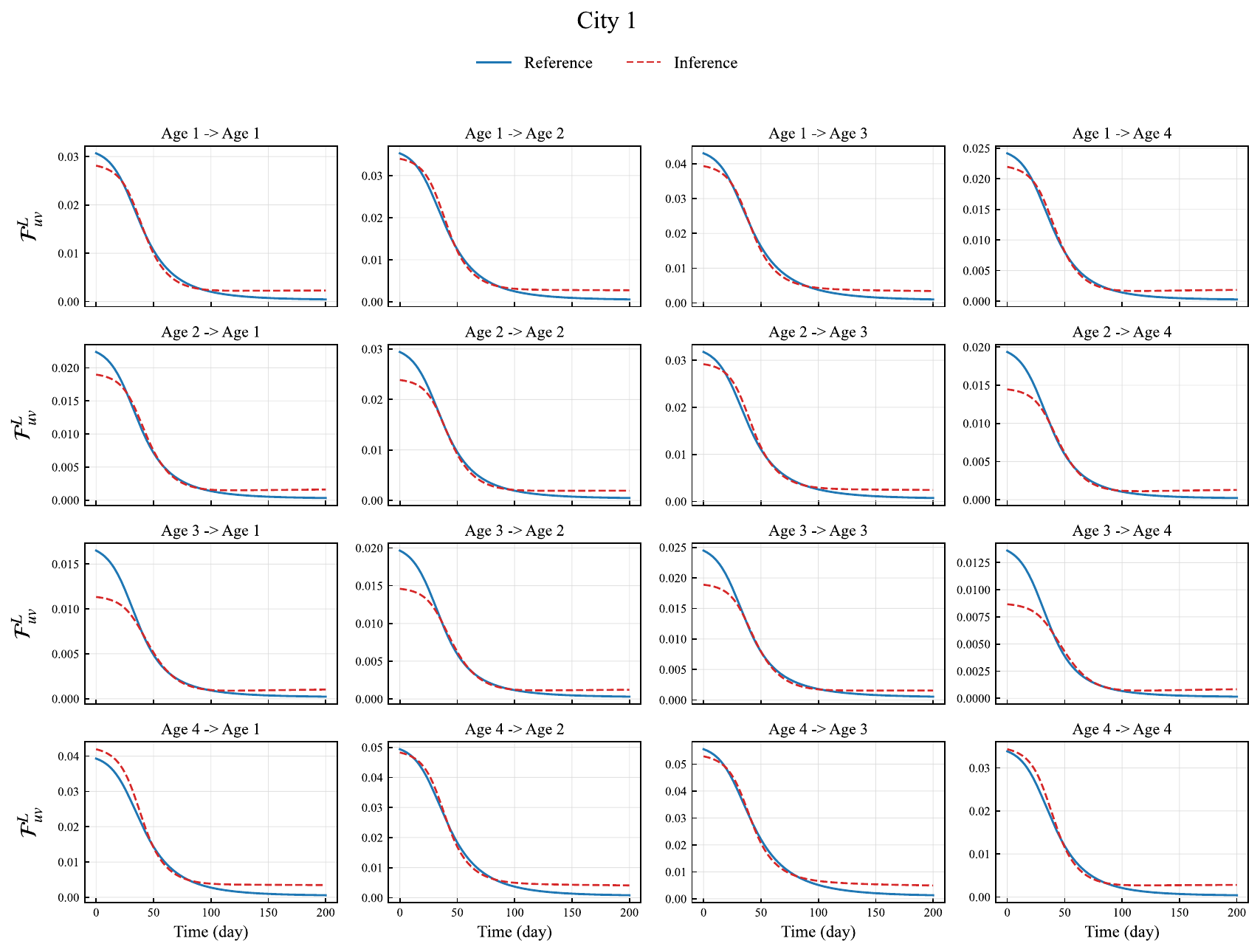}
    {IACM.}
    {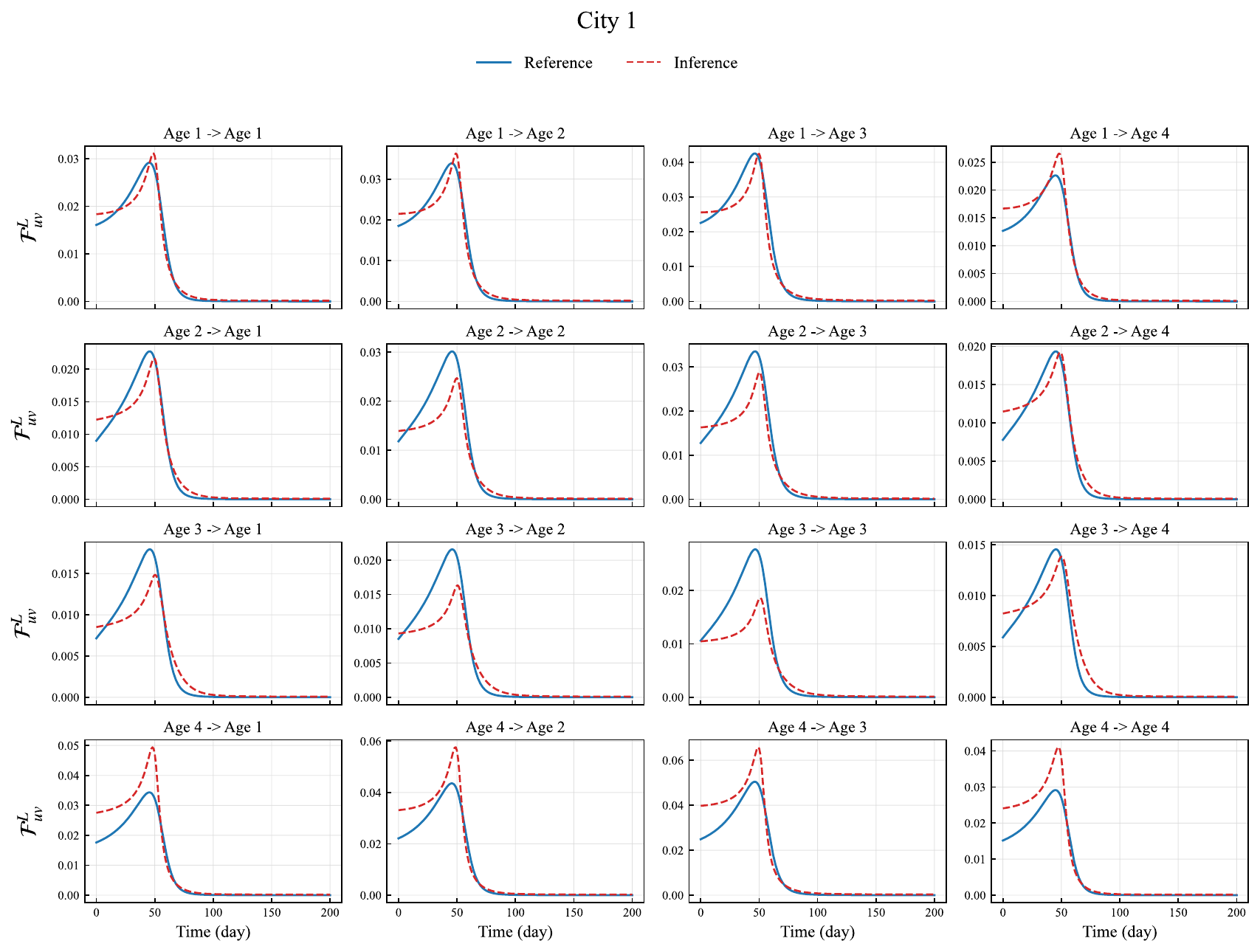}
    {NMCM.}
    {Complete ground-truth and inferred $\FL(t)$ trajectories for IACM and NMCM in the Lagrangian scenario.}
    {fig:edge-lagrangian-iacm-nmcm}

\FloatBarrier
\clearpage

\subsubsection{Eulerian Scenario}

\EdgeTrajectoryPair
    {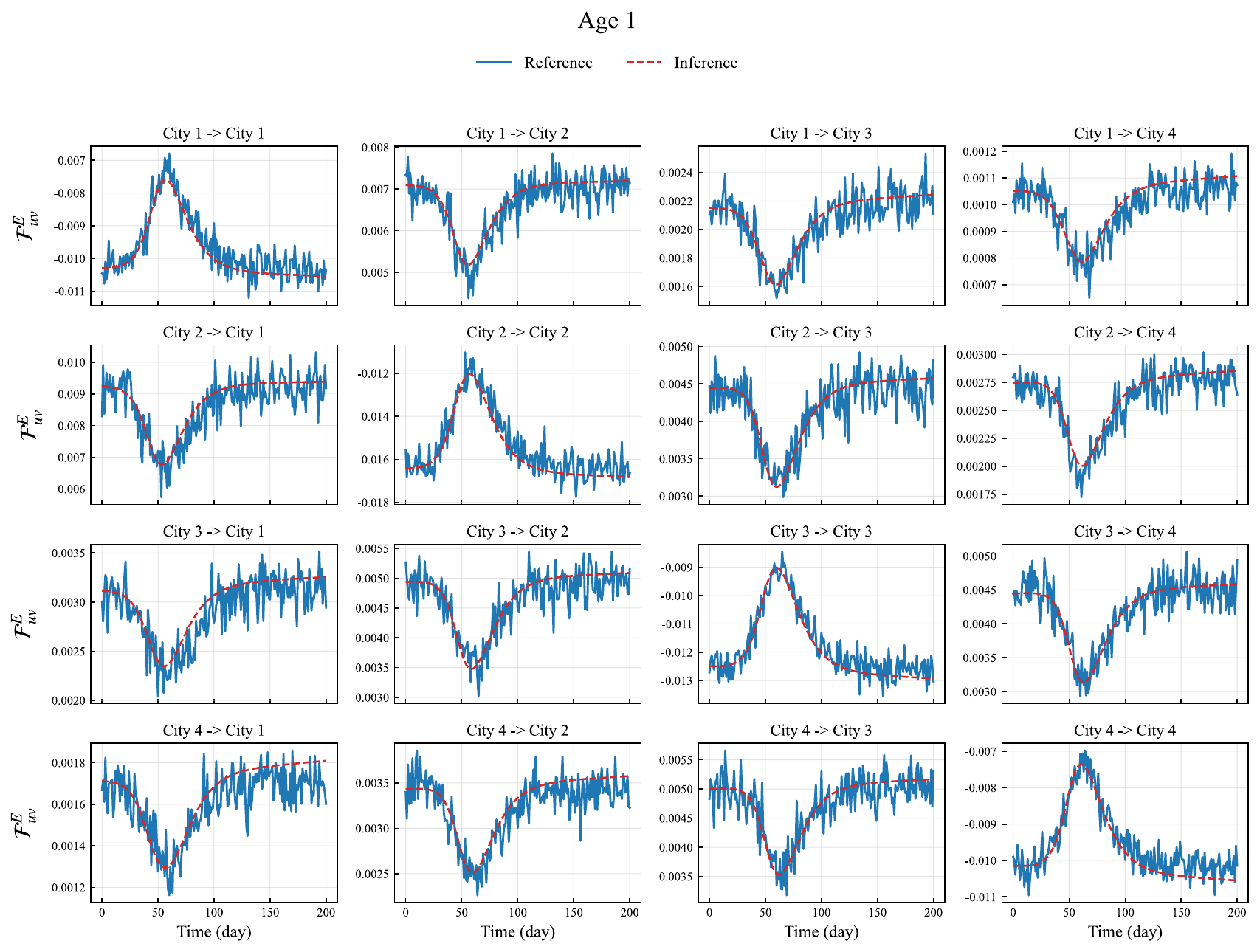}
    {BRGM.}
    {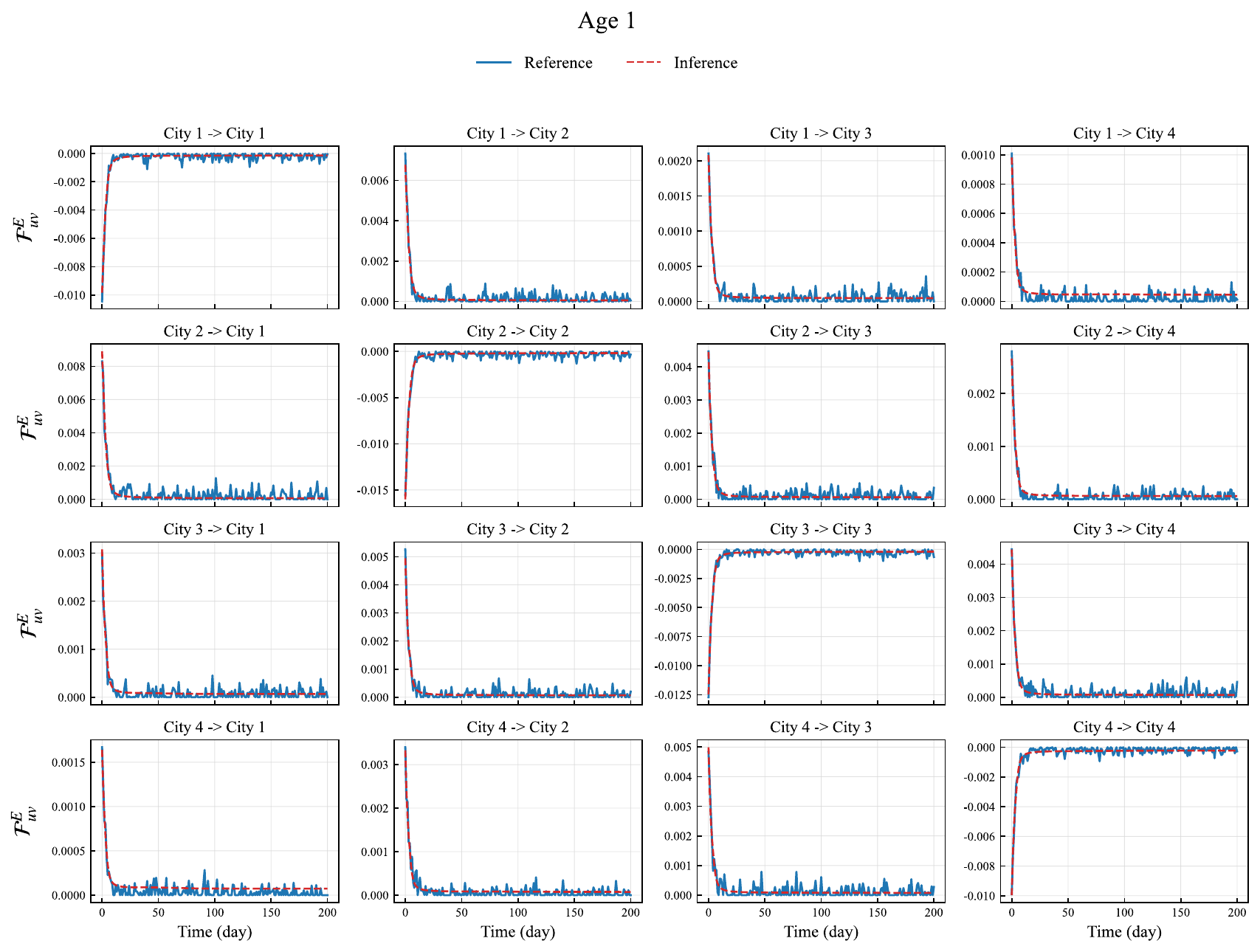}
    {CRGM.}
    {Complete ground-truth and inferred $\FE(t)$ trajectories for BRGM and CRGM in the Eulerian scenario.}
    {fig:edge-eulerian-brgm-crgm}

\FloatBarrier
\clearpage

\subsubsection{Multiscale Scenario}

For each coupled mechanism, the $\FL(t)$ and $\FE(t)$ trajectories are presented as separate four-panel figures. For $\FL(t)$, panels (a)--(d) correspond to Regions 1--4; for $\FE(t)$, panels (a)--(d) correspond to Age groups 1--4. Each operator figure is arranged in a $2\times2$ landscape layout so that all four panels and the figure legend appear on a single page.

\MixedOperatorFigure
    {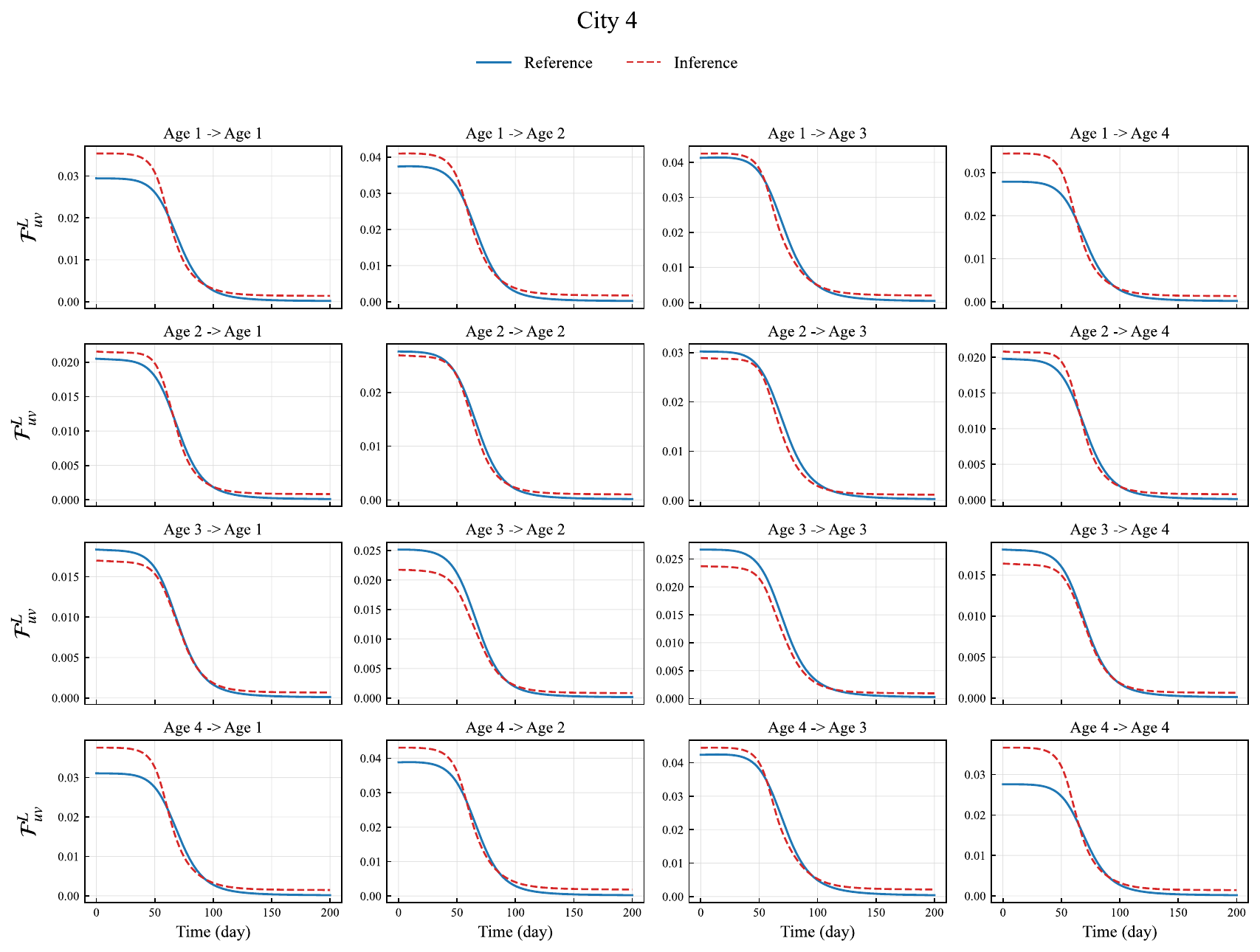}
    {BMCM--BRGM}
    {Region}
    {\FL}
    {fig:edge-mixed-bmcm-brgm-fl-operators}

\MixedOperatorFigure
    {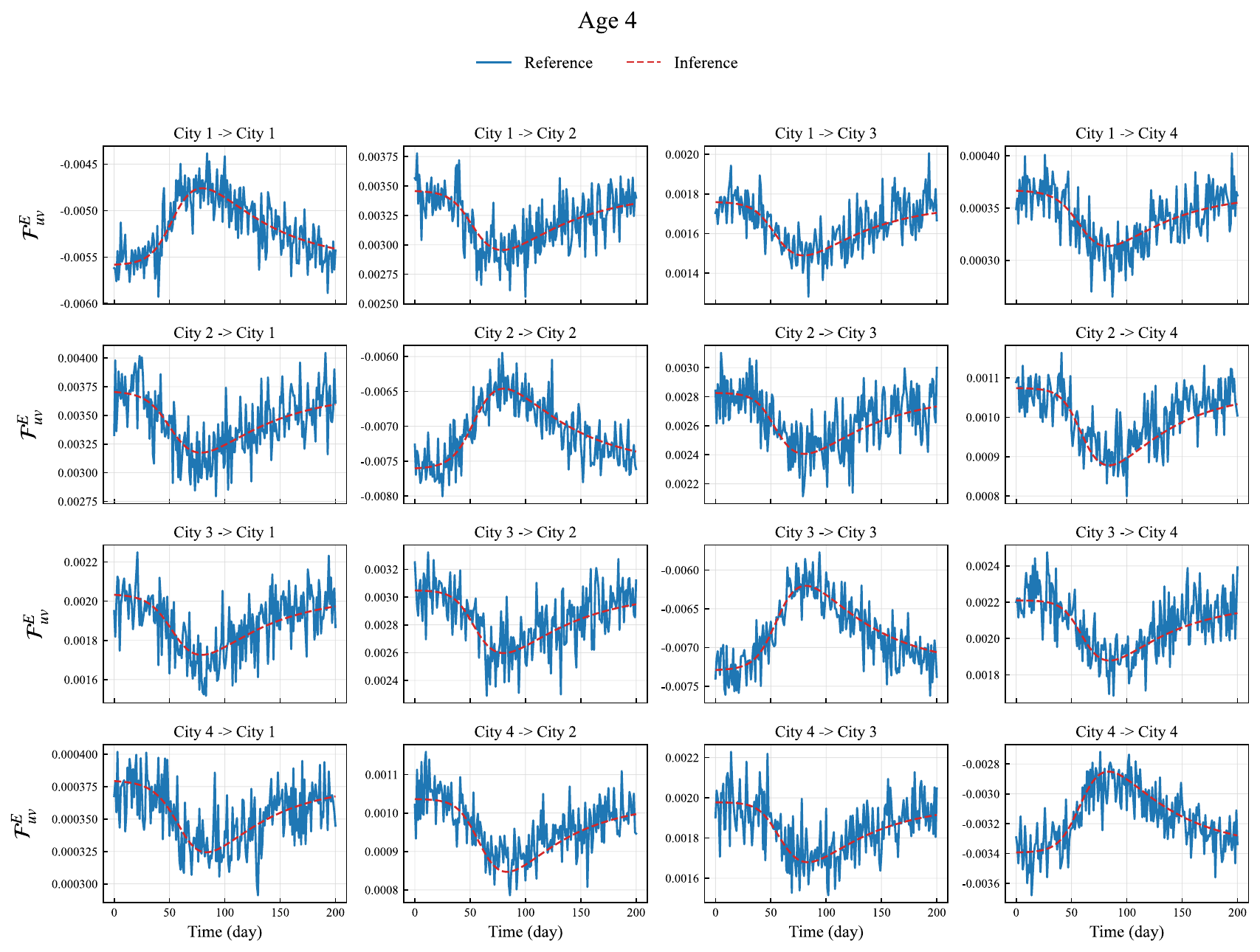}
    {BMCM--BRGM}
    {Age group}
    {\FE}
    {fig:edge-mixed-bmcm-brgm-fe-operators}

\FloatBarrier

\MixedOperatorFigure
    {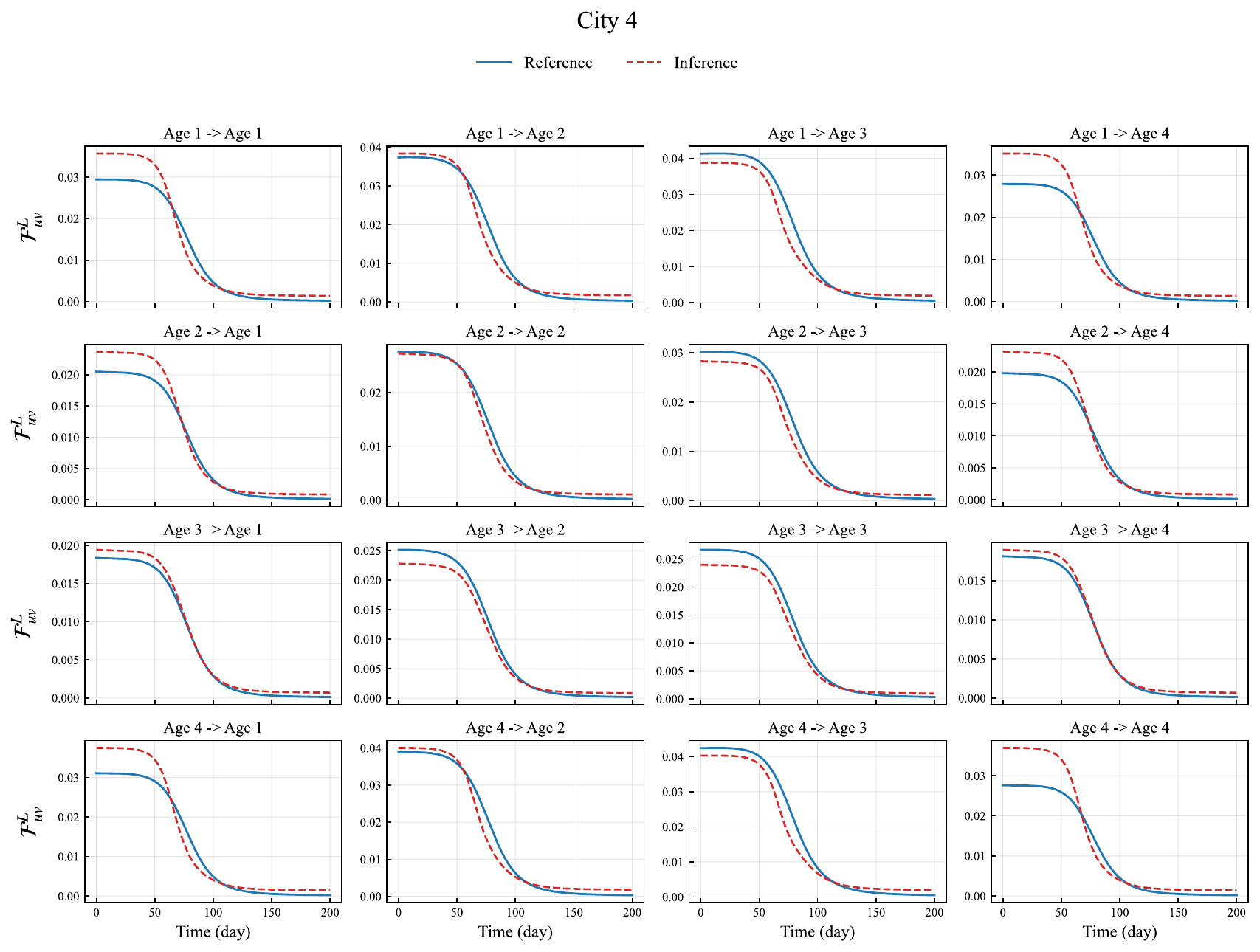}
    {BMCM--CRGM}
    {Region}
    {\FL}
    {fig:edge-mixed-bmcm-crgm-fl-operators}

\MixedOperatorFigure
    {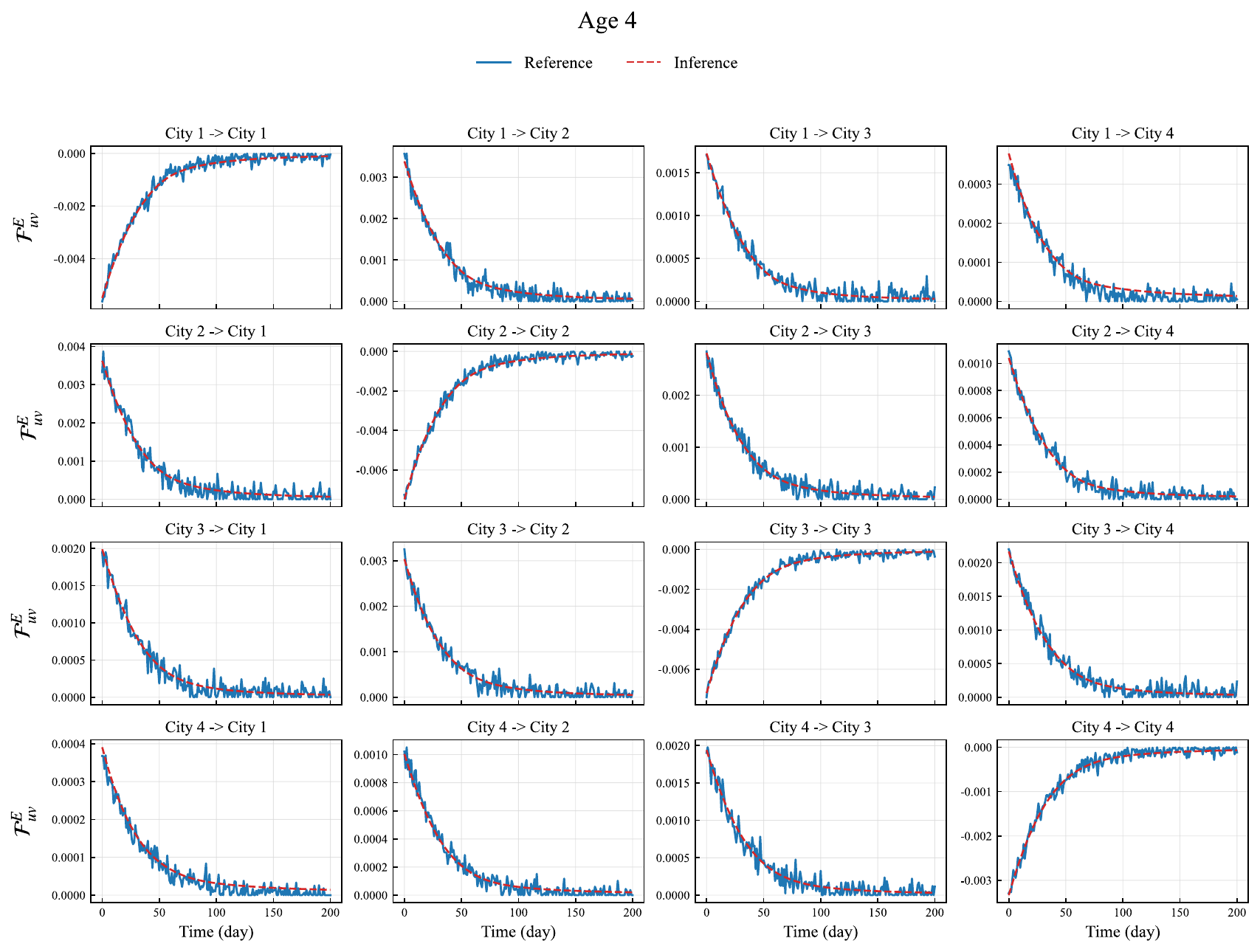}
    {BMCM--CRGM}
    {Age group}
    {\FE}
    {fig:edge-mixed-bmcm-crgm-fe-operators}

\FloatBarrier

\MixedOperatorFigure
    {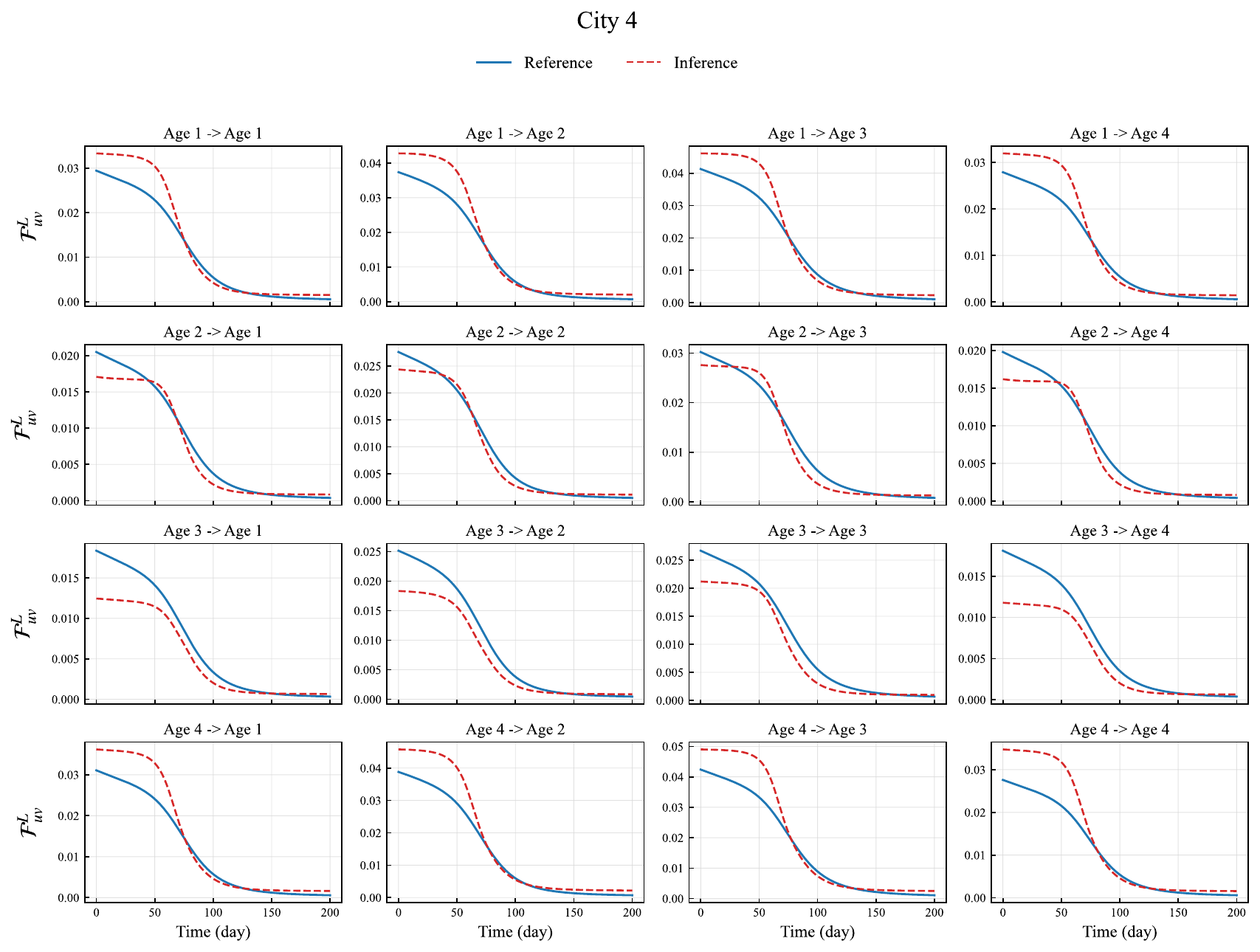}
    {TDCM--BRGM}
    {Region}
    {\FL}
    {fig:edge-mixed-tdcm-brgm-fl-operators}

\MixedOperatorFigure
    {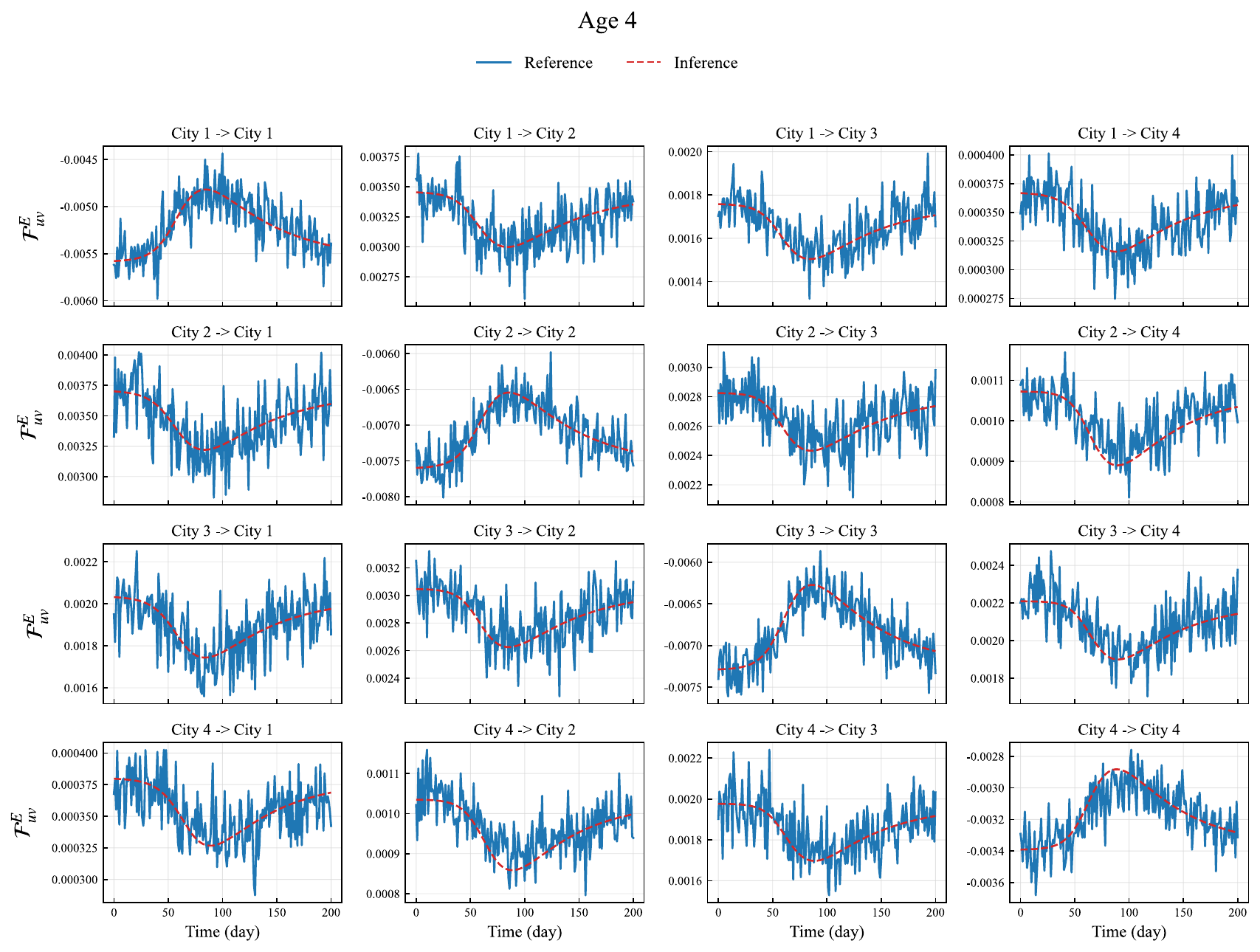}
    {TDCM--BRGM}
    {Age group}
    {\FE}
    {fig:edge-mixed-tdcm-brgm-fe-operators}

\FloatBarrier

\MixedOperatorFigure
    {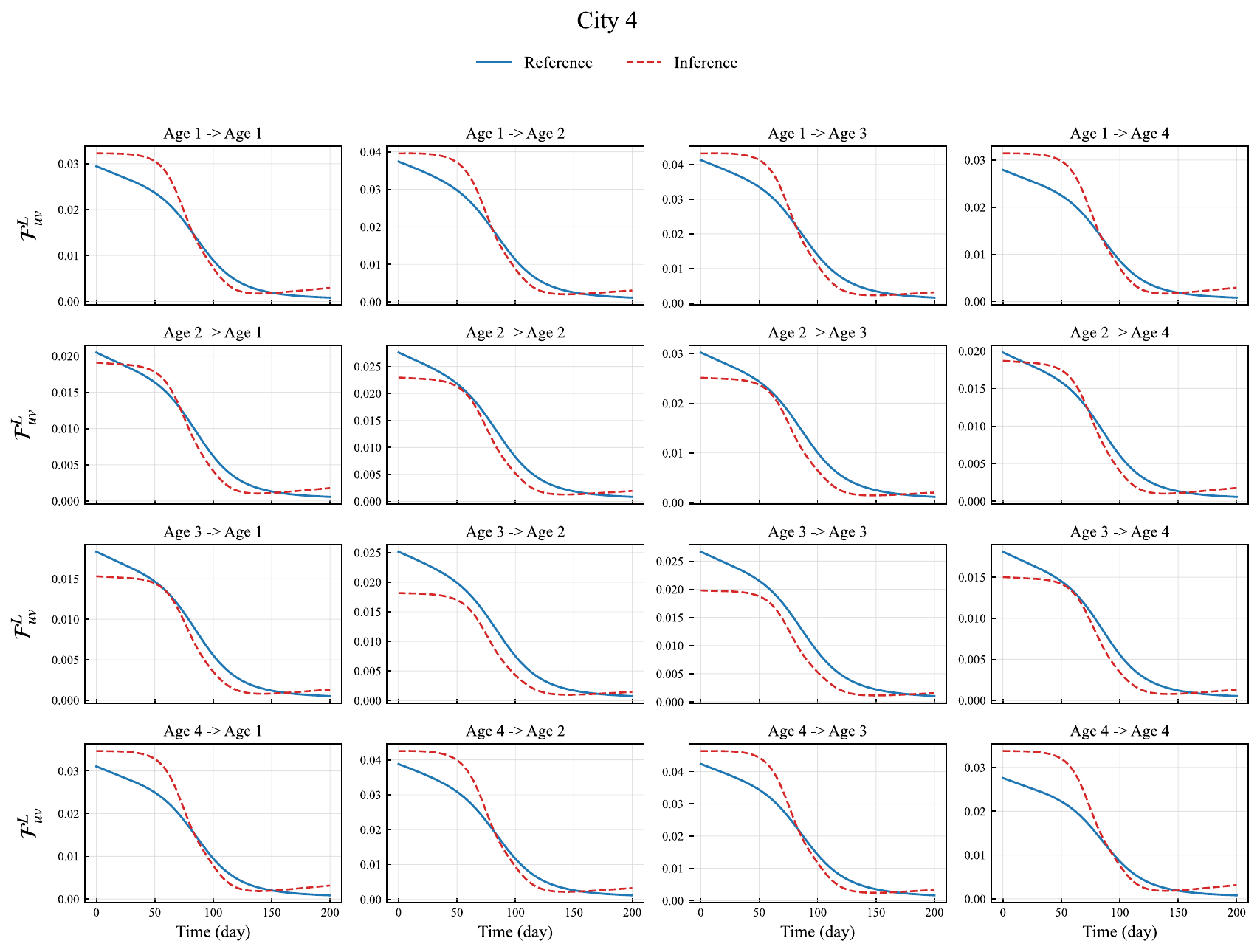}
    {TDCM--CRGM}
    {Region}
    {\FL}
    {fig:edge-mixed-tdcm-crgm-fl-operators}

\MixedOperatorFigure
    {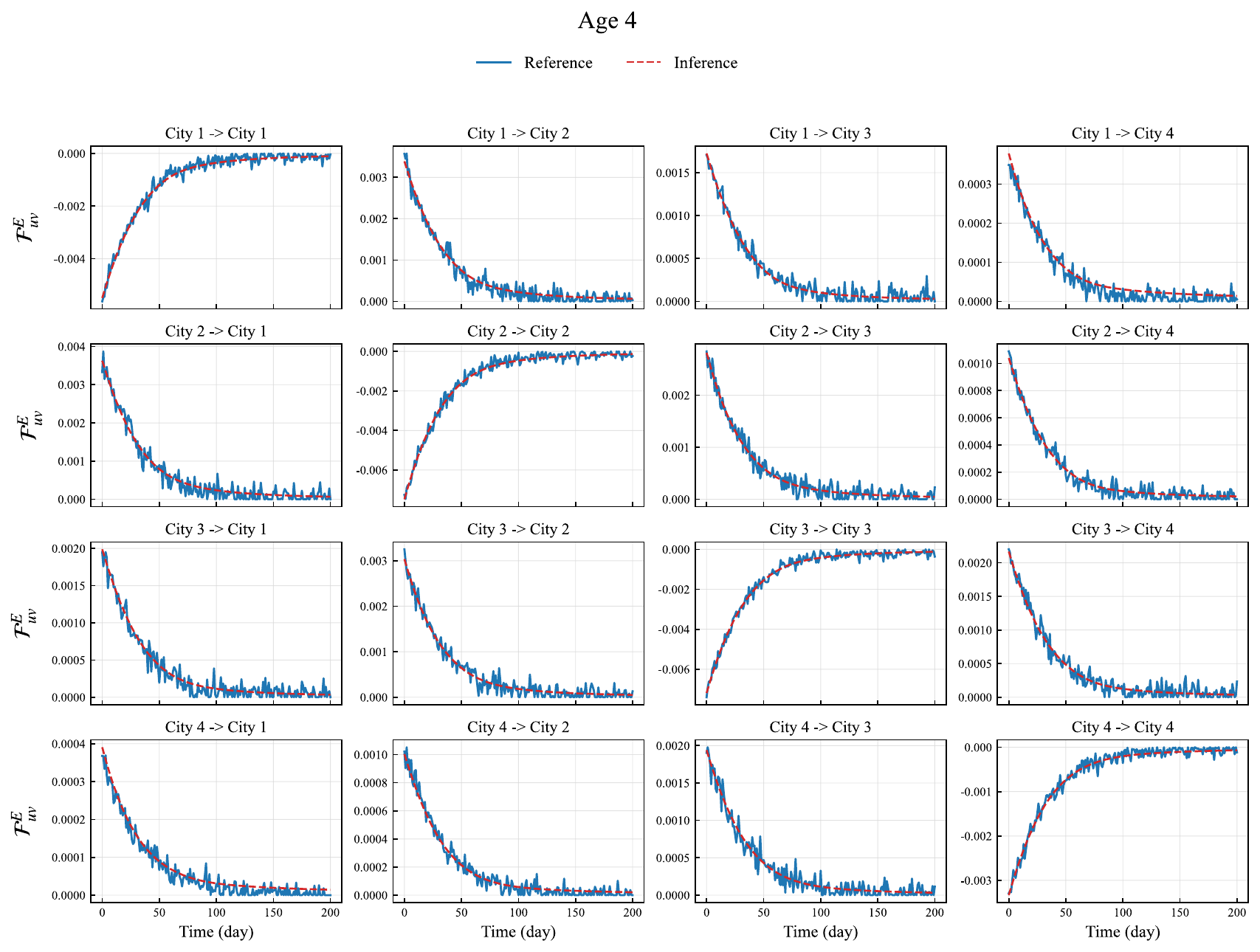}
    {TDCM--CRGM}
    {Age group}
    {\FE}
    {fig:edge-mixed-tdcm-crgm-fe-operators}

\FloatBarrier

\MixedOperatorFigure
    {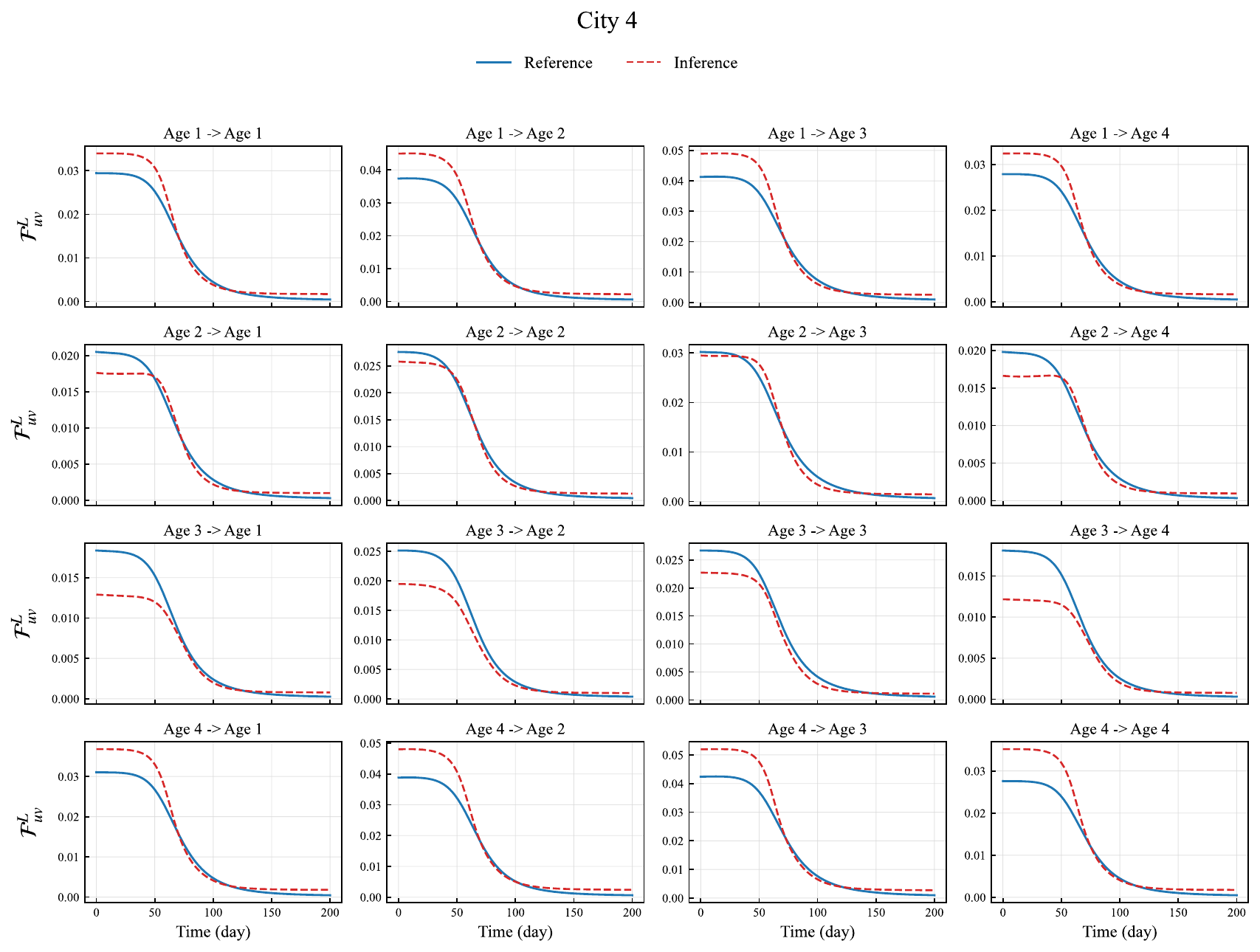}
    {IACM--BRGM}
    {Region}
    {\FL}
    {fig:edge-mixed-iacm-brgm-fl-operators}

\MixedOperatorFigure
    {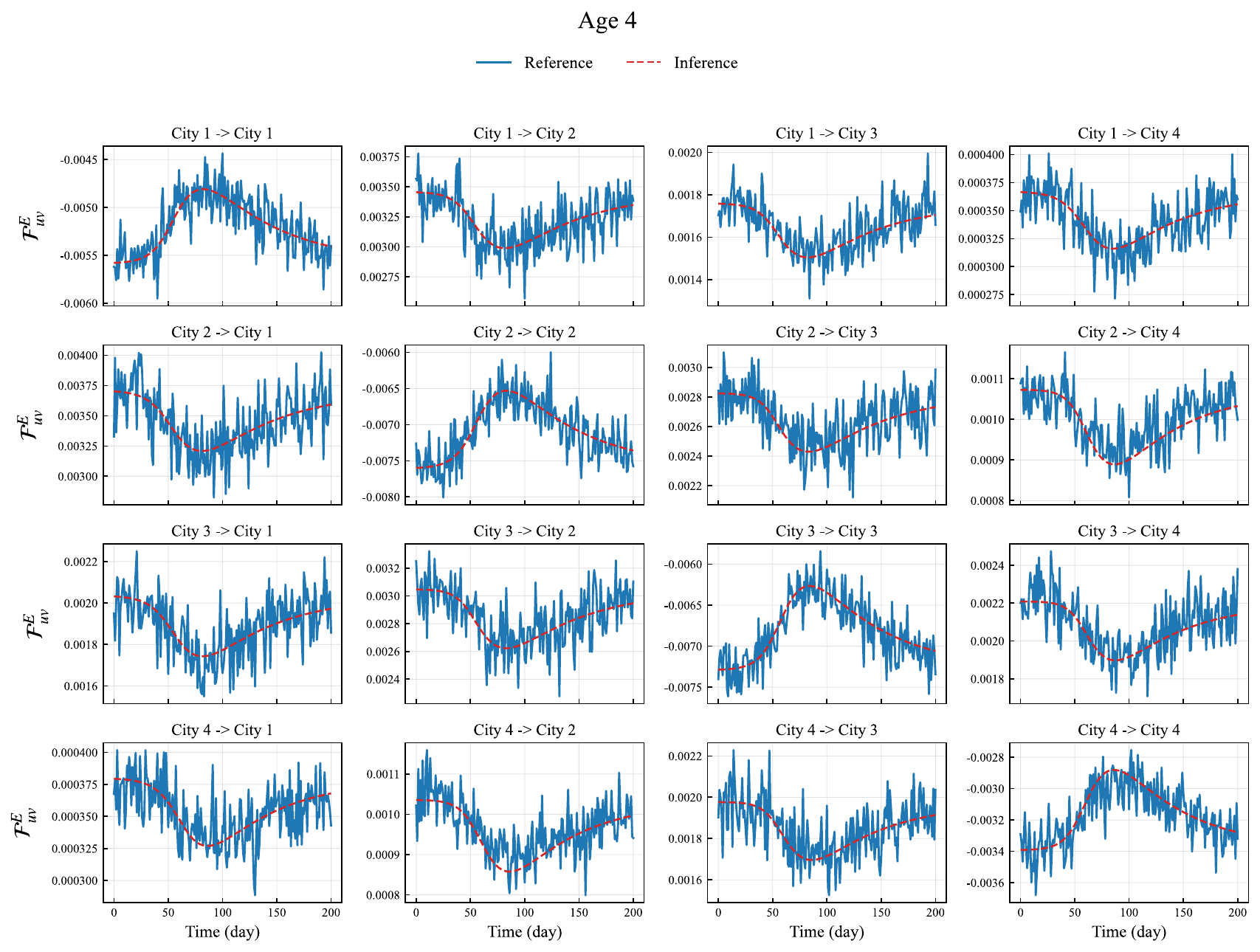}
    {IACM--BRGM}
    {Age group}
    {\FE}
    {fig:edge-mixed-iacm-brgm-fe-operators}

\FloatBarrier

\MixedOperatorFigure
    {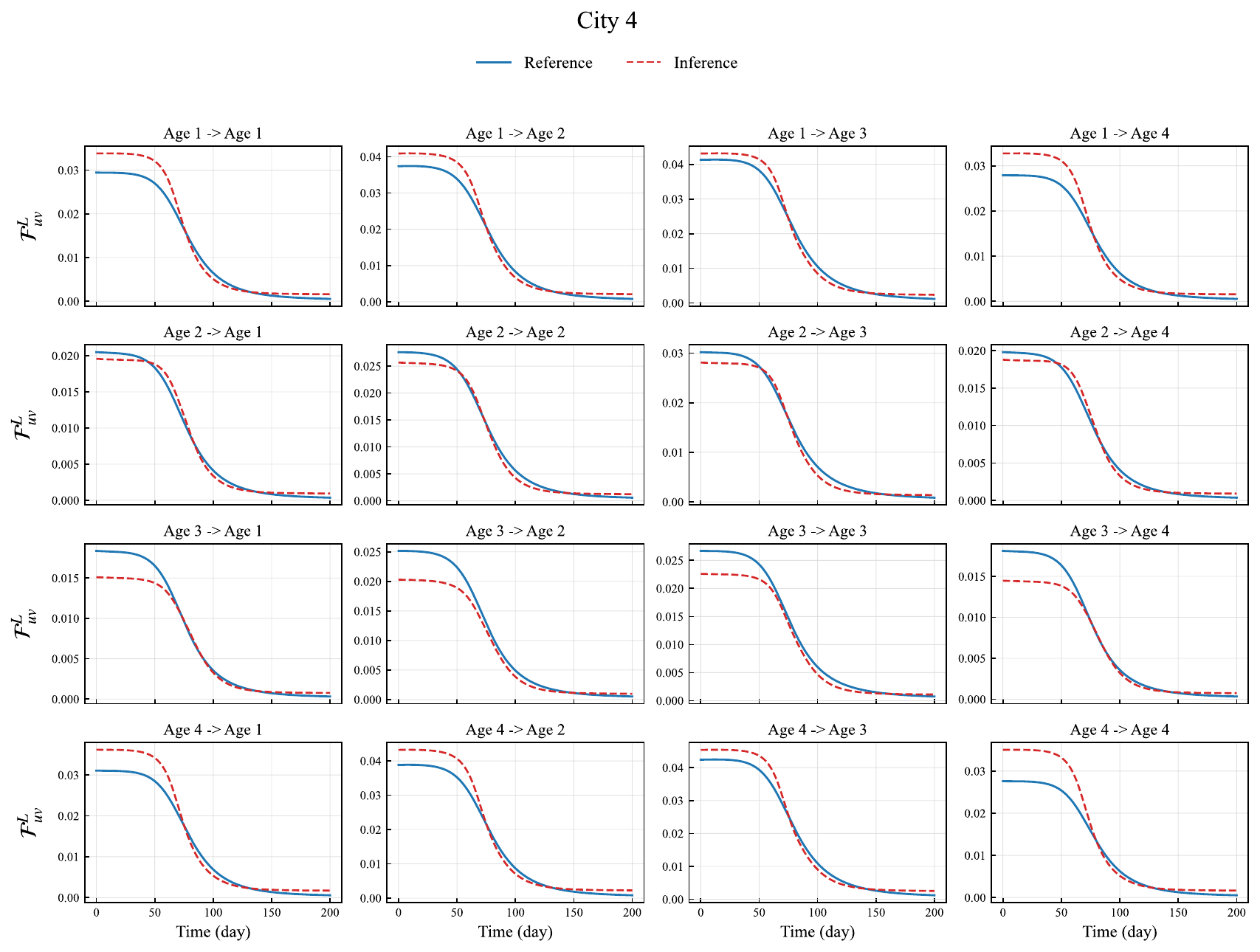}
    {IACM--CRGM}
    {Region}
    {\FL}
    {fig:edge-mixed-iacm-crgm-fl-operators}

\MixedOperatorFigure
    {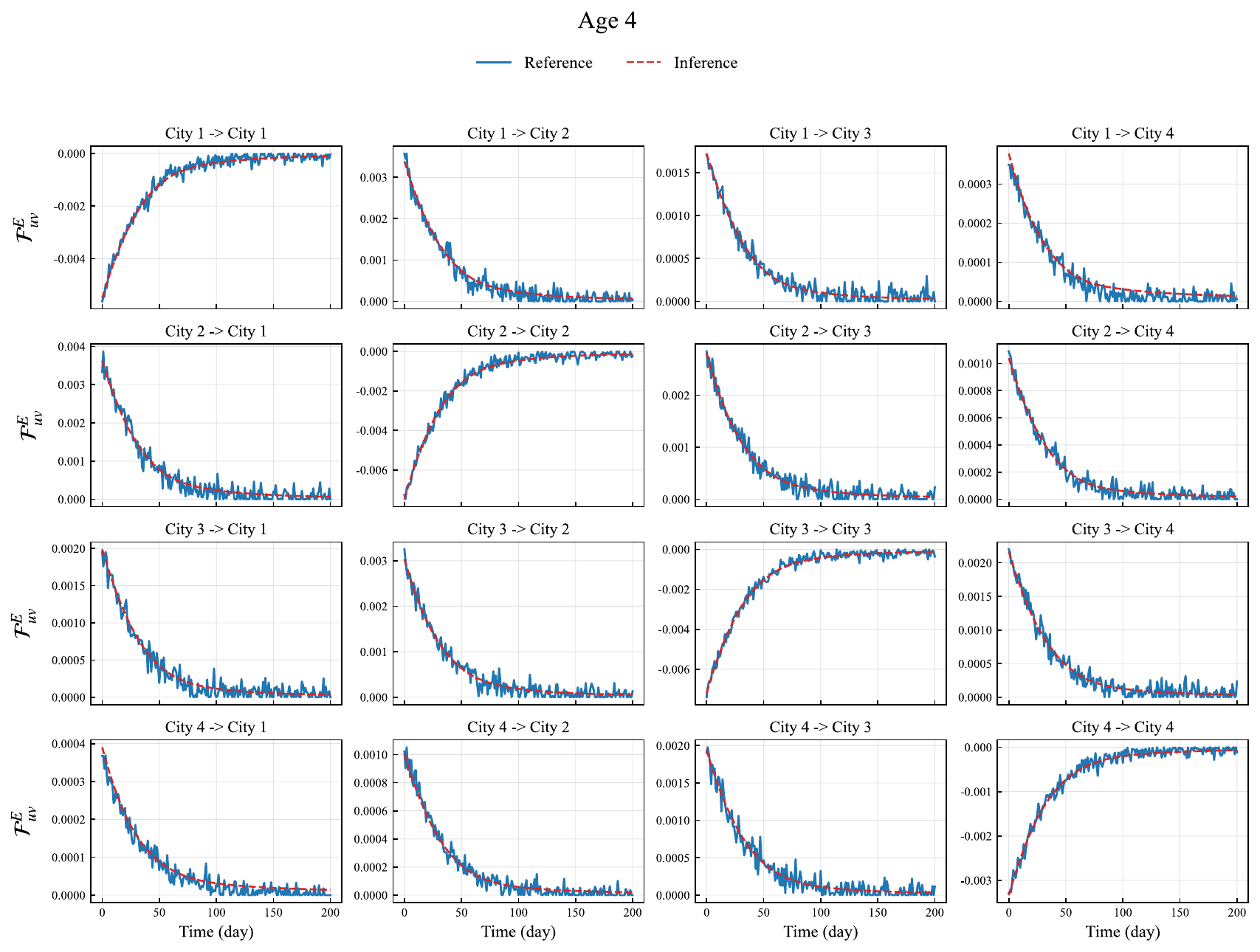}
    {IACM--CRGM}
    {Age group}
    {\FE}
    {fig:edge-mixed-iacm-crgm-fe-operators}

\FloatBarrier

\MixedOperatorFigure
    {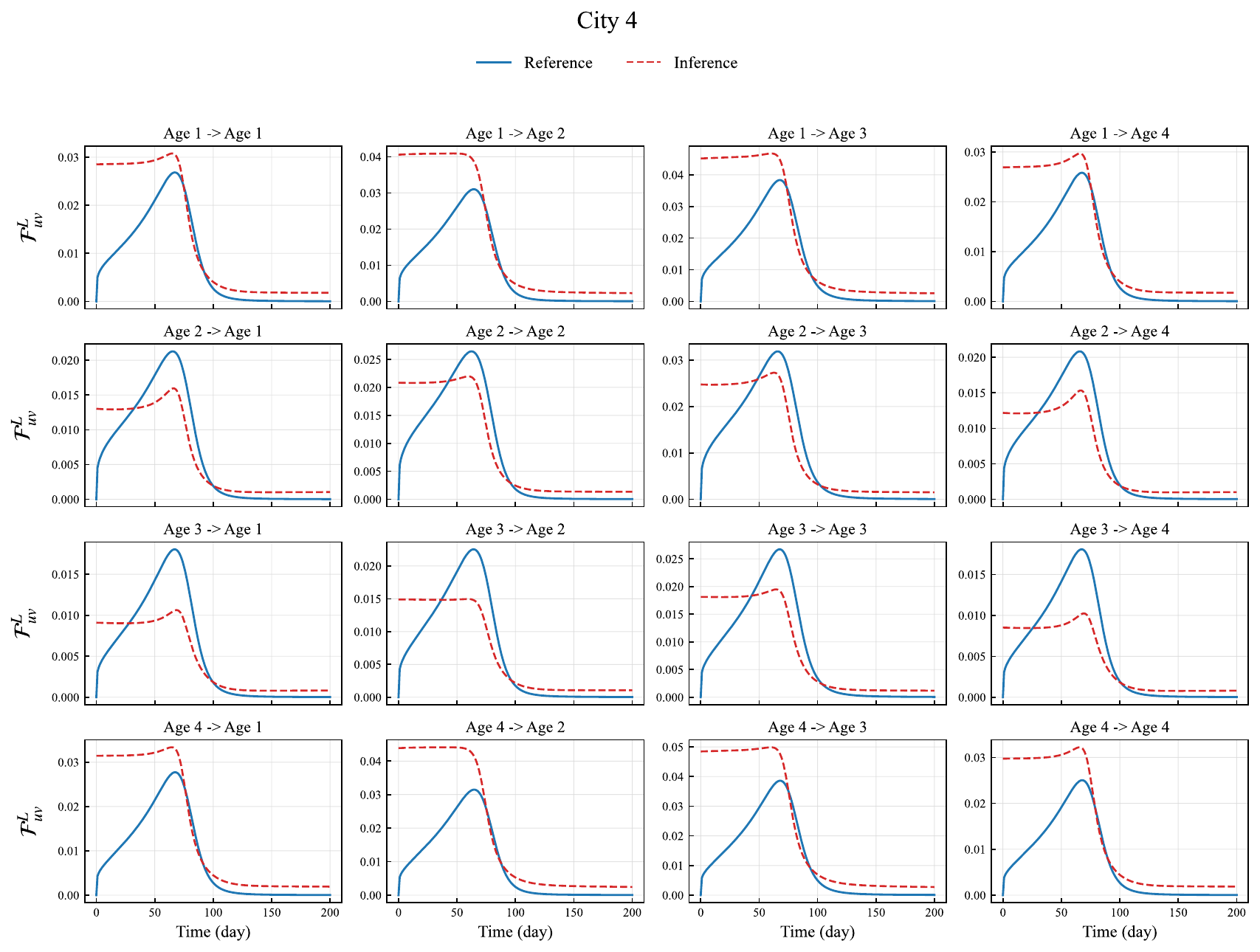}
    {NMCM--BRGM}
    {Region}
    {\FL}
    {fig:edge-mixed-nmcm-brgm-fl-operators}

\MixedOperatorFigure
    {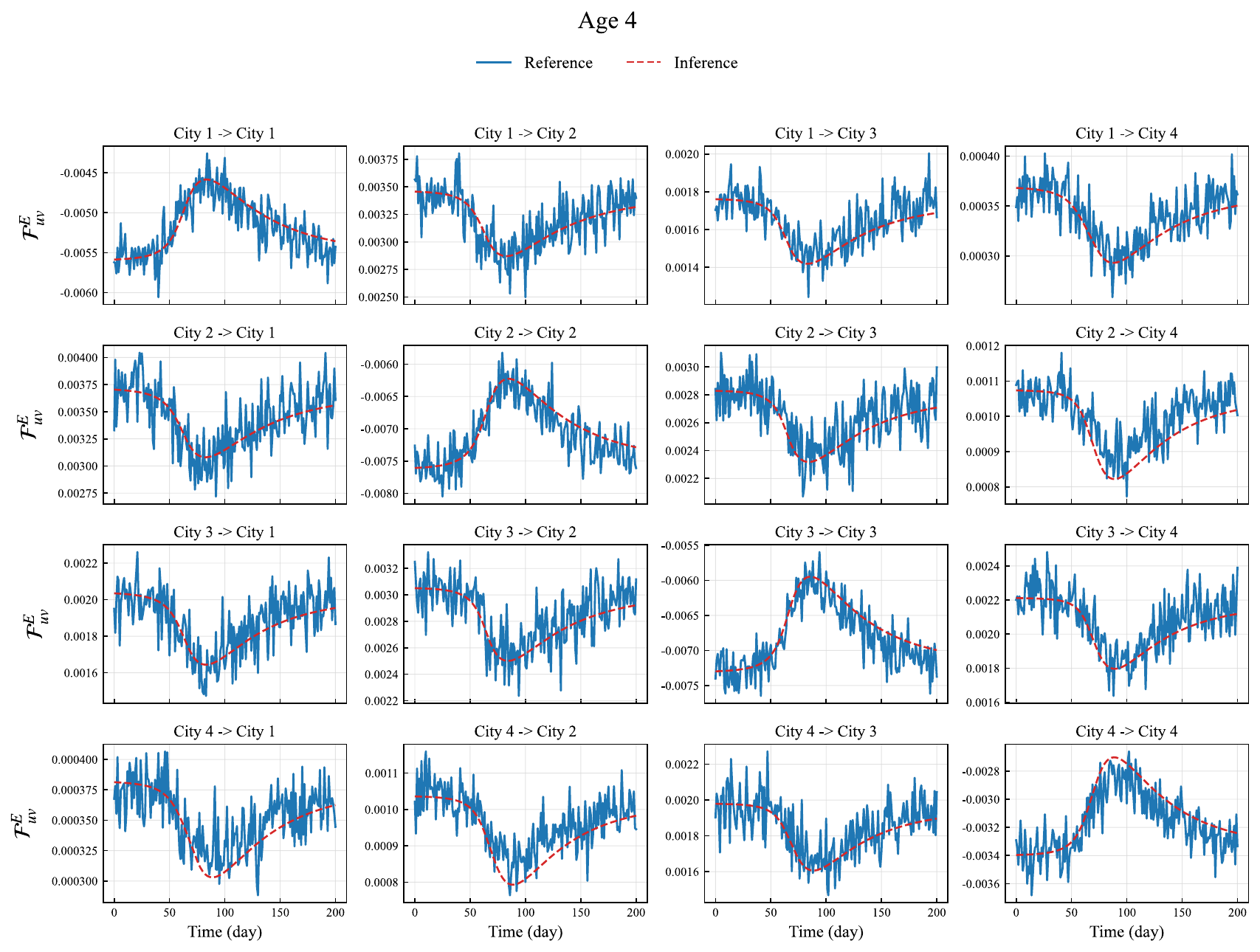}
    {NMCM--BRGM}
    {Age group}
    {\FE}
    {fig:edge-mixed-nmcm-brgm-fe-operators}

\FloatBarrier

\MixedOperatorFigure
    {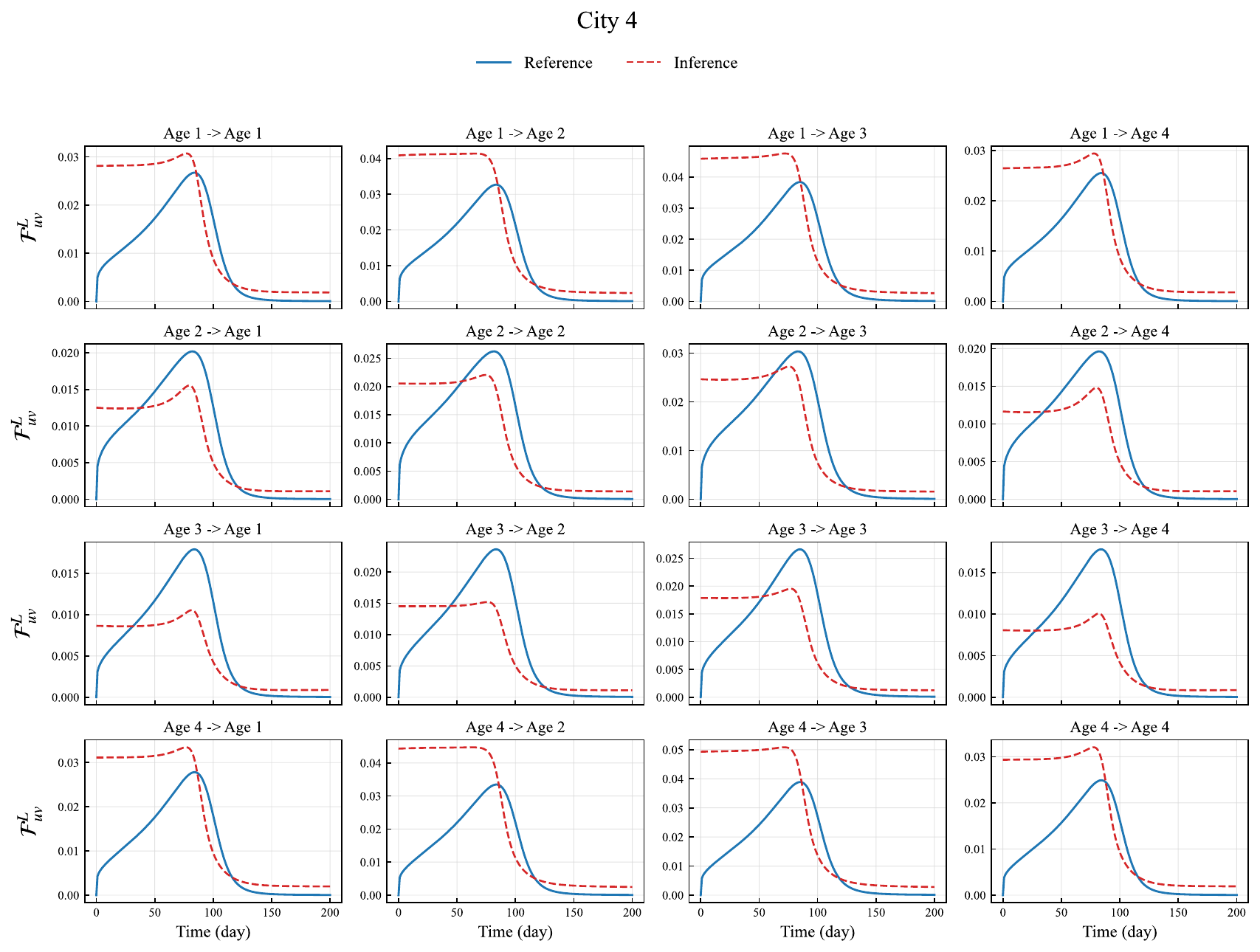}
    {NMCM--CRGM}
    {Region}
    {\FL}
    {fig:edge-mixed-nmcm-crgm-fl-operators}

\MixedOperatorFigure
    {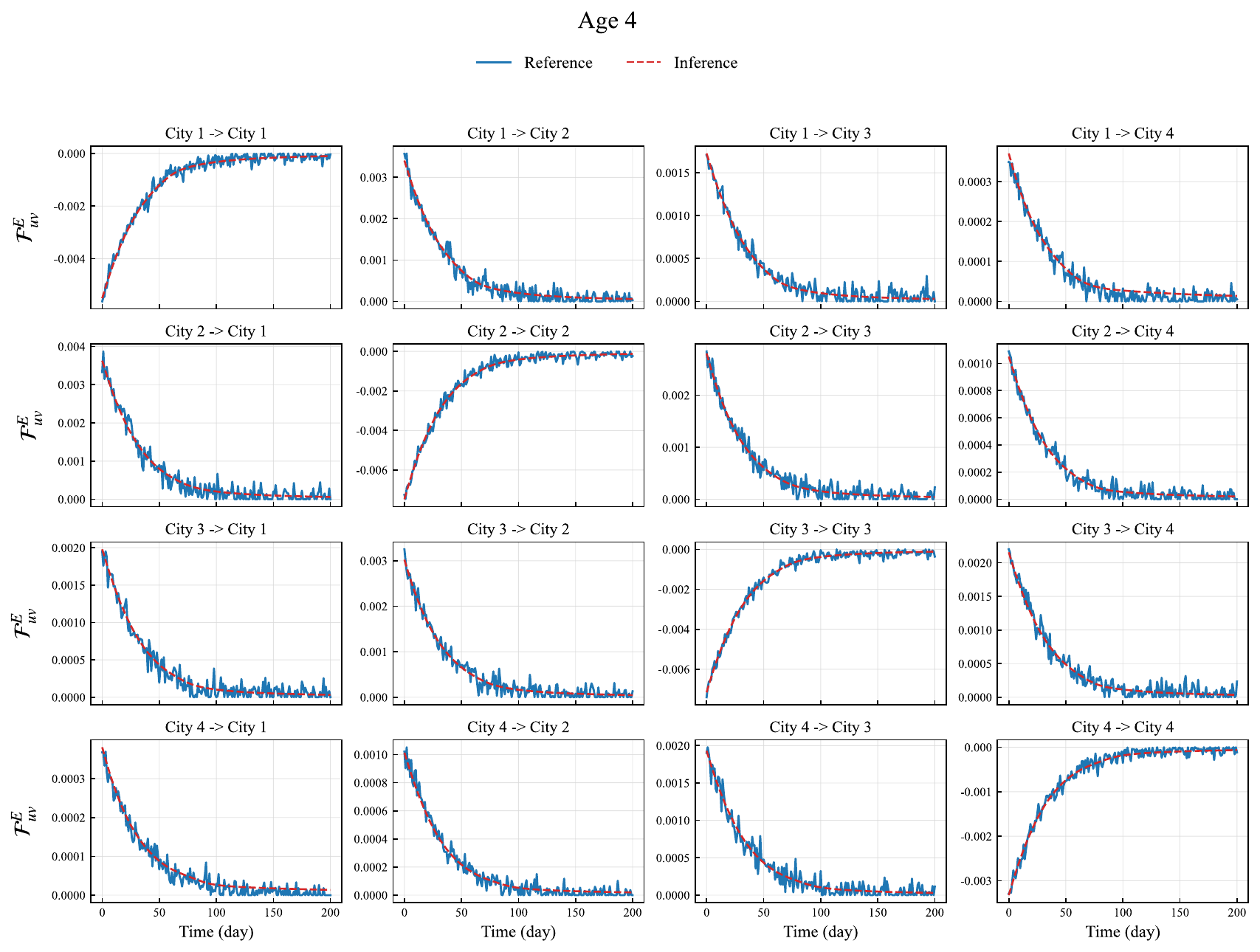}
    {NMCM--CRGM}
    {Age group}
    {\FE}
    {fig:edge-mixed-nmcm-crgm-fe-operators}

\FloatBarrier
\clearpage
\end{document}